\pdfoutput=1

\documentclass[twoside,11pt]{article}

\usepackage[hyphens]{url} % Line break URLs at hyphen.
\usepackage{jmlr2e}
\usepackage{xcolor}
\usepackage{amsmath}
\usepackage{enumerate}
\usepackage{algorithm,algpseudocode}
\usepackage{array} % centering in tables with fixed width
\usepackage{multirow} % text in tables over multiple rows
\usepackage{caption,subcaption} %subfigure
\usepackage{tikz}
\usetikzlibrary{positioning,calc,bayesnet,external}
\DeclareMathOperator*{\argmax}{arg\,max}
\DeclareMathOperator*{\argmin}{arg\,min}

\newcommand{\nodes}{Z}
\newcommand{\node}{z}
\newcommand{\nodei}[1]{\vec{\node}^{\left(#1\right)}}

\newcommand{\f}{f} % Black-box function to maximize
\newcommand{\dataspace}{D} % Space of all datasets
\newcommand{\modelspace}{M} % Space of all deep learning models
\newcommand{\model}{m} % Instance of the model space
\newcommand{\searchspace}{A} % Space of all architectures
\newcommand{\arch}{\alpha} % Instance of the search space
\newcommand{\archopt}{\arch^{*}} % Optimal architecture
\newcommand{\DLLearner}{\Lambda} % Deep Learning Mapping
\newcommand{\loss}{\mathcal{L}} % Loss function
\newcommand{\reg}{\mathcal{R}} % Regularization function
\newcommand{\objective}{\mathcal{O}} % Objective function
\newcommand{\obshist}{H} % Observation history/meta-data set
\newcommand{\D}{d} % Dataset
\newcommand{\Dtrain}{\D_{\text{train}}} % train partition of the dataset
\newcommand{\Dvalid}{\D_{\text{valid}}} % valid partition of the dataset
\newcommand{\opset}{O} % Set of allowed operations
\newcommand{\op}{o} % Symbol for operation, e.g. convolution etc.
\newcommand{\opi}[1]{\op^{\left(#1\right)}}
\newcommand{\modelparams}{\theta} % Symbol for model parameters
\newcommand{\penaltyfct}{\lambda} % Penalty function used for constrained optimization

\renewcommand{\vec}{\boldsymbol}

\jmlrheading{1}{2019}{1-52}{4/00}{10/00}{meila00a}{Martin Wistuba, Ambrish Rawat, Tejaswini Pedapati}

\ShortHeadings{A Survey on Neural Architecture Search}{Martin Wistuba, Ambrish Rawat, Tejaswini Pedapati}
\firstpageno{1}

\begin{document}

\title{A Survey on Neural Architecture Search}

\author{\name Martin Wistuba \email martin.wistuba@ibm.com\\
        \addr IBM Research AI\\
        IBM Technology Campus, Damastown Ind. Park\\
        Dublin, D15HN66, Ireland
        \AND
        \name Ambrish Rawat \email ambrish.rawat@ie.ibm.com\\
        \addr IBM Research AI\\
        IBM Technology Campus, Damastown Ind. Park\\
        Dublin, D15HN66, Ireland
        \AND
        \name Tejaswini Pedapati \email tejaswinip@us.ibm.com\\
        \addr IBM Research AI\\
        IBM T.J. Watson Research Center\\
        Yorktown Heights, NY 10598, USA}

\editor{-}

\maketitle

\begin{abstract}%   <- trailing '%' for backward compatibility of .sty file
The growing interest in both the automation of machine learning and deep learning has inevitably led to the development of a wide variety of automated methods for neural architecture search.
The choice of the network architecture has proven to be critical, and many advances in deep learning spring from its immediate improvements.
However, deep learning techniques are computationally intensive and their application requires a high level of domain knowledge.
Therefore, even partial automation of this process helps to make deep learning more accessible to both researchers and practitioners.
With this survey, we provide a formalism which unifies and categorizes the landscape of existing methods along with a detailed analysis that compares and contrasts the different approaches.
We achieve this via a comprehensive discussion of the commonly adopted architecture search spaces and architecture optimization algorithms based on principles of reinforcement learning and evolutionary algorithms along with approaches that incorporate surrogate and one-shot models.
Additionally, we address the new research directions which include constrained and multi-objective architecture search as well as automated data augmentation, optimizer and activation function search.
\end{abstract}


\section{Introduction}

Deep learning methods are very successful in solving tasks in machine translation, image and speech recognition.
This success is often attributed to their ability to automatically extract features from unstructured data such as audio, image and text.
We are currently witnessing this paradigm shift from the laborious job of manual feature engineering for unstructured data to engineering network components and architectures for deep learning methods.
While architecture modifications do result in significant gains in the performance of deep learning methods, the search for suitable architectures is in itself a time-consuming, arduous and error-prone task.
Within the last two years there has been an insurgence in research efforts by the machine learning community that seeks to automate this search process.
On a high level, this automation is cast as a search problem over a set of decisions that define the different components of a neural network.
The set of feasible solutions to these decisions implicitly defines the search space and the search algorithm is defined by the optimizer.
Arguably, the works by \cite{Zoph2017_Neural} and \cite{Baker2017_Designing} mark the beginning of these efforts where their works demonstrated that good architectures can be discovered with the use of reinforcement learning algorithms.
Shortly thereafter, \cite{Real2017_Large} showed that similar results could also be achieved by the hitherto well studied approaches in neuroevolution~\citep{Floreano2008_Neuroevolution}.
However, both these search approaches consumed hundreds of GPU hours in their respective computations.
Consequently, many of the subsequent works focused on methods that reduce this computational burden.
The successful algorithms along this line of research leverage from the principle of reusing the learned model parameters, with the works of \cite{Cai2018_Efficient} and \cite{Pham2018_ENAS} being the notable mentions.

The design of the search space forms a key component of neural architecture search.
In addition to speeding up the search process, this influences the duration of the search and the quality of the solution.
In the earlier works on neural architecture search, the spaces were designed to primarily search for chain-structured architectures.
However, with branched handcrafted architectures surpassing the classical networks in terms of performance, appropriate search spaces were proposed shortly after the initial publications~\citep{Zoph2018_Learning} and these have since become a norm in this field.

In parallel to these developments, researchers have broadened the horizons of neural architecture search to incorporate objectives that go beyond reducing the search time and generalization error of the found architectures.
Methods that simultaneously handle multiple objective functions have become relevant.
Notable works include methods that attempt to limit the number of model parameters or the like, for efficient deployment on mobile devices~\citep{Tan2018_MnasNet,Kim2017_NEMO}.
Furthermore, the developed techniques for architecture search have been extended for advanced automation of other related components of deep learning.
For instance, the search for activation functions~\citep{Ramachandran2018_Searching} or suitable data augmentation~\citep{Cubuk2018_AutoAugment}.

Currently, the automation of deep learning in the form of neural architecture search is one of the fastest developing areas of machine learning.
With new papers emerging on \url{arXiv.org} each week and major conferences publishing a handful of interesting work, it is easy to lose track.
With this survey, we provide a formalism which unifies the landscape of existing methods.
This formalism allows us to critically examine the different approaches and understand the benefits of different components that contribute to the design and success of neural architecture search.
Along the way, we also highlight some popular misconceptions pitfalls in the current trends of architecture search.
We supplement our criticism with suitable experiments.

Our review is divided into several sections.
In Section~\ref{sec:space}, we discuss various architecture search spaces that have been proposed over time.
We use Section~\ref{sec:opt} to formally define the problem of architecture search.
Then we identify four typical types of optimization methods: reinforcement learning, evolutionary algorithms, surrogate model-based optimization, and one-shot architecture search.
We define these optimization procedures and associate them to existing work and discuss it.
Section~\ref{sec:mo} highlights the architecture search, considering constraints, multiple objective functions and model compression techniques.
Alternate approaches that are motivated from the use of transfer learning are discussed in Section \ref{sec:transfer-learning}.
Similarly, the class of methods that use early termination to fasten the search process are detailed in Section \ref{sec:early-termination}.
Finally, the influence of search procedures on related areas is discussed in Section~\ref{sec:other}.
The discussion is supported with extensive illustrations to elucidate the different methods under a common formalism and relevant experimentation that examines the different aspects of neural architecture search methods.
\section{Neural Architecture Search Space}
\label{sec:space}

From a computational standpoint, neural networks represent a function that transforms input variables $\vec{x}$ to output variables $\vec{\hat{y}}$ through a series of operations.
This can be formalized in the language of computational graphs~\citep{Bauer1974_Computational} where neural networks are represented as directed acyclic graphs with a set of nodes $\nodes$.
Each node $\vec{\node}^{\left(k\right)}$ represents a tensor and is associated with an operation $\op^{\left(k\right)}\in\opset$ on its set of parent nodes $I^{\left(k\right)}$~\citep{Goodfellow2016_Deep}.
The only exception is the input node $\vec{x}$ which has neither a set of parent nodes nor an operation associated to it and is only considered as an input to other nodes.
The computation at a node $k$ amounts to
\begin{equation}
    \label{eq:space-recursive-def}
    \nodei{k} = \opi{k}(I^{\left(k\right)})\,.
\end{equation}
The set of operations includes unary operations such as convolutions, pooling, activation functions or multivariate operations such as concatenation or addition.
For notational convenience, we adopt a convention wherein we fix the unary operations to use concatenation as an implicit merge operation when acting on a set of multiple inputs and often omit it in the representation.
Note that for a neural network, any representation that specifies the set of parents and the operation for each node completely defines the architecture of the network.
We refer to such a representation as $\vec{\arch}$ and use this to unify and outline the different neural architecture search methods under the same framework.

A \emph{neural architecture search space} is a subspace of this general definition of neural architectures.
Its space of operations can be limited and certain constraints may be imposed on the architectures.
In the rest of this survey, we use search space to refer to the set of feasible solutions of a neural architecture search method.
Most works on neural architecture search address the automated search of architectures for the task of image recognition and are consequently concerned with convolutional neural networks (CNNs)~\citep{Lecun1998_Gradient}.
The corresponding search spaces in these works can be broadly classified into two categories.
The first category of search spaces is defined for the graphs that represent an entire neural architecture which we refer to as the global search space and discuss in Section~\ref{sub:space-global}.
We discuss the second category in Section~\ref{sub:space-cell} which assumes that an architecture is a combination of few cells which are repeated to build the complete network.
Although in principle the global search space encompasses the cell-based search space, we make this distinction to highlight some of the nuanced aspects of both these spaces.
An objective comparison of these search spaces is provided in Section~\ref{sub:space-global-vs-cell}.
Another common task tackled with neural architecture search is language modeling where search methods seek to find the architecture for recurrent neural networks (RNNs).
We discuss the commonly used search space in this context in Section~\ref{sub:space-rnn}.

\subsection{Global Search Space}
\label{sub:space-global}

\begin{figure}
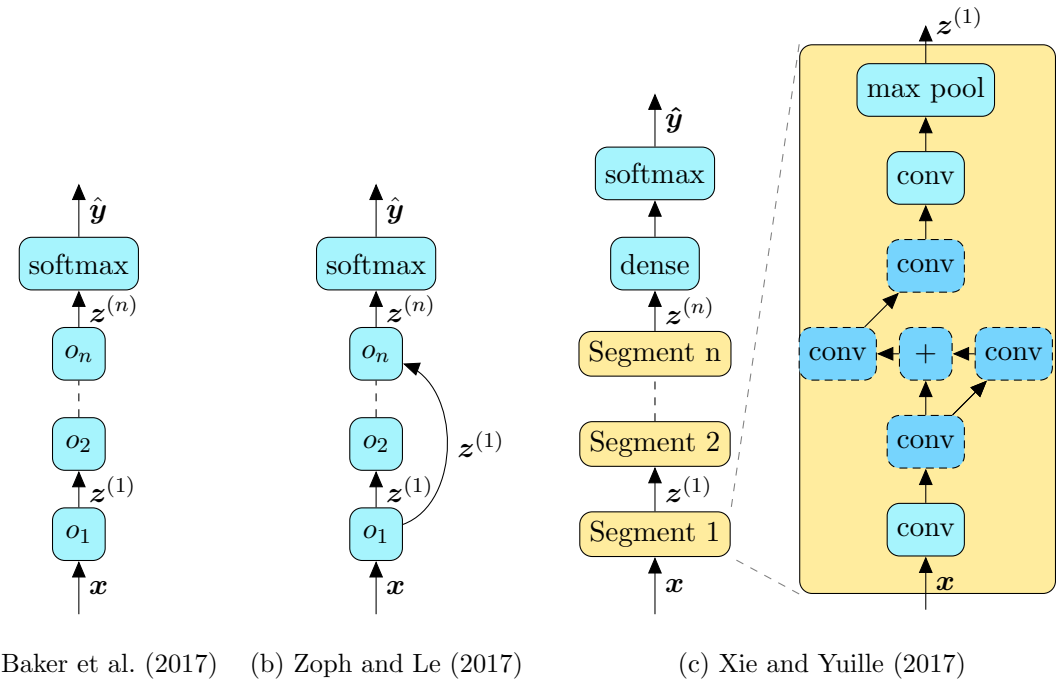

     \centering
     \begin{subfigure}[b]{0.24\textwidth}
         \centering
    \tikzsetnextfilename{space-global-baker}%
    \begin{tikzpicture}
  \input{constants.tikz}
  \node[boxstyle](input) {};
  \node[opstyle](op1) [above of=input] {$o_1$};
  \node[opstyle](op2) [above of=op1] {$o_2$};
  \node[opstyle](opn) [above of=op2] {$o_n$};
  \node[opstyle](softmax) [above of=opn]{softmax};
  \node[boxstyle](output) [above of=softmax] {};
  
  \draw[->]   (input) -- node[right]{$\vec{x}$}  (op1);
  \draw[->]   (op1) -- node[right]{$\nodei{1}$}  (op2);
  \draw[-] [dashed]  (op2) -- (opn);
  \draw[->]   (opn) -- node[right]{$\nodei{n}$} (softmax);
  \draw[->]   (softmax) -- node[right]{$\hat{\vec{y}}$} (output);
\end{tikzpicture}%

         \caption{\cite{Baker2017_Designing}}
         \label{fig:space-global-baker}
     \end{subfigure}
     \hfill
     \begin{subfigure}[b]{0.24\textwidth}
         \begin{flushright}
    \tikzsetnextfilename{space-global-zoph}%
    \begin{tikzpicture}
  \input{constants.tikz}
  \node[boxstyle](input) {};
  \node[opstyle](op1) [above of=input] {$o_1$};
  \node[opstyle](op2) [above of=op1] {$o_2$};
  \node[opstyle](opn) [above of=op2] {$o_n$};
  \node[opstyle](softmax) [above of=opn]{softmax};
  \node[boxstyle](output) [above of=softmax] {};
  
  \draw[->]   (input) -- node[right]{$\vec{x}$}  (op1);
  \draw[->]   (op1) -- node[right]{$\nodei{1}$}  (op2);
  \draw[-] [dashed]  (op2) -- (opn);
  \draw[->]   (opn) -- node[right]{$\nodei{n}$} (softmax);
  \draw[->]   (softmax) -- node[right]{$\hat{\vec{y}}$} (output);
  \draw[->]   (op1) to [bend right=70] node[right]{$\nodei{1}$} (opn);
\end{tikzpicture}%

         \caption{\cite{Zoph2017_Neural}}
         \label{fig:space-global-zoph}
         \end{flushright}
     \end{subfigure}
     \hfill
     \begin{subfigure}[b]{0.48\textwidth}
         \centering
    \tikzsetnextfilename{space-global-xie}%
    \begin{tikzpicture}
  \input{constants.tikz}
  \definecolor{myorangehighlight}{HTML}{76d4fd}
  \tikzstyle{opstylezoom}=[opstyle,node distance=1.17cm]
  \tikzstyle{segmentzoomstyle}=[segmentstyle, minimum height=7.3cm, minimum width=3.4cm]
  \tikzstyle{changeableopstylezoom}=[opstylezoom, fill=myorangehighlight,densely dashed]
  \node[boxstyle](input) {};
  \node[segmentstyle](segment1) [above of=input] {Segment 1};
  \draw[->]   (input) -- node[right]{$\vec{x}$} (segment1);
  \node[segmentstyle](segment2) [above of=segment1] {Segment 2};
  \draw[->]   (segment1) -- node[right]{$\nodei{1}$} (segment2);
  \node[segmentstyle](segmentn) [above of=segment2] {Segment n};
  \draw[-] [dashed]  (segment2) -- (segmentn);
  \node[opstyle](dense) [above of=segmentn] {dense};
  \draw[->]  (segmentn) --node[right]{$\nodei{n}$} (dense);
  \node[opstyle,align=center](softmax) [above of=dense]{softmax};
  \draw[->]   (dense) -- (softmax);
  \node[boxstyle](output) [above of=softmax]{};
  \draw[->]   (softmax) -- node[right]{$\vec{\hat{y}}$} (output);
  
  \node[segmentzoomstyle](segmentzoom2) [above right = -3.5cm and 0.9cm of segmentn] {};
  \draw[-,dashed,gray] (segmentzoom2.north west) -- (segment1.north east);
  \draw[-,dashed,gray] (segmentzoom2.south west) -- (segment1.south east);
  
  \node[changeableopstylezoom](plus4) [right of=segmentn,node distance=3.6cm] {$+$};
  \node[changeableopstylezoom](op3) [right of=plus4] {conv};
  \node[changeableopstylezoom](op4) [left of=plus4] {conv};
  \node[changeableopstylezoom](op2) [below of=plus4] {conv};
  \node[opstylezoom](op1) [below of=op2] {conv};
  \node[changeableopstylezoom](op5) [above of=plus4] {conv};
  \node[opstylezoom](op6) [above of=op5] {conv};
  \node[opstylezoom,align=center](pool) [above of=op6] {max pool};
  \node[boxstyle] (inputsegment2) [below of=op1] {};
  \node[boxstyle,node distance=1cm] (outputsegment2) [above of=pool] {};
  \draw[->]   (op1) -- (op2);
  \draw[->]   (op2) -- (op3);
  \draw[->]   (op3) -- (plus4);
  \draw[->]   (op4) -- (op5);
  \draw[->]   (op2) -- (op3);
  \draw[->]   (op2) -- (plus4);
  \draw[->]   (plus4) -- (op4);
  \draw[->]   (op5) -- (op6);
  \draw[->]   (op6) -- (pool);
  \draw[->]   (inputsegment2) --  node[right]{$\vec{x}$} (op1);
  \draw[->]   (pool) --  node[above right]{$\nodei{1}$} (outputsegment2);
\end{tikzpicture}%

         \caption{\cite{Xie2017_Genetic}}
         \label{fig:space-global-xie}
     \end{subfigure}
  \caption{Global search spaces: (a) chain-structured, (b) with skips, (c)~architecture template, only the connections between the dark blue (dashed) operations are not fixed.}
  \label{fig:space-global}
\end{figure}

Instances belonging to the global search space admit large degrees of freedom regarding the arrangement of operations.
An \emph{architecture template} may be assumed which limits the freedom of admissible structural choices within an architecture definition.
This template often fixes certain aspects of the network graph.
For instance, it may divide the architecture graph into several segments or enforce specific constraints on operations and connections both within and across these segments, thereby limiting the type of architectures that belong to a search space.
Figure \ref{fig:space-global} illustrates examples of architectures from such template-constrained search spaces.
Here, we exclude constraints which enforce a predetermined repeating pattern of subgraphs and dedicate a separate section for the discussion of such search spaces in Section \ref{sub:space-cell}.

The simplest example of a global search space is the \emph{chain-structured search space} as shown in Figure~\ref{fig:space-global-baker}.
In its most primitive form, this search space consists of architectures that can be represented by an arbitrary sequence of ordered nodes such that for any node $\nodei{k}$, $\nodei{k-1}$ is its only parent.
The data flow for neural networks with such architectures simplifies to
\begin{equation}
    \nodei{k} = \opi{k}\left(\left\{\nodei{k-1}\right\}\right)\,.
\end{equation}
\cite{Baker2017_Designing} explore this search space.
They consider a set of operations that includes convolutions, pooling, linear transformations (dense layers) with activation, and global average pooling with different hyperparameter settings such as number of filters, kernel size, stride and pooling size.
Furthermore, they consider additional constraints so as to exclude certain architectures that correspond to patently poor or computationally expensive neural network models.
For instance, architectures with pooling as the first operation do not belong to their search space.
Furthermore, architectures with dense (fully connected) layers applied as high-resolution feature maps, or as feature transformations before other operations like convolutions are excluded from their search space.

In a parallel work around the same time, \cite{Zoph2017_Neural} define a relaxed version of the chain-structured search space.
By permitting arbitrary skip connections to exist between the ordered nodes of a chain-structured architecture, members belonging to this search space exhibit a wider variety of designs.
However, in their definition the operation set is restricted to use only convolutions with different hyperparameter settings.
While they adhere to the basic recursive formulation of the chain-structured search space, they extend it with the inclusion of skip connections. 
For an architecture in this search space, the set of parents of a node $\nodei{k}$ necessarily contains $\nodei{k-1}$ with possible inclusion of other ancestor nodes,
\begin{equation}
    \nodei{k} = \opi{k}\left(\left\{\nodei{k-1}\right\}\cup\left\{\nodei{i}\ |\ \arch_{i,k}=1,\ i<k-1\right\}\right)\,.
\end{equation}
For such nodes, the merge operation is fixed as concatenation.
Figure \ref{fig:space-global-zoph} provides an example of such a skip connection.
As mentioned earlier, we omit the explicit representation of the concatenation.

\begin{figure}[t]
  \centering
    \tikzsetnextfilename{space-mnasnet}%
    \begin{tikzpicture}
\input{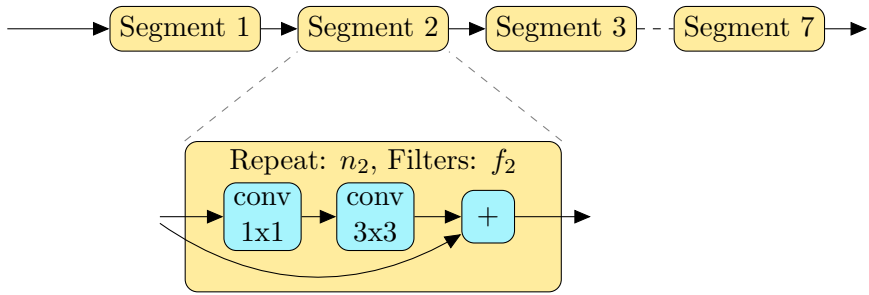}

\tikzstyle{segmentstyle2}=[segmentstyle,node distance=2.5cm]
\tikzstyle{opstyle2}=[opstyle,node distance=1.5cm]
\tikzstyle{boxstyle2}=[boxstyle,node distance=1.7cm]

\node[boxstyle2](input) {};
\edef\prevlabelmnasnet{input}
\foreach \i in {1,...,3}
{
    \xdef\labelmnasnet{segment\i}
    \node[segmentstyle2](\labelmnasnet) [right of=\prevlabelmnasnet] {Segment \i};
    \draw[->]   (\prevlabelmnasnet) -- (\labelmnasnet);
    \xdef\prevlabelmnasnet{\labelmnasnet}
}
\node[segmentstyle2](segment7) [right of=\prevlabelmnasnet] {Segment 7};
\draw[-,dashed]   (\prevlabelmnasnet) -- (segment7);
\node[boxstyle2](output) [right of=segment7] {};
\draw[->]   (segment7) -- (output);

\node[segmentstyle2,minimum width=5cm, minimum height=2cm,text height=-1.2cm](segmentzoom) [below of=segment2] {Repeat: $n_2$, Filters: $f_2$};
\draw[-,dashed,gray] (segmentzoom.north west) -- (segment2.south west);
\draw[-,dashed,gray] (segmentzoom.north east) -- (segment2.south east);

\node[boxstyle](inputcell) [right=-0.6 of $(segmentzoom.north west)!0.5!(segmentzoom.south west)$] {};
\node[opstyle2](op1) [right of=inputcell] {conv\\1x1};
\node[opstyle2](op2) [right of=op1] {conv\\3x3};
\node[opstyle2](op3) [right of=op2] {$+$};
\node[boxstyle,node distance=1.5cm](outputcell) [right of=op3] {};

\draw[->]   (inputcell) -- (op1);
\draw[->]   (op1) -- (op2);
\draw[->]   (op2) -- (op3);
\draw[->]   (op3) -- (outputcell);
\draw[->]   (inputcell) to [bend right=33] (op3);
\end{tikzpicture}%

  \caption{\cite{Tan2018_MnasNet} propose to decompose the architecture into different segments.
  Each segment $i$ has its own pattern (blue operations) which is repeated $n_i$ times and has $f_i$ filters.}
  \label{fig:space-mnasnet}
\end{figure}
\cite{Xie2017_Genetic} consider a similar search space which uses summation instead of concatenation as the merging operation.
Furthermore, the architectures in this search space are no longer sequential in the sense that the set of parents for a node $\nodei{k}$ does not necessarily contain the node $\nodei{k-1}$. 
Moreover, this search space incorporates a template which separates architectures into sequentially connected segments (e.g. three segments are used for image recognition on the CIFAR-10 dataset~\citep{Krizhevsky_CIFAR10}).
Each segment is parameterized by a set of nodes with convolutions as their operation. 
As part of the template, the segments begin with a convolution and conclude with a max pooling operation with a stride of two to reduce feature dimensions (see Figure \ref{fig:space-global-xie}).

Additionally, the maximum number of convolution operations along with their number of filters for each segment is also fixed as part of the template (for the case of CIFAR-10 this is fixed to three convolutions of filter size 64 for the first, four convolutions of filter size 128 for the second, and five convolutions of filter size 256 for the last segment).
With the operations fixed, an adjacency matrix defines the connections in the directed acyclic graphs corresponding to each segment.

An alternative work by \cite{Tan2018_MnasNet} is motivated to look for neural network models for mobile devices that perform efficiently on multiple fronts which include accuracy, inference time and number of parameters.
They design a suitable search space for this purpose that consists of architectures with a hierarchical representation (Figure~\ref{fig:space-mnasnet}).
Architectures in this search space are also formed by sequentially connecting segments.
Segments are further structured to have repeating patterns of operations and each segment is parameterized by the choice of operations and the number of repetitions of the patterns.
The structure of the patterns itself is simplified to be sequentially connected operations with an optional skip connection and an optional choice of stride.
Although they simplify the per-segment search space, they allow segments to be different which differentiates this from the cell-based search space discussed in the following section (Section \ref{sub:space-cell}). 
They argue that this flexibility provides them the necessary apparatus to tackle multi-objective designs.

\subsection{Cell-Based Search Space}
\label{sub:space-cell}
\begin{figure}[t]
  \centering
    \tikzsetnextfilename{space-nasnet-template}%
    \begin{tikzpicture}
  \input{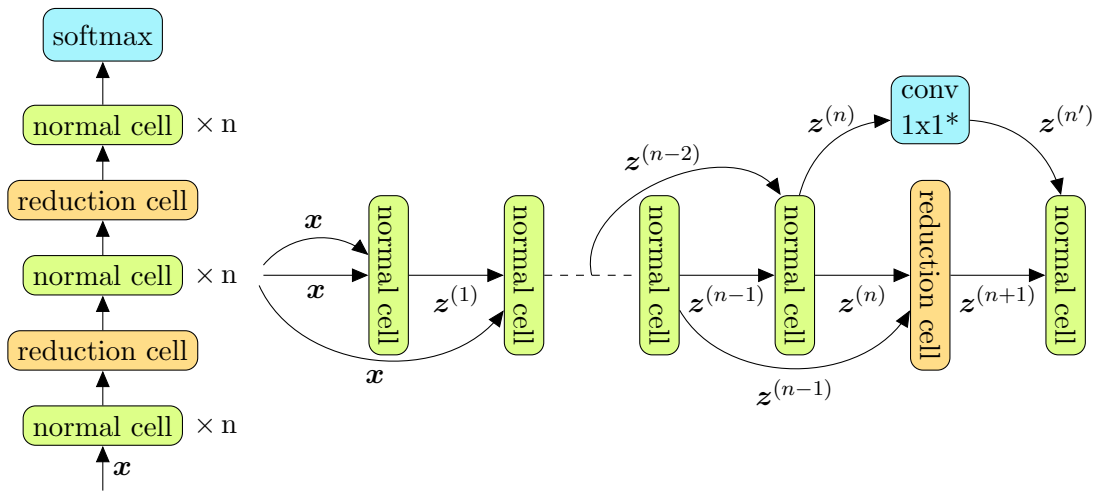}
  \tikzstyle{cellstyle}=[segmentstyle,rotate=-90,node distance=1.8cm,fill=colorCell]
  \tikzstyle{reductioncellstyle}=[cellstyle,fill=colorReductionCell]
  
  \tikzstyle{cellstyle-t}=[segmentstyle,node distance=1cm,fill=colorCell]
  \tikzstyle{reductioncellstyle-t}=[cellstyle-t,fill=colorReductionCell]
  
  % Template
  \node[boxstyle](input-t) at (-1,-3) {};
  \node[cellstyle-t](cell1-t) [above of=input-t] {normal cell};
  \node[boxstyle, node distance=1.5cm](label1-t) [right of=cell1-t] {$\times$\,n};
  \node[reductioncellstyle-t](cell2-t) [above of=cell1-t] {reduction cell};
  \node[cellstyle-t](cell3-t) [above of=cell2-t] {normal cell};
  \node[boxstyle, node distance=1.5cm](label3-t) [right of=cell3-t] {$\times$\,n};
  \node[reductioncellstyle-t](cell4-t) [above of=cell3-t] {reduction cell};
  \node[cellstyle-t](cell5-t) [above of=cell4-t] {normal cell};
  \node[boxstyle, node distance=1.5cm](label5-t) [right of=cell5-t] {$\times$\,n};
  \node[opstyle](sm) [above of=cell5-t] {softmax};
  
  \draw[->]   (input-t) -- node[right]{$\vec{x}$} (cell1-t);
  \draw[->]   (cell1-t) -- (cell2-t);
  \draw[->]   (cell2-t) -- (cell3-t);
  \draw[->]   (cell3-t) -- (cell4-t);
  \draw[->]   (cell4-t) -- (cell5-t);
  \draw[->]   (cell5-t) -- (sm);
  
  % Template Zoom-In
  \node[boxstyle](input) at (1, 0) {};
  \node[cellstyle](cell1) [above of=input] {normal cell};
  \node[cellstyle](cell2) [above of=cell1] {normal cell};
  \node[cellstyle](celln-1) [above of=cell2] {normal cell};
  \node[cellstyle](celln) [above of=celln-1] {normal cell};
  \node[reductioncellstyle](reductioncell) [above of=celln] {reduction cell};
  \node[cellstyle](cell1-1) [above of=reductioncell] {normal cell};
  \node[opstyle,node distance=2.2cm](conv) [above of=reductioncell] {conv\\1x1*};
  
  \coordinate(celln-2) at (5.5,0);

  \draw[->]   (input) -- node[below]{$\vec{x}$} (cell1);
  \draw[->]   (cell1) -- node[below]{$\nodei{1}$} (cell2);
  \draw[-] [dashed]  (cell2) -- (celln-1);
  \draw[->]  (celln-1) --node[below]{$\nodei{n-1}$} (celln);
  \draw[->]  (celln) --node[below]{$\nodei{n}$} (reductioncell);
  \draw[->]  (reductioncell) --node[below]{$\nodei{n+1}$} (cell1-1);
  
  \draw[->]   (input) to [bend left=45] node[above]{$\vec{x}$} (cell1);
  \draw[->]   (input) to [bend right=60] node[below]{$\vec{x}$} (cell2);
  \draw[->]   (celln-2) to [bend left=80] node[above]{$\nodei{n-2}$} (celln);
  \draw[->]   (celln-1) to [bend right=60] node[below]{$\nodei{n-1}$} (reductioncell);
  \draw[->]   (celln) to [bend left=36] node[above]{$\nodei{n}$} (conv);
  \draw[->]   (conv) to [bend left=36] node[above right]{$\nodei{n'}$} (cell1-1);
\end{tikzpicture}%

  \caption{Structure of the NASNet search space instances.
  $n$ normal cells followed by a reduction cell.
  This sequence is repeated several times, the reduction cell might be repeated.
  This decision is a hyperparameter and depends on the image resolution.
  The 1x1* convolution is a special operation which converts $\nodei{n}$ to match the shape of $\nodei{n+1}$.}
  \label{fig:space-nasnet-template}
\end{figure}
A cell-based search space builds upon the observation that many effective handcrafted architectures are designed with repetitions of fixed structures.
Such architectures often consist of smaller-sized graphs that are stacked to form a larger architecture.
Across the literature, these repeating structures have been interchangeably referred to as cells, blocks or units.
In the context of this survey, we refer to them as \emph{cells}.
Not only is this design known to yield high performing models, it also enables easy generalization across datasets and tasks as these units can be flexibly stacked to build larger or smaller networks.
\cite{Zoph2018_Learning} were one of the first to explore such a search space leading to the popular architecture called NASNet.
Post the development of this work, other cell-based search spaces have been proposed.
Most of these broadly stick to the structure proposed by \cite{Zoph2018_Learning} with small modifications to the choice of operations and cell-combination strategies.
In the remaining section we define and discuss these different approaches to cell-based search spaces.

\begin{figure}[t]
  \centering
    \tikzsetnextfilename{space-nasnet-reduction-cell}%
    \begin{tikzpicture}
  \input{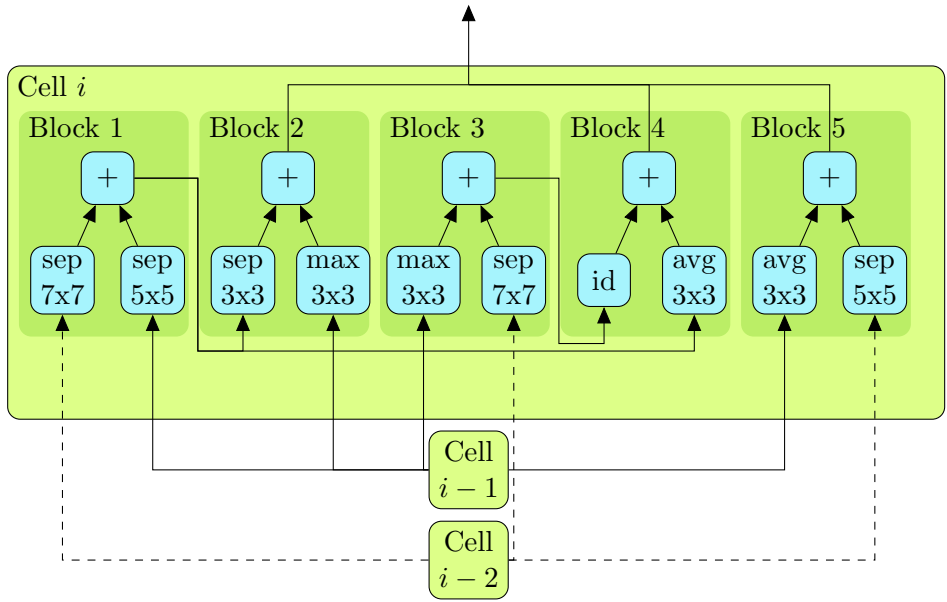}
  \tikzstyle{cellstyle}=[boxstyle,draw=black,fill=colorCell,align=center]
  \tikzstyle{blockstyle}=[boxstyle,fill=colorBlock, minimum width=2cm, minimum height=3cm,align=left,text width=2cm, text height=-2.3cm, node distance = 2.4cm]
  \tikzstyle{input1nodestyle}=[right, pos=0.805]
  \tikzstyle{input2nodestyle}=[right, pos=0.88]
  
  % Background
  \node[cellstyle, minimum width=12.2cm, minimum height=4.7cm,align=left,text width = 12.2cm, text height=-3.9cm](bg) at (0.7, 0.5) {Cell $i$};
  
  \node[blockstyle](block1) at (-4.25, 0.75) {Block 1};
  \node[blockstyle](block2) [right of=block1] {Block 2};
  \node[blockstyle](block3) [right of=block2] {Block 3};
  \node[blockstyle](block4) [right of=block3] {Block 4};
  \node[blockstyle](block5) [right of=block4] {Block 5};
  
  % Block 3
  \node[opstyle](op3-1) {max\\3x3};
  \node[opstyle](op3-2) [right of=op3-1] {sep\\7x7};
  \node[opstyle](add3) [above=1 of $(op3-1)!0.5!(op3-2)$] {$+$};
  \draw[->]   (op3-1) -- (add3);
  \draw[->]   (op3-2) -- (add3);
  
  % Block 2
  \node[opstyle](op2-2) [left of=op3-1] {max\\3x3};
  \node[opstyle](op2-1) [left of=op2-2] {sep\\3x3};
  \node[opstyle](add2) [above=1 of $(op2-1)!0.5!(op2-2)$] {$+$};
  \draw[->]   (op2-1) -- (add2);
  \draw[->]   (op2-2) -- (add2);
  
  % Block 1
  \node[opstyle](op1-2) [left of=op2-1] {sep\\5x5};
  \node[opstyle](op1-1) [left of=op1-2] {sep\\7x7};
  \node[opstyle](add1) [above=1 of $(op1-1)!0.5!(op1-2)$] {$+$};
  \draw[->]   (op1-1) -- (add1);
  \draw[->]   (op1-2) -- (add1);
  
  % Block 4
  \node[opstyle](op4-1) [right of=op3-2] {id};
  \node[opstyle](op4-2) [right of=op4-1] {avg\\3x3};
  \node[opstyle](add4) [above=1 of $(op4-1)!0.5!(op4-2)$] {$+$};
  \draw[->]   (op4-1) -- (add4);
  \draw[->]   (op4-2) -- (add4);
  
  % Block 5
  \node[opstyle](op5-1) [right of=op4-2] {avg\\3x3};
  \node[opstyle](op5-2) [right of=op5-1] {sep\\5x5};
  \node[opstyle](add5) [above=1 of $(op5-1)!0.5!(op5-2)$] {$+$};
  \draw[->]   (op5-1) -- (add5);
  \draw[->]   (op5-2) -- (add5);
  
  % Concat and Output
  \coordinate(concat) at (0.6,2.6);
  \node[boxstyle](output) [above of=concat] {};
  \draw   (add2) |- (concat);
  \draw   (add4) |- (concat);
  \draw   (add5) |- (concat);
  \draw[->]   (concat) -- (output);
  
  % Input nodes
  \node[cellstyle](input1) [below=2 of $(op3-1)!0.5!(op3-2)$]{Cell\\$i-1$};
  \draw[->]   (input1) -| (op1-2);
  \draw[->]   (input1) -| (op2-2);
  \draw[->]   (input1) -| (op3-1);
  \draw[->]   (input1) -| (op5-1);
  
  \node[cellstyle](input2) [below of=input1]{Cell\\$i-2$};
  \draw[->,dashed]   (input2) -| (op1-1);
  \draw[->,dashed]   (input2) -| (op3-2);
  \draw[->,dashed]   (input2) -| (op5-2);
  
  \draw[->]   (add1) -- ++(1.2,0) -- ++(0,-2.3) -- ++(0.6,0) --  (op2-1);
  \draw[->]   (add1) -- ++(1.2,0) -- ++(0,-2.3) -- ++(6.6,0) -- (op4-2);
  \draw[->]   (add3) -- ++(1.2,0) -- ++(0,-2.2) -- ++(0.6,0) -- (op4-1);
\end{tikzpicture}%

  \caption{Reduction cell of the NASNet-A architecture~\citep{Zoph2018_Learning} as one example how a cell in the NASNet search space can look like.
  Blocks can be used as input for other blocks (e.g. block 1 and 3), unused blocks are concatenated and are the output of the cell.}
  \label{fig:space-nasnet-cell}
\end{figure}
In the cell-based search space a network is constructed by repeating a structure called a cell in a prespecified arrangement as determined by a template (see Figure \ref{fig:space-nasnet-template}).
A cell is often a small directed acyclic graph representing a feature transformation.
While the topology of the cell is maintained across the network, its hyperparameters are often varied.
For instance the number of filters for the convolution operations can differ across different cells and a template usually specifies the fixed policy for varying these across the network structures.
One aspect of cell-based search spaces that demands attention is their handling of tensor dimensions when building the entire network.
While some approaches have dedicated cells to take care of dimension reduction like the reduction cells in the work by \cite{Zoph2018_Learning}, others achieve this by introducing pooling operations between different units in the network \citep{Zhong2018_Practical}. 
The prespecified template for connecting the various cells localizes the search to structures within a cell.
Often the template also includes a set of initial convolution layers which capture low-level features for the given dataset.
It is also worth remarking that most approaches that seek to learn a topology limit the hyperparameter choice to smaller values during the search process thereby making the search process efficient.
The final proposed network is trained with a larger number of filters ($f$) and more repetitions of the cells ($n$) with an aim to achieve better performance~\citep{Zoph2018_Learning}.

\begin{figure}[t]
  \centering
    \tikzsetnextfilename{space-blockqnn-space}%
    \begin{tikzpicture}
  \input{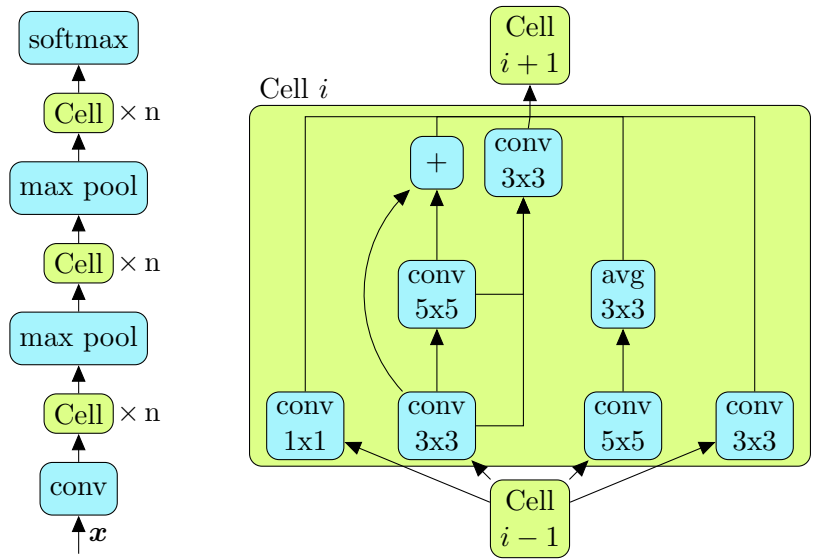}
  \tikzstyle{oprotatedstyle}=[opstyle,node distance=1.75cm]
  \tikzstyle{cellrotatedstyle}=[cellstyle,node distance=1.7cm]
  
  \tikzstyle{cellstyle-t}=[segmentstyle,node distance=1cm,fill=colorCell]
  \tikzstyle{opstyle-t}=[opstyle,node distance=1cm]
  
  % Template
  \node[boxstyle](input-t) at (-6,-0.6) {};
  \node[opstyle-t](conv-t) [above of=input-t] {conv};
  \node[cellstyle-t](cell1-t) [above of=conv-t] {Cell};
  \node[boxstyle, node distance=0.8cm](label1-t) [right of=cell1-t] {$\times$\,n};
  \node[opstyle-t](cell2-t) [above of=cell1-t] {max pool};
  \node[cellstyle-t](cell3-t) [above of=cell2-t] {Cell};
  \node[boxstyle, node distance=0.8cm](label3-t) [right of=cell3-t] {$\times$\,n};
  \node[opstyle-t](cell4-t) [above of=cell3-t] {max pool};
  \node[cellstyle-t](cell5-t) [above of=cell4-t] {Cell};
  \node[boxstyle, node distance=0.8cm](label5-t) [right of=cell5-t] {$\times$\,n};
  \node[opstyle-t](sm) [above of=cell5-t] {softmax};
  
  \draw[->]   (input-t) -- node[right]{$\vec{x}$} (conv-t);
  \draw[->]   (conv-t) -- (cell1-t);
  \draw[->]   (cell1-t) -- (cell2-t);
  \draw[->]   (cell2-t) -- (cell3-t);
  \draw[->]   (cell3-t) -- (cell4-t);
  \draw[->]   (cell4-t) -- (cell5-t);
  \draw[->]   (cell5-t) -- (sm);
  
  % Background
  \node[cellrotatedstyle, minimum width=7.2cm, minimum height=4.8cm,align=left,text width=7.2cm, text height=-5.0cm](bg) at (0, 3.1) {Cell $i$};
  
  % Input node
  \node[cellrotatedstyle](input) {Cell\\$i-1$};
  
  % Branch 2
  \node[oprotatedstyle](op2) [above left of=input] {conv\\3x3};
  \node[oprotatedstyle](op3) [above of=op2] {conv\\5x5};
  \node[oprotatedstyle](add1) [above of=op3] {$+$};
  \node[oprotatedstyle, node distance=1.15cm](op4) [right of=add1] {conv\\3x3};
  
  % Branch 1
  \node[oprotatedstyle](op1) [left of=op2] {conv\\1x1};
  
   % Branch 3
  \node[oprotatedstyle](op5) [above right of=input] {conv\\5x5};
  \node[oprotatedstyle](op6) [above of=op5] {avg\\3x3};
  
  % Branch 4
  \node[oprotatedstyle](op7) [right of=op5] {conv\\3x3};
  
  % Concat and Output
  \coordinate(concat) at (0,5.35);
  
  % Output
  \node[cellrotatedstyle, node distance=6.3cm](output) [above of=input] {Cell\\$i+1$};
  
  % Edges
  \draw[->]   (input) -- (op1);
  \draw[->]   (input) -- (op2);
  \draw[->]   (input) -- (op5);
  \draw[->]   (input) -- (op7);
  
  \draw[->]   (op2) -- (op3);
  \draw[->]   (op2) to [bend left=45] (add1);
  \draw[->]   (op3) -- (add1);
  \draw[->]   (op2) -| (op4);
  \draw[->]   (op3) -| (op4);
  
  \draw[->]   (op5) -- (op6);
  
  \draw[-]   (op1) |- (concat);
  \draw[-]   (add1) |- (concat);
  \draw[-]   (op4) -- (concat);
  \draw[-]   (op6) |- (concat);
  \draw[-]   (op7) |- (concat);
  \draw[->]   (concat) -- (output);
\end{tikzpicture}%

  \caption{Architecture template (left) and the Block-QNN-B cell discovered by \cite{Zhong2018_Practical} (right).}
  \label{fig:space-blockqnn-space}
\end{figure}
\cite{Zoph2018_Learning} are one of the first to propose a cell-based approach, popularly referred to as the \emph{NASNet search space}.
The architectures in this search space consider a template of cells as visualized in Figure~\ref{fig:space-nasnet-template}.
This search space distinguishes between two cell types, namely normal cells and reduction cells, which handle the feature dimensions across the architecture.
Operations in a normal cell have a stride of one and do not change the dimensions of the feature maps, whereas the first set of operations in a reduction cell have a stride of two and it consequently halve the spatial dimensions of the feature maps.
Each cell consists of $b$ blocks, where $b$ is a search space hyperparameter (with $b=5$ being the most popular choice). 
Each block is completely defined by its two inputs and the corresponding operations.
The computation flow for a block is defined by
\begin{equation}
    \vec{\node}^{\left(\text{block}_i\right)} = \opi{i_1}\left(\left\{\nodei{i_1}\right\}\right)+\opi{i_2}\left(\left\{\nodei{i_2}\right\}\right)\,.
\end{equation}
An example for the cell structure is visualized in Figure~\ref{fig:space-nasnet-cell}.
In their original formulation, \cite{Zoph2018_Learning} also consider concatenation as a possible merge operation of $\opi{i_1}$ and $\opi{i_2}$.
However in their experiments they note that architectures with summation operation dominate those with concatenation in terms of accuracy.
Therefore, later works such as \cite{Liu2018_Progressive} fix this decision to summation.
For a block, the set of possible inputs include the output of one of the previous two cells and the output of a previously defined block within a cell.
By including the output from previous two cells as candidates for the input to the block, this search space includes instances with skip connections across cells.
The output of a cell is determined as concatenation of the outputs of all blocks within the cell which do not serve as inputs to any other block.
Cells from this search space differ in their choice of the inputs and the operations for the different blocks.

Concurrently with \cite{Zoph2018_Learning}, \cite{Zhong2018_Practical} propose a similar cell-based search space as shown in Figure~\ref{fig:space-blockqnn-space}.
The key differences lie in their definition of cells and the use of fixed max-pooling layers as opposed to reduction cells for handling feature dimensions in the architecture template.
Since they do not decompose the cell structure any further, graphs comprising of arbitrary connections between different nodes are admissible as cells in this search space.
This alternate definition of cell admits higher degrees of freedom than the normal cell of the NASNet search space, for instance the number of operations within each cell is not fixed.
In addition to the number of operations, the instances of cells also differ in the type of operations and their input.
This search space formulation does not allow skip connections across different cells.
An example architecture discovered is presented in Figure~\ref{fig:space-blockqnn-space}.
\begin{figure}[t]
  \centering
    \tikzsetnextfilename{space-hierarchical}%
    \begin{tikzpicture}
  \input{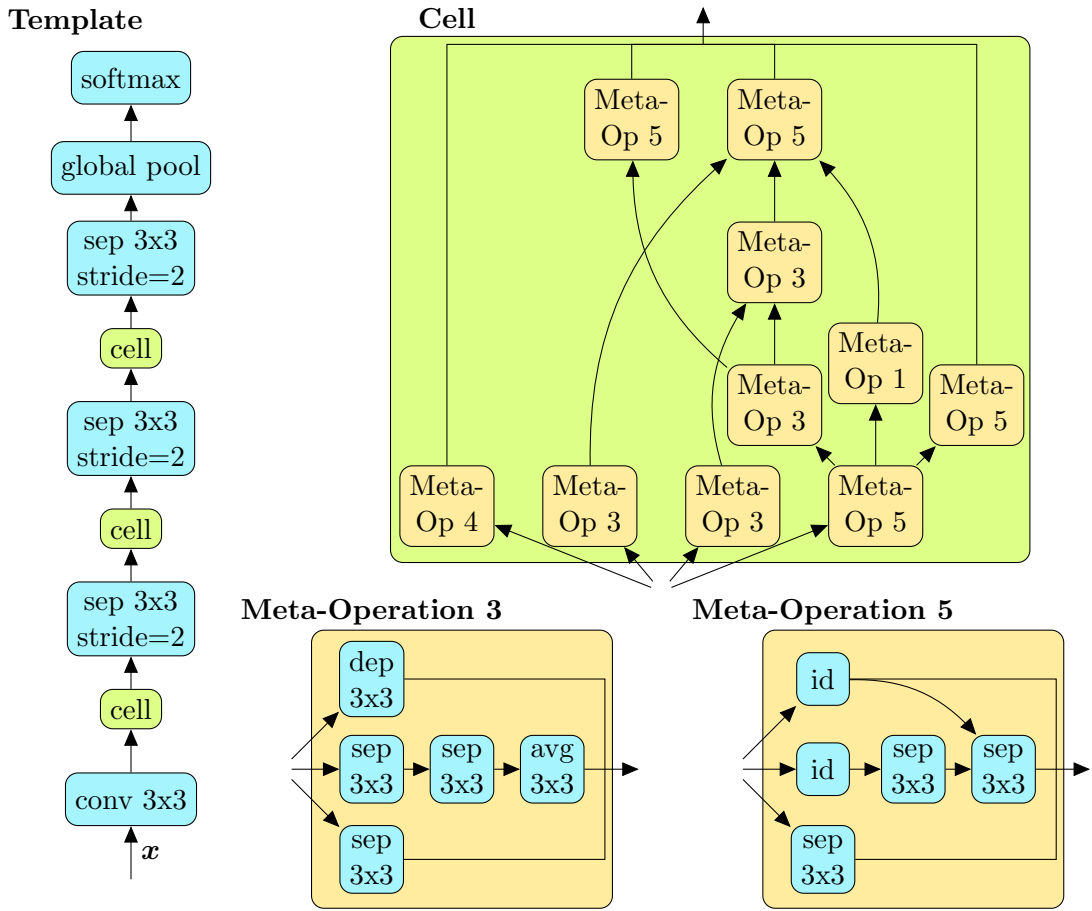}
  \tikzstyle{cellstyle-t}=[cellstyle,node distance=1.2cm]
  \tikzstyle{opstyle-t}=[opstyle,node distance=1.2cm]
  \tikzstyle{segmentstyle-c}=[segmentstyle,node distance=1.9cm]
  
  % Backgrounds
  \node[segmentstyle, minimum width=4.0cm, minimum height=3.7cm](bgop3) at (-1.6, 1) {};
  \node[segmentstyle, minimum width=4.0cm, minimum height=3.7cm](bgop5) at (4.4, 1) {};
  \node[cellstyle, minimum width=8.5cm, minimum height=7.0cm](bgcell) at (1.7, 7.25) {};
  
  % Template
  \node[boxstyle](input-t) at (-6,-0.6) {};
  \node[opstyle-t](conv-t) [above of=input-t] {conv 3x3};
  \node[cellstyle-t](cell1-t) [above of=conv-t] {cell};
  \node[opstyle-t](cell2-t) [above of=cell1-t] {sep 3x3\\stride=2};
  \node[cellstyle-t](cell3-t) [above of=cell2-t] {cell};
  \node[opstyle-t](cell4-t) [above of=cell3-t] {sep 3x3\\stride=2};
  \node[cellstyle-t](cell5-t) [above of=cell4-t] {cell};
  \node[opstyle-t](cell6-t) [above of=cell5-t] {sep 3x3\\stride=2};
  \node[opstyle-t](globalpool-t) [above of=cell6-t] {global pool};
  \node[opstyle-t](sm) [above of=globalpool-t] {softmax};
  \node[boxstyle,node distance=1.05cm](label-t) [above left of=sm] {\textbf{Template}};
  
  \draw[->]   (input-t) -- node[right]{$\vec{x}$} (conv-t);
  \draw[->]   (conv-t) -- (cell1-t);
  \draw[->]   (cell1-t) -- (cell2-t);
  \draw[->]   (cell2-t) -- (cell3-t);
  \draw[->]   (cell3-t) -- (cell4-t);
  \draw[->]   (cell4-t) -- (cell5-t);
  \draw[->]   (cell5-t) -- (cell6-t);
  \draw[->]   (cell6-t) -- (globalpool-t);
  \draw[->]   (globalpool-t) -- (sm);
  
  % Cell
  \node[segmentstyle-c](cellop21) at (2,4.5) {Meta-\\Op 3};
  \node[segmentstyle-c](cellop11) [right of=cellop21] {Meta-\\Op 5};
  \node[segmentstyle-c](cellop12) [above left of=cellop11] {Meta-\\Op 3};
  \node[segmentstyle-c](cellop13) [above of=cellop12] {Meta-\\Op 3};
  \node[segmentstyle-c](cellop14) [above of=cellop13] {Meta-\\Op 5};
  \node[segmentstyle-c](cellop15) [above of=cellop11] {Meta-\\Op 1};
  \node[segmentstyle-c](cellop16) [above right of=cellop11] {Meta-\\Op 5};
  \node[segmentstyle-c](cellop17) [left of=cellop14] {Meta-\\Op 5};
  \node[segmentstyle-c](cellop31) [left of=cellop21] {Meta-\\Op 3};
  \node[segmentstyle-c](cellop41) [left of=cellop31] {Meta-\\Op 4};
  \node[boxstyle,node distance=6.5cm](celllabel) [above of=cellop41] {\textbf{Cell}};
  \node[boxstyle](cellinput) [below=1 of $(cellop21)!0.5!(cellop31)$] {};
  \coordinate[above=1 of $(cellop17)!0.5!(cellop14)$] (cellconcat);
  \coordinate[above of=cellconcat,node distance=0.5cm] (celloutput);
  
  \draw[->]   (cellinput) -- (cellop11);
  \draw[->]   (cellinput) -- (cellop21);
  \draw[->]   (cellinput) -- (cellop31);
  \draw[->]   (cellinput) -- (cellop41);
  \draw[->]   (cellop11) -- (cellop12);
  \draw[->]   (cellop12) -- (cellop13);
  \draw[->]   (cellop13) -- (cellop14);
  \draw[->]   (cellop11) -- (cellop15);
  \draw[->]   (cellop11) -- (cellop16);
  \draw[->]   (cellop21) to [bend left=25] (cellop13);
  \draw[->]   (cellop31) to [bend left=25] (cellop14);
  \draw[->]   (cellop15) to [bend right=25] (cellop14);
  \draw[->]   (cellop12) to [bend left=25] (cellop17);

  \draw[-]   (cellop14) |- (cellconcat);  
  \draw[-]   (cellop17) |- (cellconcat);
  \draw[-]   (cellop16) |- (cellconcat);
  \draw[-]   (cellop41) |- (cellconcat);
  \draw[->]   (cellconcat) -- (celloutput);
  
  % Motif 3
  \node[boxstyle](motif3input) at (-4,1) {};
  \node[opstyle](op31) [right of=motif3input] {sep\\3x3};
  \node[opstyle](op32) [right of=op31] {sep\\3x3};
  \node[opstyle](op33) [right of=op32] {avg\\3x3};
  \node[opstyle](op34) [above of=op31] {dep\\3x3};
  \node[opstyle](op35) [below of=op31] {sep\\3x3};
  \node[boxstyle,node distance=0.9cm](label3) [above of=op34] {\textbf{Meta-Operation 3}};
  \coordinate[right of=op33,node distance=0.7cm] (concat3);
  \coordinate[right of=concat3,node distance=0.45cm] (output3);
  \draw[->]   (motif3input) -- (op31);
  \draw[->]   (motif3input) -- (op34);
  \draw[->]   (motif3input) -- (op35);
  \draw[->]   (op31) -- (op32);
  \draw[->]   (op32) -- (op33);
  \draw[-]   (op33) -- (concat3);
  \draw[-]   (op34) -| (concat3);
  \draw[-]   (op35) -| (concat3);
  \draw[->]   (concat3) -- (output3);
  
  % Motif 5
  \node[boxstyle](motif5input) at (2,1) {};
  \node[opstyle](op55) [right of=motif5input] {id};
  \node[opstyle](op51) [right of=op55] {sep\\3x3};
  \node[opstyle](op52) [right of=op51] {sep\\3x3};
  \node[opstyle](op53) [below of=op55] {sep\\3x3};
  \node[opstyle](op54) [above of=op55] {id};
  \node[boxstyle,node distance=0.9cm](label5) [above of=op54] {\textbf{Meta-Operation 5}};
  \coordinate[right of=op52,node distance=0.7cm] (concat5);
  \coordinate[right of=concat5,node distance=0.45cm] (output5);
  \draw[->]   (motif5input) -- (op55);
  \draw[->]   (motif5input) -- (op53);
  \draw[->]   (motif5input) -- (op54);
  \draw[->]   (op55) -- (op51);
  \draw[->]   (op51) -- (op52);
  \draw[->]   (op54) to [bend left=25] (op52);
  \draw[-]   (op52) -- (concat5);
  \draw[-]   (op53) -| (concat5);
  \draw[-]   (op54) -| (concat5);
  \draw[->]   (concat5) -- (output5);
\end{tikzpicture}%

  \caption{The search space proposed by \cite{Liu2018_Hierarchical} is based on an architecture template (left) which defines the sequence of cells and reduction operations.
  The choices made about an architecture is the set of meta-operations and their arrangement within the cell (right).}
  \label{fig:space-hierarchical}
\end{figure}

\cite{Liu2018_Hierarchical} propose a search space similar to the one proposed by \cite{Zhong2018_Practical}.
An architecture template that describes their high-level definition of an architecture as shown in Figure~\ref{fig:space-hierarchical}.
The main difference is that the search space is decomposed into a hierarchical search space.
The first level of hierarchy defines a fixed numbers of meta-operations.
A meta-operation is the connection of few operations to a larger segment (Figure~\ref{fig:space-hierarchical}).
The second level represents the cell by connections between the meta-operations.

\begin{figure}[t]
  \centering
    \tikzsetnextfilename{space-dpp-template}%
    \begin{tikzpicture}
  \input{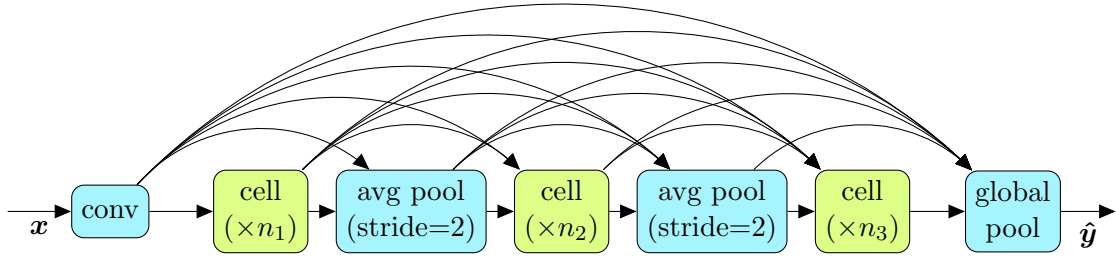}
  \tikzstyle{cellstyle}=[segmentstyle,fill=colorCell,node distance=2cm]
  \tikzstyle{opstyle2}=[opstyle,node distance=2cm]
  
  \node[boxstyle](input) {};
  \node[opstyle2,node distance=1.5cm](conv) [right of=input] {conv};
  \node[cellstyle](cell1) [right of=conv] {cell\\$\left(\times n_{1}\right)$};
  \node[opstyle2](pool1) [right of=cell1] {avg pool\\(stride=2)};
  \node[cellstyle](cell2) [right of=pool1] {cell\\$\left(\times n_{2}\right)$};
  \node[opstyle2](pool2) [right of=cell2] {avg pool\\(stride=2)};
  \node[cellstyle](cell3) [right of=pool2] {cell\\$\left(\times n_{3}\right)$};
  \node[opstyle2](pool3) [right of=cell3] {global\\pool};
  \node[boxstyle,node distance=1.5cm](output) [right of=pool3] {};
  
  \draw[->]   (input) -- node[below]{$\vec{x}$} (conv);
  \draw[->]   (conv) -- (cell1);
  \draw[->]   (cell1) -- (pool1);
  \draw[->]   (pool1) -- (cell2);
  \draw[->]   (cell2) -- (pool2);
  \draw[->]   (pool2) -- (cell3);
  \draw[->]   (cell3) -- (pool3);
  \draw[->]   (pool3) -- node[below]{$\vec{\hat{y}}$} (output);
  
  % Dense connections
  \draw[->]   (conv) to [bend left=45] (cell2);
  \draw[->]   (conv) to [bend left=45] (pool1);
  \draw[->]   (conv) to [bend left=45] (cell3);
  \draw[->]   (conv) to [bend left=45] (pool2);
  \draw[->]   (conv) to [bend left=45] (pool3);
  
  \draw[->]   (cell1) to [bend left=45] (cell2);
  \draw[->]   (cell1) to [bend left=45] (cell3);
  \draw[->]   (cell1) to [bend left=45] (pool2);
  \draw[->]   (cell1) to [bend left=45] (pool3);
  
  \draw[->]   (pool1) to [bend left=45] (cell3);
  \draw[->]   (pool1) to [bend left=45] (pool2);
  \draw[->]   (pool1) to [bend left=45] (pool3);
  
  \draw[->]   (cell2) to [bend left=45] (cell3);
  \draw[->]   (cell2) to [bend left=45] (pool3);
  
  \draw[->]   (pool2) to [bend left=45] (pool3);
\end{tikzpicture}%

  \caption{Mobile search spaces as used by \cite{Dong2018_DPP}. The entire network including the cells are densely connected.}
  \label{fig:space-dpp-template}
\end{figure}
The cell-based design paradigm has also been used for defining search spaces that are suitable for mobile devices.
\cite{Dong2018_DPP} propose a search space specifically aimed at meeting such requirements which include objectives like fewer parameters and fast inference time.
Architectures in this search space comprise of cells without branches and a fixed internal structure that alternates two normalization (e.g. batch normalization) and convolution layers.
All operations are densely connected~\citep{Huang2017_Densely} as shown in Figure~\ref{fig:space-dpp-template} which also applies for those in the cell.
The number of choices is significantly smaller than the previously discussed cell-based search spaces, making this a less challenging optimization task.

\subsection{Global vs. Cell-Based Search Space}\label{sub:space-global-vs-cell}

So far, there has been no detailed comparative study of the different search spaces proposed in the vast literature of neural architecture search.
However, cell-based search spaces, in particular the NASNet search space, are the most popular choices for developing new methods. 
Most works that examine both search spaces, the global and the cell-based space, support this choice by providing empirical evidence that well-performing can be obtained in the cell-based search space~\citep{Pham2018_ENAS}.
Regardless, cell-based search spaces benefit from the advantage that discovered architectures can be easily transferred across datasets.
Moreover, the complexity of the architectures can be varied almost arbitrarily by changing the number of filters and cells.
In general, the architectures belonging to the global search space do not exhibit all these properties, but specific cases may benefit from some of them.
For instance, architectures can be naturally modified by varying the number of filters, but transferring a discovered architecture to a new datasets with different input shapes or deepening the architecture is not trivial.
Interestingly, \cite{Tan2018_MnasNet} endorse the use of the global search space when searching for mobile architectures.
They base this argument on a hypothesis that diversity of layers is critical for achieving both high accuracy and low latency for deployment in mobile devices and that these cannot be provided by conventional cell-based search spaces.
Interestingly, \cite{Hu2018_Macro} remark on the importance of selecting initial architectures for searching in the global search space and show that architectures with comparable performance to the ones discovered in cell-based search spaces can be easily arrived at with appropriate initial conditions.
This is an interesting insight which might revive the use of global search space in the development of new methods.
However, the choice of useful initial architecture might be task-dependent and the guidelines for this selection remain unclear. 

\begin{figure}[t]
  \centering
    \tikzsetnextfilename{space-rnn}%
    \begin{tikzpicture}
\input{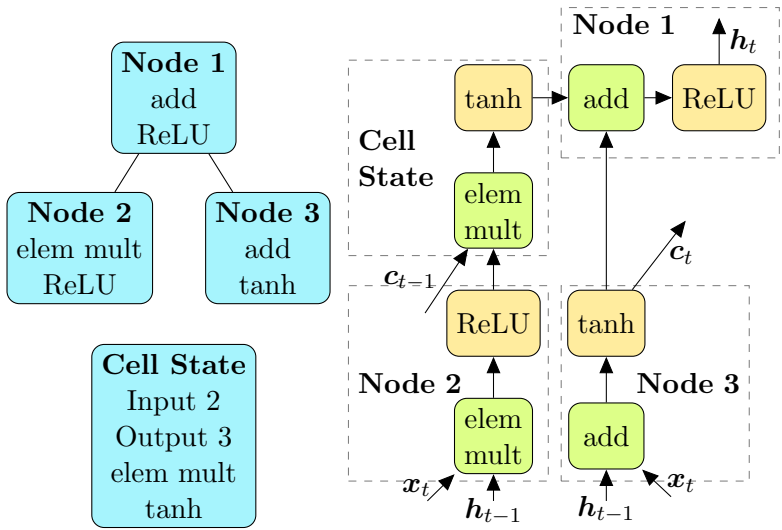}
\tikzstyle{boxstyle}=[rectangle,draw=black,minimum size=25,minimum width=20mm,align=center,rounded corners=1ex]
\tikzstyle{opstyle}=[boxstyle,fill=colorOp,node distance=1.5cm,minimum width=10mm]
\tikzstyle{unaryopstyle}=[opstyle,fill=colorSegment]
\tikzstyle{binaryopstyle}=[opstyle,fill=colorCell]

\node[rectangle,draw=gray,dashed, minimum width=2.9cm, minimum height=2cm,text height=-1.3cm,text width=2.6cm](node1) at (7.85,4.7) {\textbf{Node 1}};
\node[rectangle,draw=gray,dashed, minimum width=2.5cm, minimum height=2.6cm,text width=2.4cm](node2) at (4.9,0.7) {\textbf{Node 2}};
\node[rectangle,draw=gray,dashed, minimum width=2.5cm, minimum height=2.6cm,text width=0.5cm](node3) at (7.65,0.7) {\textbf{Node~3}};
\node[rectangle,draw=gray,dashed, minimum width=2.5cm, minimum height=2.6cm,text width=2.4cm](cellstate) at (4.9,3.7) {\textbf{Cell\\State}};
%\node[nodeboundarystyle,text width=2.6cm,minimum height=2.5cm,text height=-1.7cm](node4) at (3,-3.1) {Node 4};
\node[opstyle](uop1) at (2.5,2.5) {\textbf{Node 3}\\add\\tanh};
\node[opstyle](uop2) at (0,2.5) {\textbf{Node 2}\\elem mult\\ReLU};
\node[opstyle](bop) at (1.25,4.5) {\textbf{Node 1}\\add\\ReLU};
\node[opstyle](cellstatep) at (1.25,0) {\textbf{Cell State}\\Input 2\\Output 3\\elem mult\\tanh};

\draw[-]   (uop1) -- (bop);
\draw[-]   (uop2) -- (bop);

\node[binaryopstyle](op1) at (7,0){add};
\node[unaryopstyle](op2) [above of=op1]{tanh};
\node[binaryopstyle](op3) [left of=op1]{elem\\mult};
\node[unaryopstyle](op4) [above of=op3]{ReLU};
\node[binaryopstyle](op5) [above of=op4]{elem\\mult};
\node[unaryopstyle](op6) [above of=op5]{tanh};
\node[binaryopstyle](op7) [right of=op6]{add};
\node[unaryopstyle](op8) [right of=op7]{ReLU};
\node[rectangle](cellstate) [left of=op4]{};
\node[rectangle](cellstateout) [right=2 of op5]{};
\node[rectangle](in1) [below of=op1]{};
\node[rectangle](in2) [below right=0.5 of op1]{};
\node[rectangle,node distance=1.2cm](out) [above of=op8]{};
\node[rectangle](in3) [below of=op3]{};
\node[rectangle](in4) [below left=0.5 of op3]{};

\draw[->]   (op1) -- (op2);
\draw[->]   (op3) -- (op4);
\draw[->]   (op4) -- (op5);
\draw[->]   (op5) -- (op6);
\draw[->]   (op6) -- (op7);
\draw[->]   (op2) -- (op7);
\draw[->]   (op7) -- (op8);
\draw[->]   (cellstate) --node[left]{$\vec{c}_{t-1}$} (op5);
\draw[->]   (op2) --node[right]{$\vec{c}_{t}$} (cellstateout);
\draw[->]   (op8) --node[right]{$\vec{h}_{t}$} (out);
\draw[->]   (in1) --node[below]{$\vec{h}_{t-1}$} (op1);
\draw[->]   (in2) --node[right]{$\vec{x}_{t}$} (op1);
\draw[->]   (in3) --node[below]{$\vec{h}_{t-1}$} (op3);
\draw[->]   (in4) --node[left]{$\vec{x}_{t}$} (op3);
\end{tikzpicture}%

  \caption{On the left is an example for a full binary tree describing a recurrent cell.
  Each node has a merging and an activation function associated.
  The cell state information is stored independently.
  It points to two nodes from the binary tree which are associated to the input and output as well as a merging and activation function.
  The right shows how this representation is translated to a computational graph.}
  \label{fig:space-rnn}
\end{figure}

\subsection{Search Space for Recurrent Cells}\label{sub:space-rnn}

Recurrent neural networks (RNNs) are the popular class of deep learning models used for sequence-to-sequence modelling tasks in natural language processing and other domains. 
\cite{Zoph2017_Neural} were the first and only to define a search space for recurrent cells.
A recurrent cell ($g$) can be described by
\begin{equation}
    \vec{h}_t = g_{\vec{\theta},\vec{\arch}}\left(\vec{x}_t, \vec{h}_{t-1}, \vec{c}_{t-1}\right)\,,
\end{equation}
where $\vec{\theta}$ is the vector of trainable parameters, $\vec{\arch}$ the structure of $g$, $\vec{h}_t$ the hidden state, $\vec{c}_t$ the cell state, and $\vec{x}_t$ the input at step $t$.
The architecture $\vec{\arch}$ is defined by a full binary tree where every node is associated with a binary merge operation (e.g. addition, element-wise multiplication) and an activation function (e.g. tanh, sigmoid) as shown in Figure~\ref{fig:space-rnn}.
Every node $i$ represents a parameterized function
\begin{equation}
    g^{(i)}_{\vec{\theta}}\left(\nodei{i_1},\nodei{i_2}\right)=\opi{\text{act}}\left(\opi{\text{merge}}\left(\nodei{i_1},\nodei{i_2}\right)\right)\,,
\end{equation}
where $\nodei{i_1}$ and $\nodei{i_2}$ are the outputs computed by the child nodes of node $i$.
Each leaf node takes as input $\vec{x}_t$ and $\vec{h}_{t-1}$, the cell state is considered independently.
A node within the tree has to be selected to provide $\vec{c}_t$.
$\vec{c}_{t-1}$ is used in combination with an arbitrary node using the selected merge operation and activation function.

\section{Optimization Methods}\label{sec:opt}

With a comprehensive coverage of the search spaces, we are now in a position to formally define the problem of neural architecture search.
We denote the space of all datasets as $\dataspace$, the space of all deep learning models as $\modelspace$, and the architecture search space as $\searchspace$.
Given this setup, a general deep learning algorithm $\DLLearner$ is defined as the mapping
\begin{equation}
    \DLLearner\ :\ \dataspace\times\searchspace\rightarrow\modelspace\ .
\end{equation}
In the context of this definition, an architecture of the search space $\vec{\arch}\in\searchspace$ defines more than just the topology.
It further encodes all properties required to train a network on a dataset which includes the choice of model parameter optimization algorithm, regularization strategies and all other hyperparameters.

Given a dataset $\D$, which is split into a training partition $\Dtrain$ and a validation partition $\Dvalid$, the general deep learning algorithm $\DLLearner$ estimates the model $\model_{\boldsymbol{\arch},\boldsymbol{\modelparams}}\in\modelspace_{\boldsymbol{\arch}}$.
This model is estimated by minimizing a loss function $\loss$ which is penalized with a regularization term $\reg$ with respect to the training data.
That is,
\begin{equation}
    \label{eq:nas_opt}
    \DLLearner\left(\boldsymbol{\arch},\D\right) = \argmin_{\model_{\boldsymbol{\arch},\boldsymbol{\modelparams}}\in\modelspace_{\boldsymbol{\arch}}}\,\loss\left(\model_{\boldsymbol{\arch},\boldsymbol{\modelparams}}, \Dtrain\right) + \reg\left(\boldsymbol{\modelparams}\right)\ .
\end{equation}
Neural architecture search is the task of finding the architecture $\vec{\archopt}$ which maximizes an objective function $\objective$ on the validation partition $\Dvalid$.
Formally,
\begin{equation}
    \vec{\archopt} = \argmax_{\boldsymbol{\arch}\in\searchspace}\, \objective\left(\DLLearner\left(\boldsymbol{\arch},\Dtrain\right),\Dvalid\right)= \argmax_{\boldsymbol{\arch}\in\searchspace} \f\left(\boldsymbol{\arch}\right)\ .
    \label{eq:opt-problem-definition}
\end{equation}
The objective function $\objective$ can be the same as the negative loss function $\loss$.
For the classification problem it is often the case that the loss is the negative cross-entropy and the objective function the classification accuracy.
This problem definition falls under the broader scope of hyperparameter optimization~\citep{Bergstra2012_Random}.

Optimizing the \emph{response function} $\f$ is a global black-box optimization problem.
In the following, we discuss several optimization strategies based on reinforcement learning, evolutionary algorithms and others.

\begin{figure}[t]
  \centering
    \tikzsetnextfilename{opt-rl-general-framework}%
    \begin{tikzpicture}[->,shorten >=1pt,thick,node distance=1cm and 3cm]
  \input{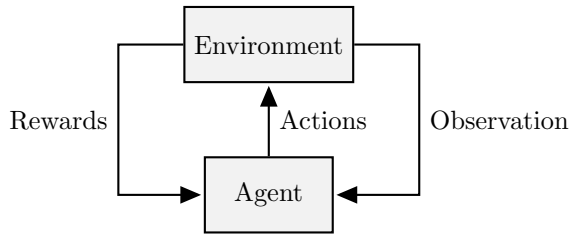}
  \tikzstyle{boxstyle}=[rectangle,minimum width=17mm,draw=black,minimum height=10mm]
  \begin{scope}
    \node [boxstyle,fill=gray!10] (controller) at (2,0) {\small Agent};
    \node [boxstyle,fill=gray!10] (environment) at (2,2)  {\small Environment};

    \draw[->]   (controller) -- node[right]{\small Actions} (environment);
    \draw[->]   (environment) to (4,2) to node [right,pos=0.5]{\small Observation} (4,0) to (controller);
    \draw[->]   (environment) to (0,2) to node [left,pos=0.5]{\small Rewards} (0,0) to (controller);
    
  \end{scope}
\end{tikzpicture}%

  \caption{A general framework for reinforcement learning algorithms.}
  \label{fig:opt-rl-general-framework}
\end{figure}
\subsection{Reinforcement Learning}\label{sub:opt-rl}
Reinforcement learning (RL) approaches are useful for modeling a process of sequential decision making where an agent interacts with an environment with the sole objective of maximizing its future return.
The agents learns to improve its behavior through multiple episodes of interaction with the environment.
At every step $t$, the agent executes an action $a^{\left(t\right)}$ and receives an observation of the state $s^{\left(t\right)}$ along with a reward $r^{\left(t\right)}$.
The environment on the other hand receives the agent's action, accordingly transits to a new state, and emits the corresponding observation and scalar reward (Figure \ref{fig:opt-rl-general-framework}).
The objective of the agent lies in maximizing its return which is expressed as a discounted sum of rewards over $T$ decision steps as defined by $\sum_{t=0}^T \gamma r^{\left(t\right)}$, where $\gamma$ is the discounting factor.
Naturally, such approaches are well suited for neural architecture search where the agent, namely the search algorithm, takes decisions to modify the system's state, i.e. the architecture, so as to maximize the long term objective of high performance which is often measured in terms of accuracy.

Reinforcement learning systems are often modeled as a \emph{Markov decision process} encompassing cases where the environment is fully observable.
Most approaches consider simplified scenarios where the set of actions and states is finite and the setup consists in a finite horizon problem where episodes terminate after a finite number of steps.
Even in this setting the number of possible options is combinatorially high.
An RL agent seeks to learn a policy $\pi$ which serves as the mapping from a state to a probability distribution over the set of actions.
While one class of RL methods indirectly learns this policy with the use of value functions and state-action value functions, other approaches directly learn a parameterized policy.
The state-value function $v_\pi(s)$ defines the expected return for the agent given that the starting state is $s$ and the policy $\pi$ is followed thereafter.
The value of taking an action $a$ in state $s$ under a policy $\pi$ is similarly defined by $q_\pi$.
Both value functions satisfy the consistency equation defined by the Bellman equation~\citep{Sutton1998_Reinforcement}.
Furthermore, the value function induces a partial order over policies and the optimal policy is the one that ascribes largest values to all states.
The goal of an agent is to find this optimal policy.

\paragraph{Temporal Difference Learning}
Approaches like SARSA~\citep{Rummery1994_online}, TD-$\lambda$~\citep{Sutton1998_Reinforcement}, and Q-learning~\citep{Watkins1989_learning} attempt to find this policy implicitly via approximating the optimal value functions.
The optimal policy is subsequently determined as the greedy-policy with respect to the optimal value function.
The optimal value functions $v^\ast(s)$ and $q^\ast(a,s)$ satisfy the Bellman optimal criterion.
Q-learning \citep{Watkins1989_learning} learns an action-value function $Q(s,a)$ that directly approximates $q^\ast$.
The agent learns $Q$ independently from the policy it follows.
The Q-learning algorithm uses the following update rule for the action-value function $Q$,
\begin{equation}
    \label{eq:q_learning}
    Q(s^{\left(t\right)},a^{\left(t\right)})  \leftarrow (1-\eta)\,Q(s^{\left(t\right)},a^{\left(t\right)}) + \eta\left[r^{(t)} + \gamma\max_{a^\prime}Q\left(s^{(t+1)},a^\prime\right)\right]\,.
\end{equation}
Here, $\eta$ is the Q-learning rate.
The learned policy and correspondingly the proposed architecture is subsequently derived by greedily selecting the actions for every state which maximize $Q(s,a)$.

\paragraph{Policy Gradient Methods}
An alternative approach in RL consists in policy gradient methods which do not appeal to value functions and instead directly learn policies as defined by a collection of parameters, $\pi_{\vec{\theta}}(a|s)$.
These methods select actions without explicitly consulting a value function.
The parameters of the policy $\vec{\theta}$ are adjusted so as to move in the direction of the agent's performance measure via classical gradient ascent updates.
The required gradient is often not directly available as the performance depends both on the actions selected and the distribution of states under which the actions is taken.
Hence an empirical estimate of the gradient is used to perform the necessary update.
REINFORCE~\citep{Williams1992_Simple} is a classical algorithm that estimates this gradient as,
\begin{equation}
    \label{eq:reinforce}
    \mathbb{E}_\pi\left[\sum_{t=0}^T \nabla_{\vec{\theta}}\ln\pi_{\vec{\theta}}(a^{(t)}|s^{(t)}) G^{(t)}\right]\,.
\end{equation}
where $G^{(t)}$ is the return from step $t$, $\left(\sum_t^T\gamma r^{(t)}\right)$.
However these empirical gradient estimates are often noisy and additional tricks like inclusion of a baseline are incorporated for useful learning. 
Moreover, alternative formulations of this gradient as derived through objective functions of importance sampling have led to other approximations.
These approaches include Trust Region Policy Optimization (TRPO)~\citep{Schulman2015_Trust} and Proximal Policy Optimization (PPO)~\citep{Schulman2017_Proximal}.
It is worth noting that these methods are on-policy methods as the agents follow the policy they seek to learn.

\paragraph{Optimization with Q-Learning}
\cite{Baker2017_Designing} were one of the first to propose the use of RL-based algorithms for neural architecture search.
They use a combination of Q-learning, $\epsilon$-greedy, and experience replay in the design of their algorithm.
The actions in their approach are the choice of different layers to add to an architecture as well as the option to terminate building the architecture and declare it as finished (we discussed this search space in Section~\ref{sub:space-global}).
Subsequently, the states are the premature architectures.
The trajectories sampled from this state space correspond to models which are subsequently trained to compute the validation accuracy.
The $Q$ function is appropriately updated with experience replay.
In order to trade off between exploration and exploitation, they incorporate an $\epsilon$-greedy strategy where random trajectories are sampled with a probability of $\epsilon$.
This is an off-policy method where the agent does not use the optimal policy during the episodes.
Additionally, the action-value function utilized in this case is deterministic.
The algorithm works as follows: after initializing the action-value function, the agent samples a trajectory which comprises of multiple decision steps, eventually leading to a terminal state.
The algorithm then trains the model corresponding to the trajectory and updates the action-value function as defined in the Q-learning algorithm.
\cite{Zhong2018_Practical} also use a Q-learning-based algorithm to search for network architectures. 
However, they perform the search in a cell-based search space which we discussed in Section~\ref{sub:space-cell}.

\begin{figure}[t]
  \centering
    \tikzsetnextfilename{opt-rl-controller}%
    \begin{tikzpicture}
\input{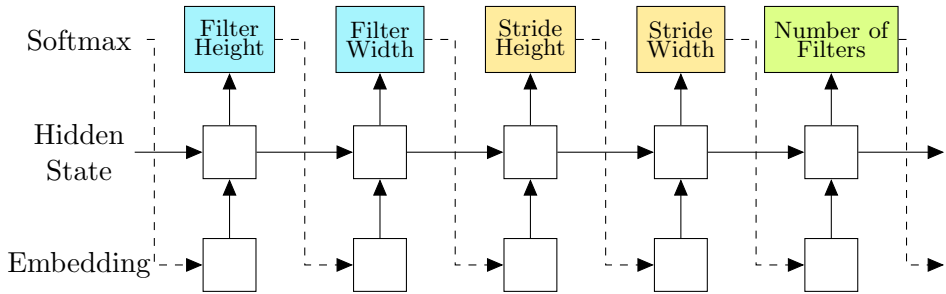}
\tikzstyle{boxstyle}=[rectangle,draw=black,minimum size=20,align=center, font=\fontsize{9}{0}\selectfont]
\tikzstyle{opstyle}=[boxstyle,fill=colorOp, minimum height=25]
\tikzstyle{unaryopstyle}=[opstyle,fill=colorSegment]
\tikzstyle{binaryopstyle}=[opstyle,fill=colorCell]

\edef\prevhiddenlabel{labelinput}
\edef\xdim{2.0}

\node[rectangle,align=center](\prevhiddenlabel) at (0,0) {Hidden\\State};
\node[rectangle,align=center](labelemb) at (0,-1.5) {Embedding};
\node[rectangle,align=center](labelsm) at (0,1.5) {Softmax};

\node[opstyle](sm1) at (\xdim,1.5) {Filter\\Height};
\node[opstyle](sm2) at (2*\xdim,1.5) {Filter\\Width};
\node[unaryopstyle](sm3) at (3*\xdim,1.5) {Stride\\Height};
\node[unaryopstyle](sm4) at (4*\xdim,1.5) {Stride\\Width};
\node[binaryopstyle](sm5) at (5*\xdim,1.5) {Number of\\Filters};

\foreach \i in {1,...,5}
{
    \xdef\hiddenlabel{hidden\i}
    
    \node[boxstyle](\hiddenlabel) at (\xdim*\i,0) {};
    \draw[->]   (\prevhiddenlabel) -- (\hiddenlabel);
    \xdef\prevhiddenlabel{\hiddenlabel}
    
    \xdef\embeddinglabel{emb\i}
    \node[boxstyle](\embeddinglabel) at (\xdim*\i,-1.5) {};
    \draw[->]   (\embeddinglabel) -- (\hiddenlabel);
    
    \draw[->]   (\hiddenlabel) -- (sm\i);
}
\foreach \i in {1,...,4}
{
    \pgfmathtruncatemacro\result{\i + 1}
    \draw[->,dashed]   (sm\i) to (\xdim*\i+\xdim/2,1.5) to (\xdim*\i+\xdim/2,-1.5) to (emb\result);
}

\draw[->,dashed]   (0.9,1.5) to (\xdim/2,1.5) to (\xdim/2,-1.5) to (emb1);
\draw[->,dashed]   (sm5) to (\xdim*5.5,1.5) to (\xdim*5.5,-1.5)  to (\xdim*5.75,-1.5);
\draw[->]   (hidden5) to (\xdim*5.75,0);
\end{tikzpicture}%

  \caption{The controller used by \cite{Zoph2017_Neural} (predictions for skip connections are not shown) to predict configuration of one layer.}
  \label{fig:opt-rl-controller}
\end{figure}
\paragraph{Optimization with Policy Gradient Methods}
Alternate approaches based on policy gradient method have also been used for neural architecture search.
The work by \cite{Zoph2017_Neural} was the first to consider this modeling approach.
In their approach they directly model a controller whose predictions can be considered as actions that are used to construct a neural architecture.
The controller parameterized by $\vec{\theta}$ defines the stochastic policy $\pi_{\vec{\theta}}(a|s)$.
They incorporate an autoregressive controller which predicts the action based on previous actions, and model it with a Recurrent Neural Network (RNN).
During an episode, every action is sampled from the probability distribution implied by a softmax operation and then fed into the next time step as input (Figure~\ref{fig:opt-rl-controller}).
The RNN-controller in their approach samples layers which are sequentially appended to construct the final network.
The final network is trained to obtain the validation accuracy and the parameters of the controller are updated as per the REINFORCE update rule (Equation~\eqref{eq:reinforce}) which involves scaling the gradient with the obtained validation accuracy.
As discussed in Section~\ref{sub:space-global}, they consider a space of chain-structured networks parameterized by only convolutional layers with different filter width, filter height, stride parameters, and number of filters.
Additionally, they include anchor points in their encoding to allow for skip connections.

Policy gradient approaches have also been explored to learn controllers for cell-based search spaces.
\cite{Zoph2018_Learning} define an RNN-controller that outputs actions which sequentially determines the inputs and operations for a prespecified number of blocks within a NASNet search space cell as defined in Section \ref{sub:space-cell}.
The sampled architectures are trained and the parameters of the controller are updated according to the PPO update rule.

\cite{Cai2018_Efficient} also model a controller to parameterized the policy and train it with REINFORCE.
However the action space of the controller differs and it requires an initial architecture to start with.
There are two types of actions, widening a layer or inserting a new layer to deepen the network.
Both actions are performed by means of \emph{function-preserving transformations}~\citep{Chen2016_Net2Net}.
This means that although the structure of the architecture changes, the function that it models remains unchanged.
Thus, the accuracy of this network does not change at first.
This is ensured by initializing the additionally added parameters in a special way.
In the example of adding a layer, it is initialized to correspond to the identity function.
At every step and every layer, the controller predicts whether it wants to change it by means of widening or deepening it.
As this involves predicting transformation decisions for a variable length sequence, they incorporate a bidirectional recurrent network~\citep{Schuster1997_Bidirectional} with an embedding layer to learn a low-dimensional representation of the network which is subsequently fed into actors that compute the transformation decisions.
For the predicted actions the function-preserving transformations are computed which enable efficient reuse of weights for training the newly sampled architecture which reduces the overall search duration.

The work by \cite{Cai2018_Path} builds upon these transformations and proposes new transformations to include branches in the network architectures.
A convolutional layer or an identity map is first converted to its equivalent multi-branch motif such that the functionality is preserved.
This is achieved by splitting or replicating the input to a convolutional or identity map into multiple branches, applying the corresponding transformation (convolutional or identity) to each branch, and respectively concatenating or adding the outputs of each branch.
The function-preserving transformations are subsequently applied on branches to explore more complex architectures.
Similar to the work by \cite{Cai2018_Efficient}, the meta-controller in their approach also uses an encoder network to learn a low-dimensional representation of the architecture and provide a distribution over transformation actions.
They propose to learn the meta-controller with REINFORCE.

\paragraph{Monte Carlo Tree Search} 
An alternative way of solving the reinforcement learning problem lies in the use of Monte Carlo methods which are based on averaging sample returns~\citep{Sutton1998_Reinforcement}.
A popular algorithm in this domain is the UCT algorithm~\citep{Kocsis2006_Bandit} which has been adopted in different ways for neural architecture search.
This algorithm is designed for a tree-structured state-action space which is incrementally explored and expanded. 
A policy based on upper confidence bound~\citep{Auer2002_Using} is utilized to explore this tree.
The work of \cite{Negrinho2017_DeepArchitect} adopts the vanilla UCT algorithm with bisection for neural architecture search.
\cite{Wistuba2018_Practical} builds on this work and employs function-preserving operations to speed up the training process and uses surrogate models to predict the reward for unexplored states.
Similarly, \cite{Wang2019_AlphaX} make use of a surrogate model to predict the state reward and search on the NASNet search space and use the weight sharing paradigm (Section~\ref{sub:opt-one-shot}).

\subsection{Evolutionary Algorithms}\label{sub:opt-ea}

\begin{figure}[t]
  \centering
    \tikzsetnextfilename{opt-ea-general-framework}%
    \begin{tikzpicture}[->,shorten >=1pt,thick,node distance=1cm and 3cm]
  \input{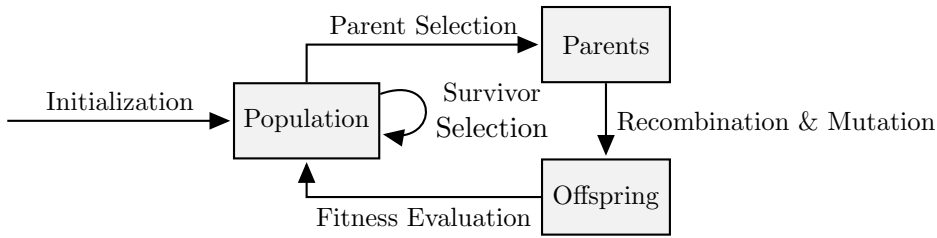}
  \tikzstyle{boxstyle}=[rectangle,minimum width=17mm,draw=black,minimum height=10mm]
  \begin{scope}
    \node [boxstyle,fill=gray!10] (parents)   {\small Parents};
    \node [boxstyle,fill=gray!10] (offspring) [below=of parents]  {\small Offspring};
    \path (parents) -- (offspring) coordinate[midway] (aux);
    \node [boxstyle, left=of aux,fill=gray!10] (population) {\small Population};
    \node [left=of population] (init_aux) {};
    \draw[->]   (population) |- node[above,pos=0.75]{\small Parent Selection} (parents);
    \draw[->]   (offspring) -| node[below,pos=0.25]{\small Fitness Evaluation} (population);
    \draw[->]   (parents) -- node[right]{\small Recombination \& Mutation} (offspring);
    \draw[->]   (init_aux) -- node[above]{\small Initialization} (population);
    \path (population) edge [out=380,in=350,looseness=4] node[right,align=center] {\small Survivor\\Selection} (population);
  \end{scope}
\end{tikzpicture}%

  \caption{A general framework for evolutionary algorithms.}
  \label{fig:opt-ea-general-framework}
\end{figure}

Evolutionary algorithms (EA) are population-based global optimizer for black-box functions which consist of following essential components: initialization, parent selection, recombination and mutation, survivor selection~\citep{DeJong2006_Evolutionary}.
The initialization defines how the first generation of the population is generated.
After the initialization the optimizer repeats the following steps until termination (see Figure \ref{fig:opt-ea-general-framework}):
\begin{enumerate}
    \item Select parents from the population for reproduction.
    \item Apply recombination and mutation operations to create new individuals.
    \item Evaluate the fitness of the new individuals.
    \item Select the survivors of the population.
\end{enumerate}

In this broad class of algorithms, the choices for mutation and recombination operators along with the choice of fitness and parent selection functions guide the overall search process.
The choice of operators used for recombination and mutation is motivated to trade off diversity and similarity in the population, akin to the exploration and exploitation trade-off in reinforcement learning-based search algorithms.
Similarly, the choice of fitness functions reflects the optimization objective and the choice of survivor selection enables competition between the individuals of the population.

In the context of neural architecture search, the population consists of a pool of network architectures.
A parent architecture or a pair of architectures is selected in step 1 for mutation or recombination, respectively.
The steps of mutation and recombination refer to operations that lead to novel architectures in the search space which are evaluated for fitness in step 3 and the process is repeated till termination.
Often only mutation operators are used in the domain of neural architecture search.
There is no indication that a recombination operation applied to two individuals with high fitness would result into an offspring with similar or better fitness.
On the contrary, oftentimes the fitness of the resultant offspring is much lower.
For this reason and other reasons of simplicity, this part is often omitted.

The most common parent selection method in neural architecture search is \emph{tournament selection} \citep{Goldberg1990_Tournament_Selection}.
This method selects the individuals for further evolution in two steps.
First, it selects $k$ individuals from the population at random.
Then it iterates over them in the descending order of their fitness while selecting individuals for further steps with some (prespecified) high probability $p$.
An alternative to tournament selection is \emph{fitness proportionate selection}.
In this approach an individual is selected proportional to its fitness.
Thus, in a population $\left\{\arch_{1},\ldots,\arch_{N}\right\}$, the $i$-th individual is selected with probability $\frac{\f\left(\arch_i\right)}{\sum_{j=1}^{N}\f\left(\arch_j\right)}$, where $\f\left(\arch_i\right)$ is the fitness of the individual $\arch_{i}$.

\begin{figure}[t]
  \centering
    \tikzsetnextfilename{opt-ea-genotype-phenotype}%
    \begin{tikzpicture}
  \input{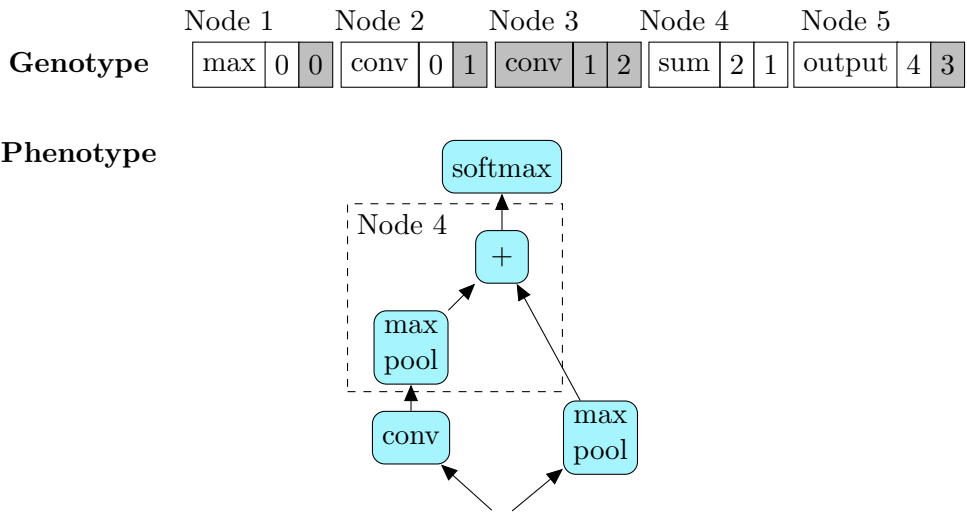}
  \tikzstyle{nodestyle}=[rectangle,minimum height=6mm,draw=black,fill=white]
  \tikzstyle{inactivenodestyle}=[nodestyle,fill=lightgray]
  \tikzstyle{activenodestyle}=[nodestyle,fill=white]
  \tikzstyle{nodeboundarystyle}=[rectangle,draw=black,dashed]
  
  \node[activenodestyle](n11) {max};
  \node[activenodestyle,node distance=7mm](n12) [right of=n11] {0};
  \node[inactivenodestyle,node distance=4.5mm](n13) [right of=n12] {0};
  \node[rectangle,node distance=6mm](l1) [above of=n11] {Node 1};
  
  \node[activenodestyle,node distance=8.5mm](n21) [right of=n13] {conv};
  \node[activenodestyle,node distance=7.5mm](n22) [right of=n21] {0};
  \node[inactivenodestyle,node distance=4.5mm](n23) [right of=n22] {1};
  \node[rectangle,node distance=6mm](l2) [above of=n21] {Node 2};
  
  \node[inactivenodestyle,node distance=8.5mm](n31) [right of=n23] {conv};
  \node[inactivenodestyle,node distance=7.5mm](n32) [right of=n31] {1};
  \node[inactivenodestyle,node distance=4.5mm](n33) [right of=n32] {2};
  \node[rectangle,node distance=6mm](l3) [above of=n31] {Node 3};
  
  \node[activenodestyle,node distance=8mm](n41) [right of=n33] {sum};
  \node[activenodestyle,node distance=7mm](n42) [right of=n41] {2};
  \node[activenodestyle,node distance=4.5mm](n43) [right of=n42] {1};
  \node[rectangle,node distance=6mm](l4) [above of=n41] {Node 4};
  
  \node[activenodestyle,node distance=10mm](n51) [right of=n43] {output};
  \node[activenodestyle,node distance=9mm](n52) [right of=n51] {4};
  \node[inactivenodestyle,node distance=4.5mm](n53) [right of=n52] {3};
  \node[rectangle,node distance=6mm](l5) [above of=n51] {Node 5};
  
  \node[rectangle,node distance=2cm](genotype) [left of=n11] {\textbf{Genotype}};
  
  \node[opstyle](softmax) [below=1 of $(n23)!0.5!(n31)$]{softmax};
  \node[opstyle](add) [below of=softmax]{$+$};
  \node[opstyle](max1) [below left=0.5 of add]{max\\pool};
  \node[opstyle](conv1) [below of=max1]{conv};
  \node[opstyle](max2) [right=1.5 of conv1]{max\\pool};
  \node[rectangle](input) [below =3 of add]{};
  
  \node[nodeboundarystyle,text width=2.6cm,minimum height=2.5cm,text height=-1.7cm](node4) at (3,-3.1) {Node 4};
  
  \node[rectangle,node distance=1.2cm](phenotype) [below of=genotype] {\textbf{Phenotype}};
  
  \draw[->]   (input) -- (conv1);
  \draw[->]   (input) -- (max2);
  \draw[->]   (conv1) -- (max1);
  \draw[->]   (max2) -- (add);
  \draw[->]   (max1) -- (add);
  \draw[->]   (add) -- (softmax);
\end{tikzpicture}%

  \caption{The genetic encoding used by \cite{Suganuma2017_A}.
  Gray shaded parts of the genotype are inactive.
  Inactive parts such as Node 3 are not present in the phenotype.
  The additional max pooling layer introduced by Node 4 is an example how not explicitly encoded operations in the genotype can appear in the phenotype.}
  \label{fig:opt-ea-phenotype-genotype}
\end{figure}

After the recombination and mutation step, the population has grown.
The intent of the survivor selection component is to reduce the population size and enable competition within the individuals.
Several different policies are used to achieve this, ranging from selecting only the best (\emph{elitist selection}) to selecting all individuals.
One class of evolutionary algorithm that has been widely adopted for neural architecture search is genetic algorithms.
These approaches bear their name from the representation of individuals in their methodology which is done by means of a fixed-length encoding, called the \emph{genotype}.
Mutation and recombination operations are performed directly on this representation.
Further, a genotype is used to create the physical appearance of the individual, which in the case of neural architecture search is the architecture.
This materialization is called the \emph{phenotype}.
It is important to note that parts of the genotype can be conditionally active or inactive.
The information of the genotype is referred to as \emph{active} if its modification results in a change in the phenotype (given all other information remains the same).
For an example see Figure \ref{fig:opt-ea-phenotype-genotype}.

\begin{table}
  \caption{High-level details of various evolutionary algorithms for neural architecture search.}
  \label{tab:opt-ea-overview}
  \centering
  \begin{tabular}{lcccc}
    \hline\noalign{\smallskip}
    Method & Search Space & Init & Parent Sel. & Survivor Sel. \\
    \noalign{\smallskip}
    \hline
    \noalign{\smallskip}
    \cite{Real2017_Large} & global & simple & tournament & tournament \\
    \cite{Xie2017_Genetic} & global & random & all & fitness prop. \\
    \cite{Suganuma2017_A} & global & random & n/a & elitist \\
    \cite{Liu2018_Hierarchical} & cell-based & random & tournament & all \\
    \cite{Real2019_Aging} & cell-based & random & tournament & youngest \\
    \cite{Elsken2018_Simple} & global & simple & n/a & elitist \\
    \cite{Wistuba2018_Deep} & cell-based & simple & tournament & all \\
    \hline
  \end{tabular}
\end{table}
In the remaining section we discuss the popular choices for search space, mutation operators and selection functions that have been utilized for neural architecture search.
We note that EA-based neural architecture search methods include a set of highly diverse approaches which have benefited from the varied choices of encoding the search space along with the choices of mutation operators and selection functions.
In the context of this work we describe six notable works in EA-based neural architecture search.
A broad overview of these approaches is provided in Table \ref{tab:opt-ea-overview}.

\begin{figure}[t]
  \centering
    \tikzsetnextfilename{opt-ea-real2017-architecture}%
    \begin{tikzpicture}
  \input{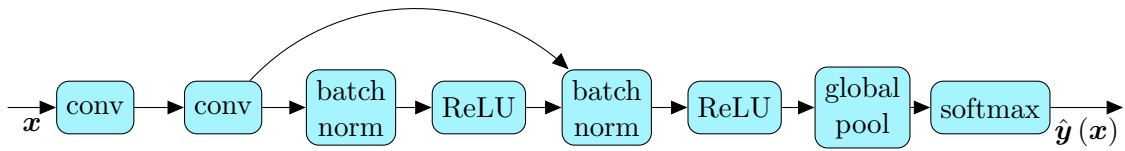}
  \tikzstyle{opstyle2}=[opstyle,node distance=1.7cm]
  
  \node[boxstyle](input) {};
  \node[opstyle2,node distance=1.3cm](conv1) [right of=input] {conv};
  \node[opstyle2](conv2) [right of=conv1] {conv};
  \node[opstyle2](bn1) [right of=conv2] {batch\\norm};
  \node[opstyle2](a1) [right of=bn1] {ReLU};
  \node[opstyle2](bn2) [right of=a1] {batch\\norm};
  \node[opstyle2](a2) [right of=bn2] {ReLU};
  \node[opstyle2](globalpool) [right of=a2] {global\\pool};
  \node[opstyle2](softmax) [right of=globalpool] {softmax};
  \node[boxstyle,node distance=1.9cm](output) [right of=softmax] {};
  
  \draw[->]   (input) -- node[below]{$\vec{x}$} (conv1);
  \draw[->]   (conv1) -- (conv2);
  \draw[->]   (conv2) -- (bn1);
  \draw[->]   (bn1) -- (a1);
  \draw[->]   (a1) -- (bn2);
  \draw[->]   (bn2) -- (a2);
  \draw[->]   (a2) -- (globalpool);
  \draw[->]   (globalpool) -- (softmax);
  \draw[->]   (softmax) -- node[below]{$\hat{\vec{y}}\left(\vec{x}\right)$} (output);
  
  \draw[->]   (conv2) to [bend left=45] (bn2);
\end{tikzpicture}%

  \caption{A possible architecture discovered by the algorithm described by \cite{Real2017_Large}.
  Noticeable are the redundant operations such as two convolutions in a row without an activation.
  }
  \label{fig:opt-ea-real2017-architecture}
\end{figure}
\paragraph{\cite{Real2017_Large}}
\cite{Real2017_Large} were one of the first to propose an evolutionary algorithm to find competitive convolutional neural network architectures for image classification.
Their approach begins with one thousand copies of the simplest possible architecture, i.e. a global pooling layer followed by the output layer.
In the parent selection step of their approach they propose to sample a pair of architectures for further processing.
While the better of the two architectures is copied (along with the weights), mutated, trained for 25,600 steps with a batch size 50, and added to the population, the other one is removed from the population.
This is equivalent to using tournament selection with $k=2$ and $p=1$.
The set of mutations consists of simple operations such as adding convolutions (possibly with batch normalization or ReLU activation) at arbitrary locations before the global pooling layer, along with mutations that alter kernel size, number of channels, stride and learning rate, add or remove skip connections, remove convolutions, reset weights, and a mutation that corresponds to no change.
Since no further domain constraints are enforced, architectures with redundant components can be potentially sampled in this approach (see Figure \ref{fig:opt-ea-real2017-architecture}).

\paragraph{\cite{Xie2017_Genetic}}
\cite{Xie2017_Genetic} made their work available only one day later.
In contrast to \cite{Real2017_Large}, their work develops a genetic algorithm over a more structured search space which comprises of a sequence of segments with possibly parallel convolution operations.
In their approach they consider networks to be composed of three segments, each consisting of multiple convolution layers (see Section \ref{sub:space-global} for more details).
Each segment is described by an adjacency matrix which defines a directed acyclic graph over the layers within the segment.
The genotype of the entire network is obtained as a fixed length binary string of the adjacency matrices of the three segments.
The algorithm begins with a population of 20 random samples from the search space.
In the parent selection step, all pairs $\left(\arch_i,\arch_{i+1}\right),\ i \mod 2 = 1$ are considered for a cross-over operation.
With a probability $p$, the cross-over operation swaps segments between the two selected networks.
As the next step, all individuals which have undergone no modifications in the previous step are considered for mutation.
In their algorithm, mutations are defined as random flip operations on the adjacency matrices that define the segments.
Finally, the obtained offspring is trained from scratch and its fitness is evaluated.
The fitness of an individual is defined as the difference between its validation accuracy and the minimum accuracy among all individuals of the population.
As fitness proportionate selection is used in their approach, this modification ensures that the weakest individual survives with a probability of zero.
It is worth remarking that this is one of the first works to demonstrate a successful transfer of an architecture automatically discovered on a smaller dataset to a larger one, namely from CIFAR-10 to ImageNet~\citep{Russakovsky2015_ImageNet}.

\paragraph{\cite{Suganuma2017_A}}
\cite{Suganuma2017_A} present another optimizer based on genetic algorithms but consider a wider set of operations in their search space, which includes convolutions and pooling along with concatenation and summation of vectors.
In their approach, they encode the entire network as a sequence of blocks represented by a triple that defines the operation and inputs of the block. 
The string obtained by concatenating the block encoding constitutes the genotype.
Inactive parts are admissible in such a definition of a genotype, for instance unused inputs for certain operations or disconnected blocks.
The inactive parts do not materialize in the phenotype (for an example see Figure~\ref{fig:opt-ea-phenotype-genotype}).
They propose to use the (1+$\lambda$)-evolutionary strategy ($\lambda=2$)~\citep{Beyer2002_Evolution} approach to guide the evolution where mutation and selection operators are applied in a loop until termination.
Beginning with a genotype that corresponds to the parent individual (random for the first iteration), $\lambda$ offspring are generated by forcefully modifying the active parts of the genotype.
A mutation is also applied to the parent on one of the inactive parts of its genotype.
However this does not result in any change of the phenotype and hence does not add to the overall training cost.
This action ensures progress of the search even in cases where the parent of this generation will become the parent of the next generations again.
The offspring networks are trained to obtain validation accuracies and the parent for the next iteration is selected using elitist selection strategy.

\paragraph{\cite{Liu2018_Hierarchical}}
\cite{Liu2018_Hierarchical} propose another evolutionary algorithm which uses the hierarchically organized search space as described in Section~\ref{sub:space-cell}.
At every step a mutation selects the hierarchy level to modify along with the modification to the representations at that level. 
The population is initialized with 200 trivial genotypes which are diversified by applying 1000 random mutations.
Furthermore, parents are selected using tournament selection (5\% of the population) and no individuals are removed during the evolutionary process.
In their setting, mutations can add, alter and remove operations from the genotype.

\paragraph{\cite{Real2019_Aging}}
The follow-up work by \cite{Real2019_Aging} is one of the most significant works in the direction of using evolutionary algorithms for architecture search.
It is primarily known for finding the architectures AmoebaNet-B and AmoebaNet-C which set new records for the task of image classification on CIFAR-10 and ImageNet dataset.
However, their search process used a total of 3,150 GPU hours.
Their evolutionary search algorithm operates on the NASNet search space (details in Section \ref{sub:space-cell}) with tournament selection used to select parents at each iteration.
The selected individuals are mutated with a random change in an operation or a connection in either a normal or a reduction cell and are trained for 25 epochs.
As opposed to previous works, their algorithm does not solely rely on validation accuracy and incorporates age in the selection of survivors.
This is motivated to restrain repeated selection of well-performing models for mutation and introduce diversity to the population.
This basically adds a regularization term to the objective function which makes sure that we are searching for architectures which are not only capable of reaching high validation accuracy once but every time.
\vspace{0.5cm}
\\The previously discussed evolutionary algorithms are mostly concerned about finding a good performing architecture ignoring GPU budget limitations.
The shortest search time used among those methods is still seventeen GPU days.
The two works discussed in the following are more concerned about efficiency and report comparable results within a day.
The method to accomplish this is a combination of mutations which are function-preserving transformations (see Section \ref{sub:opt-rl}) and a more aggressive, greedy evolution.

\paragraph{\cite{Elsken2018_Simple}}
\cite{Elsken2018_Simple} propose a simple yet efficient evolutionary algorithm inspired by the function-preserving transformations.
Similar to \cite{Suganuma2017_A} they follow the (1+$\lambda$) evolutionary strategy.
A parent is selected with elitist selection strategy, eight different offspring are generated with function-preserving mutations, and each is trained for a total of 17 epochs.
The initial architecture comprises of three convolution and two max pooling layers.
As discussed earlier, the function-preserving operations significantly reduce the training time per architecture and therefore the entire search duration.

\begin{figure}[t]
  \centering
    \tikzsetnextfilename{opt-ea-wistuba2018-cell}%
    \begin{tikzpicture}
  \input{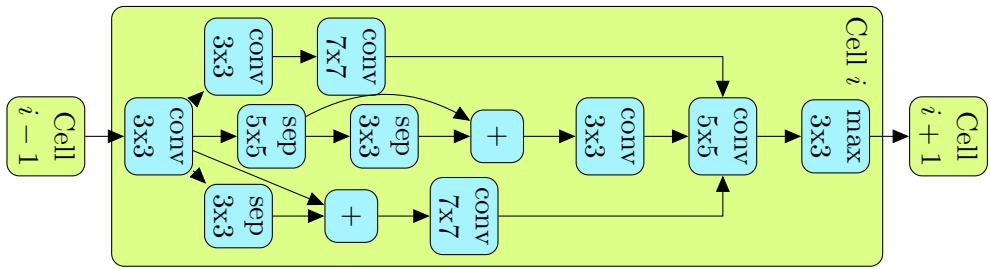}
  \tikzstyle{oprotatedstyle}=[opstyle,rotate=-90,node distance=1.5cm]
  \tikzstyle{cellrotatedstyle}=[cellstyle,rotate=-90,node distance=1.5cm]
  
  % Background
  \node[cellrotatedstyle, minimum width=3.2cm, minimum height=10.25cm,align=left,text width=3.2cm, text height=-9.3cm](bg) at (6, 0) {Cell $i$};
  
  % Input node
  \node[cellrotatedstyle](input) {Cell\\$i-1$};
  \node[oprotatedstyle](op1) [above of=input] {conv\\3x3};
  % Branch 1
  \node[oprotatedstyle](op2) [above left of=op1] {conv\\3x3};
  \node[oprotatedstyle](op3) [above of=op2] {conv\\7x7};
  % Branch 2
  \node[oprotatedstyle](op4) [above of=op1] {sep\\5x5};
  \node[oprotatedstyle](op5) [above of=op4] {sep\\3x3};
  \node[oprotatedstyle](add1) [above of=op5] {$+$};
  \node[oprotatedstyle](op6) [above of=add1] {conv\\3x3};
  % Branch 3
  \node[oprotatedstyle](op7) [above right of=op1] {sep\\3x3};
  \node[oprotatedstyle](add2) [above of=op7] {$+$};
  \node[oprotatedstyle](op8) [above of=add2] {conv\\7x7};
  
  \node[oprotatedstyle](op9) [above of=op6] {conv\\5x5};
  
  \node[oprotatedstyle](max) [above of=op9] {max\\3x3};
  
  \node[cellrotatedstyle](output) [above of=max] {Cell\\$i+1$};
  
  % Edges
  \draw[->]   (input) -- (op1);
  \draw[->]   (op1) -- (op2);
  \draw[->]   (op1) -- (op4);
  \draw[->]   (op1) -- (op7);
  
  \draw[->]   (op2) -- (op3);
  \draw[->]   (op3) -| (op9);
  
  \draw[->]   (op4) -- (op5);
  \draw[->]   (op5) -- (add1);
  \draw[->]   (add1) -- (op6);
  \draw[->]   (op6) -- (op9);
  
  \draw[->]   (op7) -- (add2);
  \draw[->]   (add2) -- (op8);
  \draw[->]   (op8) -| (op9);
  
  \draw[->]   (op9) -- (max);
  \draw[->]   (max) -- (output);
  
  \draw[->]   (op4) to [bend left=30] (add1);
  \draw[->]   (op1) -- (add2);
\end{tikzpicture}%

  \caption{Exemplary outcome of the evolutionary algorithm proposed by \cite{Wistuba2018_Deep}.}
  \label{fig:opt-ea-wistuba2018-cell}
\end{figure}

\paragraph{\cite{Wistuba2018_Deep}}
In a similar line of work, \cite{Wistuba2018_Deep} also utilizes function-preserving mutations but uses a cell-based search space.
The proposed search method begins with a simple network and performs function-preserving mutations to grow the population.
The initial population is diversified by generating fifteen additional children from the base network.
During the evolutionary algorithm, parents are selected using tournament selection and the offspring is trained for fifteen epochs prior to their fitness evaluation. 
Similar to \cite{Liu2018_Hierarchical}, all individuals survive the evolution process to ensure diversity.
Discovered cells have arbitrary structures, an example is provided in Figure \ref{fig:opt-ea-wistuba2018-cell}.

\subsection{Surrogate Model-Based Optimization}  \label{sub:opt-smbo} 

As the name implies, surrogate model-based optimizers use a surrogate model $\hat{\f}$ to approximate the response function $\f$ (Equation \eqref{eq:nas_opt})~\citep{Jones2001_A}.
For the case of neural architecture search this is motivated to get an approximate response for an architecture by circumventing the time-consuming training step and thereby improve the overall efficiency of the search process.
The surrogate itself is modeled as a machine learning model and is trained on a meta-dataset which contains architecture descriptions along with their response values, which are gathered during the architecture search:
\begin{equation}
    \obshist = \left\{\left(\vec{\arch}_1,\f\left(\vec{\arch}_1\right)\right),\left(\vec{\arch}_2,\f\left(\vec{\arch}_2\right)\right),\ldots\right\}
\end{equation}
Surrogate models are generally trained to minimize the squared error:
\begin{equation}
    \label{eq:surrogate-loss}
    \sum_{\left(\vec{\arch},\f\left(\vec{\arch}\right)\right) \in \obshist} \left(\hat{\f}\left(\vec{\arch}\right) - \f\left(\vec{\arch}\right)\right)^{2}\,.
\end{equation}
However, often in practice only a ranking for the architectures is desired and in such cases a low loss value is not necessitated as long as the surrogates provide a useful ranking.

\begin{figure}[t]
  \centering
    \tikzsetnextfilename{opt-smbo-general-framework}%
    \begin{tikzpicture}[->,shorten >=1pt,thick,node distance=1cm and 4cm]
  \input{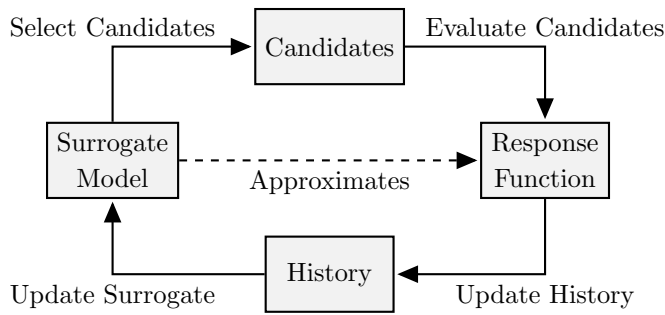}
  \tikzstyle{boxstyle}=[rectangle,minimum width=17mm,draw=black,minimum height=10mm]
  \begin{scope}
    \node [boxstyle,fill=gray!10,align=center] (surrogate) {\small Surrogate\\\small Model};
    \node [boxstyle,right=of surrogate,fill=gray!10,align=center] (response) {\small Response\\\small Function};
    \path (surrogate) -- (response) coordinate[midway] (aux);
    \node [boxstyle,above=of aux,fill=gray!10] (candidates)   {\small Candidates};
    \node [boxstyle,below=of aux,fill=gray!10] (history)  {\small History};
    
    \draw[->]   (surrogate) |- node[above,pos=0.5]{\small Select Candidates} (candidates);
    \draw[->]   (candidates) -| node[above,pos=0.5]{\small Evaluate Candidates} (response);
    \draw[->]   (response) |- node[below,pos=0.5]{\small Update History} (history);
    \draw[->]   (history) -| node[below,pos=0.5]{\small Update Surrogate} (surrogate);
    \draw[dashed,->]   (surrogate) -- node[below]{\small Approximates} (response);
  \end{scope}
\end{tikzpicture}%

  \caption{A general framework for surrogate model-based optimization.}
  \label{fig:opt-smbo-general-framework}
\end{figure}
The predictions from surrogate model are often used to identify promising candidate architectures.
These candidates are evaluated, their corresponding new meta-instances are added to $\obshist$ and the surrogate model is accordingly updated (Figure~\ref{fig:opt-smbo-general-framework}).
These steps are executed until a convergence criterion is reached.
We describe three different approaches to surrogate model-based optimization that have been used with an intent to improve the efficiency of neural architecture search.

\paragraph{\cite{Kandasamy2018_Neural}}
\cite{Kandasamy2018_Neural} cast this as a black-box optimization problem and tackle it with Bayesian optimization~\citep{Snoek2012_Practical}.
Bayesian optimization methods use a combination of a probabilistic surrogate model and an acquisition function to obtain suitable candidates.
The acquisition function measures the utility by accounting for both, the predicted response and the uncertainty in the prediction.
In their approach they model the surrogate with a Gaussian process~\citep{Rasmussen2006_Gaussian} and use expected improvement as the acquisition function~\citep{Mockus1978} for the optimization.
The architecture with the highest expected utility is selected for evaluation and the consequent meta-instances are added to $\obshist$.
The surrogate model is updated and the previous steps are repeated.
The authors render two changes to the standard Bayesian optimization algorithm.
First, a novel kernel function to compute similarity between two network architectures is proposed and second, an evolutionary algorithm is used to maximize the acquisition function.
The kernel value computation in this work is modeled as an optimal transport program~\citep{Villani2008_Optimal}.
In essence, the graph topology, the location and frequency of operations as well as the number of feature maps are used to compute the similarity between two architectures.

\paragraph{\cite{Liu2018_Progressive}}
\cite{Liu2018_Progressive} also incorporate a surrogate model and search for architectures in the NASNet search space (Section~\ref{sub:space-cell}).
They explore this search space by progressively increasing the number of blocks in a cell, which, as previously discussed, is a hyperparameter and fixed in the template. 
While searching for models with an additional block, they simultaneously train a surrogate model which can predict the validation accuracy of the architecture given the encoding of the cell.
The overall search process is designed as a beam search where at every step, a new set of candidates is generated by expanding the current pool of cells with an additional block and the top $k$ of those are trained as per the predictions of the surrogate model.
Further, in each step the surrogate model is updated by considering the newly obtained $k$ architectures in the meta-dataset.
Given the design of their setup, the surrogate model is required to handle variable sized inputs.
In addition to an RNN and an Long Short Term Memory (LSTM)~\citep{Hochreiter1997_LSTM} which are natural choices for this modeling, they also investigate a multi-layer perceptron (MLP) for this model.

\begin{figure}[t]
    \centering
    \tikzsetnextfilename{opt-smbo-nao-autoencoder}%
    \begin{tikzpicture}
\input{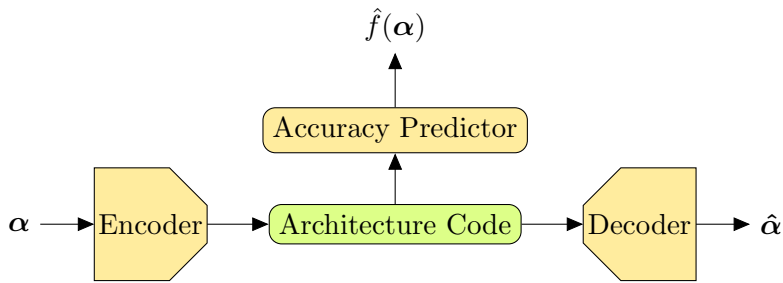}

\node[rectangle](input) at (-1,-0.75){$\vec{\arch}$};
\node[rectangle](output) at (9,-0.75){$\vec{\hat{\arch}}$};
\draw[fill=colorSegment] (0, 0) -- (1,0) -- (1.5,-0.5) -- (1.5,-1) -- (1,-1.5) -- (0,-1.5) -- cycle;
\node[rectangle] at (0.75,-0.75){Encoder};

\node[cellstyle](archcode) at (4,-0.75){Architecture Code};
\node[segmentstyle](dense) at (4,0.5){Accuracy Predictor};
\node[rectangle](surrogate) at (4,1.9){$\hat{\f}(\vec{\arch})$};

\draw[fill=colorSegment] (8, 0) -- (7,0) -- (6.5,-0.5) -- (6.5,-1) -- (7,-1.5) -- (8,-1.5) -- cycle;
\node[rectangle] at (7.25,-0.75){Decoder};

\draw[->]   (1.5,-0.75) -- (archcode);
\draw[->]   (archcode) -- (6.5,-0.75);
\draw[->]   (dense) -- (surrogate);
\draw[->]   (input) -- (0,-0.75);
\draw[->]   (archcode) -- (dense);
\draw[->]   (8,-0.75) -- (output);
\end{tikzpicture}%

    \caption{\cite{Luo2018_NAO} propose to combine an autoencoder with the surrogate model.
    This model is jointly learned to achieve $\vec{\arch}\approx\vec{\hat{\arch}}$ and $\f(\vec{\arch})\approx\hat{\f}(\vec{\arch})$.}
    \label{fig:opt-smbo-nao-autoencoder}
\end{figure}
\paragraph{\cite{Luo2018_NAO}}
In an interesting approach, \cite{Luo2018_NAO} jointly learn an autoencoder for the architecture representation with the surrogate model which uses the continuous encoding provided by the autoencoder, the architecture code, as its input (Figure~\ref{fig:opt-smbo-nao-autoencoder}).
A key difference lies in their search algorithm which uses the surrogate model to sample new architectures by taking gradient steps with respect to the architecture code.
The architecture code is changed by moving in the direction which yields better accuracy according to the surrogate model.
The new architecture is obtained by mapping the potentially better architecture code back using the decoder learned as part of the autoencoder.
At every step, these samples are trained, the meta-dataset is expanded, and the surrogate model and autoencoder are updated accordingly.
The encoder, decoder and the surrogate model are jointly trained to minimize a weighted sum of the reconstruction loss and the performance prediction loss.

\subsection{One-Shot Architecture Search}\label{sub:opt-one-shot}
We define an architecture search method as one-shot if it trains a single neural network during the search process.\footnote{Not to be confused with the one-shot learning paradigm which refers to algorithms that seek to learn from one instance.}
This neural network is then used to derive architectures throughout the search space as candidate solutions to the optimization problem.
Most of these one-shot architectures are based on an over-parameterized network~\citep{Saxena2016_Convolutional,Pham2018_ENAS,Liu2018_DARTS,Xie2019_SNAS,Casale2019_ProbNAS}.
The advantage of this family of methods is the relatively low search effort which is only slightly greater than the training costs of one architecture in the search space.
As we describe later, this methodology can be conveniently combined with many of the previously discussed optimization methods.
\begin{figure}[t]
    \centering
    \tikzsetnextfilename{opt-one-shot-weight-sharing}%
    \begin{tikzpicture}
\input{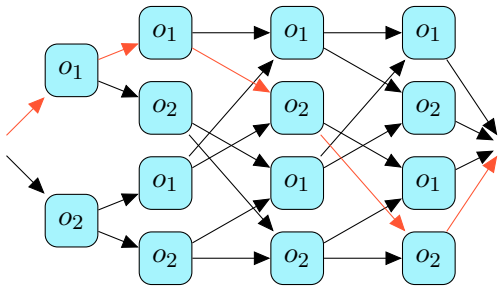}

\definecolor{red}{HTML}{ff5533}

\node[boxstyle](input) at (0,0) {};
\node[opstyle](op1-1) at (1,1) {$\op_1$};
\node[opstyle](op1-2) at (1,-1) {$\op_2$};

\node[opstyle](op2-1) at (2.25,-0.5) {$\op_1$};
\node[opstyle](op2-2) at (2.25,-1.5) {$\op_2$};
\node[opstyle](op2-3) at (2.25,1.5) {$\op_1$};
\node[opstyle](op2-4) at (2.25,0.5) {$\op_2$};

\node[opstyle](op3-1) at (4,-0.5) {$\op_1$};
\node[opstyle](op3-2) at (4,-1.5) {$\op_2$};
\node[opstyle](op3-3) at (4,1.5) {$\op_1$};
\node[opstyle](op3-4) at (4,0.5) {$\op_2$};

\node[opstyle](op4-1) at (5.75,-0.5) {$\op_1$};
\node[opstyle](op4-2) at (5.75,-1.5) {$\op_2$};
\node[opstyle](op4-3) at (5.75,1.5) {$\op_1$};
\node[opstyle](op4-4) at (5.75,0.5) {$\op_2$};

\node[boxstyle](output) at (6.75,0) {};

\draw[->,color=red]   (input) -- (op1-1);
\draw[->]   (input) -- (op1-2);

\draw[->]   (op1-2) -- (op2-1);
\draw[->]   (op1-2) -- (op2-2);
\draw[->,color=red]   (op1-1) -- (op2-3);
\draw[->]   (op1-1) -- (op2-4);

\draw[->]   (op2-1) -- (op3-3);
\draw[->]   (op2-1) -- (op3-4);
\draw[->]   (op2-2) -- (op3-1);
\draw[->]   (op2-2) -- (op3-2);
\draw[->]   (op2-3) -- (op3-3);
\draw[->,color=red]   (op2-3) -- (op3-4);
\draw[->]   (op2-4) -- (op3-1);
\draw[->]   (op2-4) -- (op3-2);

\draw[->]   (op3-1) -- (op4-3);
\draw[->]   (op3-1) -- (op4-4);
\draw[->]   (op3-2) -- (op4-1);
\draw[->]   (op3-2) -- (op4-2);
\draw[->]   (op3-3) -- (op4-3);
\draw[->]   (op3-3) -- (op4-4);
\draw[->]   (op3-4) -- (op4-1);
\draw[->,color=red]   (op3-4) -- (op4-2);

\draw[->]   (op4-1) -- (output);
\draw[->,color=red]   (op4-2) -- (output);
\draw[->]   (op4-3) -- (output);
\draw[->]   (op4-4) -- (output);
\end{tikzpicture}%

  \caption{Example for weight sharing with only two operations in the chain-structured search space.
  Every box has its own weights, every path (e.g. the red one: $\op_1\rightarrow \op_1\rightarrow \op_2\rightarrow \op_2$) is one architecture in the search space.
  Thus, weights are shared across different architectures.}
    \label{fig:opt-one-shot-weight-sharing}
\end{figure}
\paragraph{Weight Sharing}
\cite{Pham2018_ENAS} search in a subspace of the NASNet search space (see Section~\ref{sub:space-cell}) and operate on an over-parameterized network that covers the entire search space.
They use reinforcement learning to learn a controller to sample architectures (which are only subgraphs of the complete architecture) from the search space (see Figure~\ref{fig:opt-one-shot-weight-sharing}).
The weights of the controller and a part of the over-parameterized network are alternately updated by gradient descent.
During the network parameter update step, a batch is selected and an architecture is sampled from the search space.
Only the parameters of this architecture are updated, the remaining parameters and the parameters of the controller remain unchanged. During the update of the controller, architectures are sampled and evaluated on a validation batch.
The obtained reward is used to update the controller similar to \cite{Zoph2017_Neural} (details provided in Section~\ref{sub:opt-rl}).
This idea is referred to as weight sharing, as it corresponds to training multiple networks with shared weights.
\cite{Bender2018_Understanding} suggest to replace the controller with uniform sampling.
\cite{Casale2019_ProbNAS} learn the parameters of an independent categorical distribution to sample architectures.
In all these variations, after the training step, the final architecture is selected according to the one-shot model's validation performance and trained from scratch.

\paragraph{Differentiable Architecture Search}
By introducing a set of binary variables $\{\arch_{i,j,k}\}$, an equivalent formulation of Equation \eqref{eq:space-recursive-def} can be obtained as
\begin{equation}
    \nodei{k} = \sum_{j\in\left|\opset\right|} \opi{j}\left(\left\{\nodei{i}\ |\ \arch_{i,j,k}=1,\ i<k\right\}\right)\ .
\end{equation}
For notational convenience, we will assume that each operation which determines $\nodei{k}$ can only use a single input from $\mathcal{I}^{(k)}$.
Then, this definition simplifies to
\begin{equation}
    \nodei{k} = \sum_{i\in\mathcal{I}^{(k)}}\sum_{j\in\left|\opset\right|} \arch_{i,j,k}\cdot\opi{j}\left(\left\{\nodei{i}\right\}\right)\,,
\end{equation}
with $\arch_{i,j,k}\in\left\{0,1\right\}$.
In the works discussed thus far, the assumption has been that every operation is either part of the network or not.
\cite{Liu2018_DARTS} relax this assumption and instead assumes a linearly weighted combination where the $\arch_{i,j,k}$ can assume any real value in the range of 0 to 1, which allows for softer decisions on paths.
They parameterize $\vec{\alpha}$ by structural parameters $\vec{\beta}$,
\begin{equation}
    \arch_{i,j,k}=\frac{\exp\left(\beta_{i,j,k}\right)}{\sum_{i\in\mathcal{I}^{(k)}}\sum_{j\in\left|\opset\right|}\exp\left(\beta_{i',j',k}\right)}\,.
\end{equation}
This softmax operation ensures that for every $k$
\begin{equation}
    \sum_{i\in\mathcal{I}^{(k)}}\sum_{j\in\left|\opset\right|} \arch_{i,j,k}=1\,.
\end{equation}
From Equation \eqref{eq:opt-problem-definition} a new, differentiable loss function for both structural parameters and model parameters $\vec{\modelparams}$ can be derived:
\begin{equation}
    \min_{\vec{\arch}\left(\vec{\beta}\right)\in\searchspace} \loss\left(\argmin_{\model_{\vec{\arch}\left(\vec{\beta}\right),\vec{\modelparams}}\in\modelspace_{\vec{\arch}\left(\vec{\beta}\right)}}\,\loss\left(\model_{\vec{\arch}\left(\vec{\beta}\right),\vec{\modelparams}}, \Dtrain\right) + \reg\left(\vec{\modelparams}\right),\Dvalid\right)\,.
\end{equation}
\cite{Liu2018_DARTS} propose an alternating optimization method which learns the model parameters by minimizing the loss on the training set and the structural parameters by minimizing the loss on the validation set using gradient-based optimization methods.
After training this network, the final architecture is chosen based on the values of $\arch_{i,j,k}$, the larger the better.
This method is a very elegant solution that makes all parameters differentiable.
However, it has a significant drawback: all parameters must be kept in memory all the time.
Since the network covers the entire search space, this is a significant disadvantage.
The following two works overcome this by introducing update rules which require to keep only a part of the network in memory at any one time.

The work by \cite{Xie2019_SNAS} builds upon the work of \cite{Liu2018_DARTS}, but has three major differences.
First, they represent the architecture by $p\left(\vec{\beta}\right)$, where $p\left(\vec{\beta}\right)$ is fully factorizable and is modeled with a concrete distribution~\citep{Maddison2017_The}.
Choosing this distribution results in only one path of the over-parameterized network being selected for training.
This path corresponds to a feasible solution in the search space and accordingly requires less memory and can be easily kept in memory.
Second, they minimize the loss of $\Dtrain$ with respect to the two types of parameters $\vec{\beta}$ and $\vec{\modelparams}$.
The validation partition $\Dvalid$ is used only to select the final architecture.
And finally, gradients are estimated from Monte Carlo samples rather than analytic expectation.
\cite{Cai2019_Proxyless} achieve the same effect with the use of binary gates~\citep{Courbariaux2015_BinaryConnect}.
Each binary gate chooses a part of the path based on a learned probability and in this sense is very similar to the idea of \cite{Xie2019_SNAS}.
Both these works significantly reduce the memory overload by not loading the entire over-parameterized model in the memory.
Discrete architectures are derived by selecting the most likely operation and the top two predecessors in the over-parameterized network.
This architecture is then trained again from scratch.

\paragraph{Hypernetworks}
\cite{Brock2018_SMASH} propose the use of dynamic hypernetworks \citep{Ha2017_HyperNetworks}, a neural network which generates the weights for another network conditioned on a variable, i.e. in this case the architecture description.
The hypernetwork is trained to generate network weights for a variety of architectures.
It can be used to rank different architectures and derive a final architecture which is then trained from scratch.
This method also shares weights, but most of them are shared in the hypernetwork.
Hypernetworks save training time of candidate architectures during the search as the weights are obtained as predictions from these hypernetworks.

\cite{Zhang2019_Graph} extend this idea by combining it with a graph neural network~\citep{Scarselli2009_The} which directly operates on a graph defined by nodes and edges.
Every node is a recurrent neural network, and messages between these networks can be propagated using the edges.
Each node stores its own state, which is updated by means of message propagation.
For each architecture, a graph neural network is generated which is homomorphic.
This means we have one node for each operation in the architecture, and for each connection between operations we have an edge in the graph neural network.
The hypernetwork is then conditioned on the states of the graph neural network to infer the weights of the architecture.

\paragraph{Discussion}
All of the one-shot approaches share the weights across different sampled architectures, and rely on the hypothesis that the trained over-parameterized network can be used to rank architectures for quality.
This belief is supported by \cite{Bender2018_Understanding} who show a correlation between the validation error obtained for a sample of the over-parameterized network and the one obtained after training the corresponding architecture from scratch.
However, they acknowledge that training the over-parameterized network is not trivial and that the practical approaches of batch normalization, dropout rate and regularization play an important role.
\cite{Zhang2019_Graph} conduct a similar experiment using a separate implementation and confirm the correlation between the two variables.
The correlation for the hypernetwork-based optimizers has been independently confirmed by \cite{Brock2018_SMASH} and \cite{Zhang2019_Graph}.
Thus, we have sufficient empirical evidence that weight sharing is indeed a useful tool for the search of CNN architectures.
As a result, the search only takes up insignificantly additional time than training a single candidate in the search space and is consequently incredibly efficient compared to other optimization methods.
\cite{Sciuto2019_Evaluating} challenge this widespread belief for its applicability for searching recurrent cells.
They show that the ranking of architectures implied by the various methods \citep{Pham2018_ENAS,Liu2018_DARTS,Luo2018_NAO} does not correlate with the true ranking.
The reason that the search methods still deliver good results is solely due to the very limited search space.
They provide empirical evidence that a random search outperforms many of the previously described methods in the search for an RNN cell.

\begin{table}
  \centering
  \caption{We give an overview of the results obtained and the search time required on CIFAR-10 by the various search algorithms discussed.
  In addition, we list the results of various random searches and human-designed architectures.}
  \label{tab:opt-results-cifar10}
  \begin{tabular}{c|lc>{\centering\arraybackslash}p{1.8cm}>{\centering\arraybackslash}p{1.2cm}}
    \hline\noalign{\smallskip}
    \multicolumn{1}{l}{} & Reference & Error (\%) & Params (Millions) & GPU Days \\
    \noalign{\smallskip}
    \hline
    \noalign{\smallskip}
    \parbox[t]{3mm}{\multirow{8}{*}{\rotatebox[origin=c]{90}{RL}}}
    &\cite{Baker2017_Designing} & 6.92 & 11.18 & 100\\
    &\cite{Zoph2017_Neural} & 3.65 & 37.4 & 22,400\\
    &\cite{Cai2018_Efficient} & 4.23 & 23.4 & 10\\
    &\cite{Zoph2018_Learning} & 3.41 & 3.3 & 2,000\\
    &\cite{Zoph2018_Learning} + Cutout & 2.65 & 3.3 & 2,000\\
    &\cite{Zhong2018_Practical} & 3.54 & 39.8 & 96\\
    &\cite{Cai2018_Path} & 2.99 & 5.7 & 200\\
    &\cite{Cai2018_Path} + Cutout & 2.49 & 5.7 & 200\\
    \noalign{\smallskip}
    \hline
    \noalign{\smallskip}
    \parbox[t]{3mm}{\multirow{7}{*}{\rotatebox[origin=c]{90}{EA}}}
    &\cite{Real2017_Large} & 5.40 & 5.4 & 2,600\\
    &\cite{Xie2017_Genetic} & 5.39 & N/A & 17\\
    &\cite{Suganuma2017_A} & 5.98 & 1.7 & 14.9\\
    &\cite{Liu2018_Hierarchical} & 3.75 & 15.7 & 300\\
    &\cite{Real2019_Aging} & 3.34 & 3.2 & 3,150\\
    &\cite{Elsken2018_Simple} & 5.2 & 19.7 & 1\\
    &\cite{Wistuba2018_Deep} + Cutout & 3.57 & 5.8 & 0.5\\
    \noalign{\smallskip}
    \hline
    \noalign{\smallskip}
    \parbox[t]{3mm}{\multirow{3}{*}{\rotatebox[origin=c]{90}{SMBO}}}
    &\cite{Kandasamy2018_Neural} & 8.69 & N/A & 1.7\\
    &\cite{Liu2018_Progressive} & 3.41 & 3.2 & 225\\
    &\cite{Luo2018_NAO} & 3.18 & 10.6 & 200\\
    \noalign{\smallskip}
    \hline
    \noalign{\smallskip}
    \parbox[t]{3mm}{\multirow{9}{*}{\rotatebox[origin=c]{90}{One-Shot}}}
    &\cite{Pham2018_ENAS} & 3.54 & 4.6 & 0.5\\
    &\cite{Pham2018_ENAS} + Cutout & 2.89 & 4.6 & 0.5\\
    &\cite{Bender2018_Understanding} & 4.00 & 5.0 & N/A\\
    &\cite{Casale2019_ProbNAS} + Cutout & 2.81 & 3.7 & 1\\
    &\cite{Liu2018_DARTS} + Cutout & 2.76 & 3.3 & 4\\
    &\cite{Xie2019_SNAS} + Cutout & 2.85 & 2.8 & 1.5\\
    &\cite{Cai2019_Proxyless} + Cutout & 2.08 & 5.7 & 8.33\\
    &\cite{Brock2018_SMASH} & 4.03 & 16.0 & 3\\
    &\cite{Zhang2019_Graph} & 2.84 & 5.7 & 0.84\\
    \noalign{\smallskip}
    \hline
    \noalign{\smallskip}
    \parbox[t]{3mm}{\multirow{4}{*}{\rotatebox[origin=c]{90}{Random}}}
    & \cite{Liu2018_Hierarchical} & 3.91 & N/A & 300\\
    & \cite{Luo2018_NAO} & 3.92 & 3.9 & 0.3\\
    & \cite{Liu2018_DARTS} + Cutout & 3.29 & 3.2 & 4\\
    & \cite{Li2019_Random} + Cutout & 2.85 & 4.3 & 2.7\\
    \noalign{\smallskip}
    \hline
    \noalign{\smallskip}
    \parbox[t]{3mm}{\multirow{5}{*}{\rotatebox[origin=c]{90}{Human}}}
    &\cite{Zagoruyko2016_Wide} & 3.87 & 36.2 & - \\
    &\cite{Gastaldi2017_Shake} (26 2x32d) & 3.55 & 2.9 & - \\
    &\cite{Gastaldi2017_Shake} (26 2x96d) & 2.86 & 26.2 & - \\
    &\cite{Gastaldi2017_Shake} (26 2x112d) & 2.82 & 35.6 & - \\
    &\cite{Yamada2016_Deep} + ShakeDrop & 2.67 & 26.2 & - \\
    \hline
  \end{tabular}
\end{table}
\begin{table}
  \centering
  \caption{Results obtained on ImageNet by architectures discovered by neural architecture search methods as well as human-designed architectures.}
  \label{tab:opt-results-imagenet}
  \begin{tabular}{c|l>{\centering\arraybackslash}p{2.3cm}>{\centering\arraybackslash}p{1.8cm}>{\centering\arraybackslash}p{1.8cm}>{\centering\arraybackslash}p{1.2cm}}
    \hline\noalign{\smallskip}
    \multicolumn{1}{l}{} & Reference & Top 1/Top 5 Accuracy (\%) & Params (Millions) & Image Size (squared) & GPU Days \\
    \noalign{\smallskip}
    \hline
    \noalign{\smallskip}
    \parbox[t]{3mm}{\multirow{3}{*}{\rotatebox[origin=c]{90}{RL}}}
    &\cite{Zoph2018_Learning} & 82.7/96.2 & 88.9 & 331 & 2,000\\
    &\cite{Zhong2018_Practical} & 77.4/93.5 & N/A & 224 & 96\\
    &\cite{Cai2018_Path} & 74.6/91.9 & 594 & 224 & 200\\
    \noalign{\smallskip}
    \hline
    \noalign{\smallskip}
    \parbox[t]{3mm}{\multirow{4}{*}{\rotatebox[origin=c]{90}{EA}}}
    &\cite{Xie2017_Genetic} & 72.1/90.4 & 156 & 224 & 17\\
    &\cite{Liu2018_Hierarchical} & 79.7/94.8 & 64.0 & N/A & 300\\
    &\cite{Real2019_Aging} & 82.8/96.1 & 86.7 & 331 & 3,150\\
    &\cite{Real2019_Aging} & 83.9/96.6 & 469 & 331 & 3,150\\
    \noalign{\smallskip}
    \hline
    \noalign{\smallskip}
    \parbox[t]{3mm}{\multirow{2}{*}{\rotatebox[origin=c]{90}{SMBO}}}
    &\cite{Liu2018_Progressive} & 74.2/91.9 & 5.1 & 224 & 225\\
    &\cite{Liu2018_Progressive} & 82.9/96.2 & 86.1 & 331 & 225\\[1.3ex]
    \noalign{\smallskip}
    \hline
    \noalign{\smallskip}
    \parbox[t]{3mm}{\multirow{7}{*}{\rotatebox[origin=c]{90}{One-Shot}}}
    &\cite{Bender2018_Understanding} & 75.2/N/A & 11.9 & 224 & N/A\\
    &\cite{Casale2019_ProbNAS} & 74.0/91.6 & 5.6 & N/A & 1\\
    &\cite{Liu2018_DARTS} & 73.3/91.3 & 4.7 & 224 & 4\\
    &\cite{Xie2019_SNAS} & 72.7/90.8 & 4.3 & 224 & 1.5\\
    &\cite{Cai2019_Proxyless} & 75.1/92.5 & N/A & 224 & 8.33\\
    &\cite{Brock2018_SMASH} & 61.4/83.7 & 16.2 & 32 & 3\\
    &\cite{Zhang2019_Graph} & 73.0/91.3 & 6.1 & 224 & 0.84\\
    \noalign{\smallskip}
    \hline
    \noalign{\smallskip}
    \parbox[t]{3mm}{\multirow{5}{*}{\rotatebox[origin=c]{90}{Human}}}
    &\cite{Howard2017_Mobilenets} & 70.6/89.5 & 4.2 & 224 & - \\
    &\cite{Zhang2018_ShuffleNet} & 70.9/89.8 & 5.0 & 224 & - \\
    &\cite{Xie2017_Aggregated} & 80.9/95.6 & 83.6 & 320 & - \\
    &\cite{Zhang2017_PolyNet} & 81.3/95.8 & 92.0 & 331 & - \\
    &\cite{Chen2017_Dual} & 81.5 95.8 & 79.5 & 320 & - \\
    \hline
  \end{tabular}
\end{table}
\subsection{Conclusions}
In this section, we reviewed various optimization algorithms based on methods such as reinforcement learning, evolutionary algorithms, surrogate model-based optimization, and one-shot models.
Naturally, the imminent question is - which method should be prescribed?
In order to answer that we need to inspect if the results can be fairly compared.
In our opinion, this is not the case.
The different experiments differ drastically in terms of search space, search duration and data augmentation.
However, we can safely conclude that the NASNet search space is the most commonly used search space, presumably because the imposed restrictions in its definition favor the discovery of well-performing architectures.
Similarly, strategies that reuse parameters (through function-preserving transformations or weight sharing) are effective in reducing the search duration.
However, we would like to re-emphasize the questionable benefits of weight sharing for RNNs as noted in our previous discussion.

At this point we need to criticize the lack of fair baselines.
Although architecture search can be considered as a special way of optimizing hyperparameters, most of the related work is disregarded.
In particular, random search which has proven to be an extremely strong baseline.
In fact, \cite{Sciuto2019_Evaluating} show that random search finds better RNN cells than any other optimizer.
\cite{Li2019_Random} also confirm this result and additionally show that random search finds architectures that perform at least as well as the ones obtained from established optimizers for CNNs.
It is worth noting that these random optimizers work on a search space that is known to sample well-performing architectures.
\cite{Xie2019_Exploring} go one step further and use a graph generator to generate random graph structures that no longer adhere to the rules of established search spaces.
Although these graph generators do not have any deep learning specific prior, the generated architectures perform better than those found by complex architecture optimizers.
Architectures designed by humans often serve as a motivation to justify the search for neural architectures.
However, state-of-the-art architectures are often not considered for comparison or not trained under equivalent experimental conditions.
Thus, the effect of learning rate decay strategies, augmenting techniques (cutout~\citep{DeVries2018_Improved}), regularization tricks (ScheduledDropPath~\citep{Zoph2018_Learning}), and other nuanced strategies that effect training dynamics gets overshadowed.
We include the results for the state-of-the-art architectures for CIFAR-10 in Table~\ref{tab:opt-results-cifar10}  and report additional results for models trained with the popular data-augmentation technique of cutout in Table~\ref{tab:other-autoaugment-results}.
Results for ImageNet are reported in Table~\ref{tab:opt-results-imagenet}.
We note that the performance gap between human-designed and discovered architectures is smaller than what is often claimed or reported, in particular for CIFAR-10.

We recognize that in recent years many interesting and creative architecture search methods have been developed, but we argue that the source of performance gains still remains unclear.
This remains a highly relevant and unanswered question for this research area.

\section{Early Termination of Training Processes}\label{sec:early-termination}

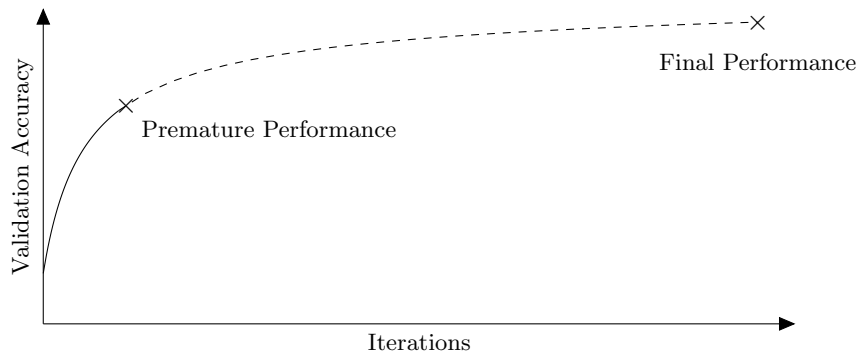
\begin{figure}[t]
  \centering
    \tikzsetnextfilename{early-termination-learning-curve}%
    \begin{tikzpicture}
  \draw[->] (0,0) -- (10,0) node[below,pos=0.5]{\footnotesize Iterations};
  \draw[->] (0,0) -- (0,4.2) node[above,pos=0.5,rotate=90]{\footnotesize Validation Accuracy};
  \draw[scale=0.5,domain=0:2,smooth,variable=\x,black] plot ({\x},{10-6/ln(\x+2)});
  \draw[scale=0.5,domain=2:19,dashed,variable=\x,black] plot ({\x},{10-6/ln(\x+2)});
  \node at (9.5,4) (mp-cross) {$\times$} ;
  \node[below=0.01 of mp-cross] {\footnotesize Final Performance};
  \node at (1.1,2.9) {$\times$} ;
  \node at (3,2.6) {\footnotesize Premature Performance};
\end{tikzpicture}%

  \caption{Learning curve prediction tries to predict from the premature learning curve (solid line) the final performance. This can significantly reduce the number of required training iterations.}
  \label{fig:early-termination-learning-curve}
\end{figure}
Although expensive, training is an inevitable step to obtain the validation accuracy on a dataset.
For neural architecture search, however, it is not necessary to know the validation accuracy, it is sufficient to know whether an architecture candidate is better or worse than the current best solution.
Human experts estimate the likelihood if the current candidate can surpass the current best solution by analyzing the behavior of its learning curve.
We refer to the function that maps the number of iterations used for training a model to its validation accuracy at that time as its learning curve (Figure~\ref{fig:early-termination-learning-curve}).
If there is a high probability that a current training process does not lead to an improved solution, the process is terminated.
The time thus saved can now be used for other training processes.
Learning curve prediction methods are automated methods that try to estimate this probability.
The previously discussed surrogate models (Section~\ref{sub:opt-smbo}) are related to these methods.
The main difference is that learning curve prediction methods require additional training of a candidate and take the learning curve into account.

Many approaches in neural architecture search do not train models to completion.
Instead, the validation accuracy of this premature model is used as an approximation for the final performance.
One popular hyperparameter optimization which relies on this premature performance is Hyperband~\citep{Li2017_Hyperband}.
In this work, a random set of candidate models is trained in parallel and after every few iterations (the considered number of iterations is increased exponentially), training of a large fraction of candidates is terminated based on the premature performance.
In this way the number of candidates reduces exponentially, while at the same time the training time per candidate is also increased at an exponential rate.
Eventually only one candidate remains which serves as the output of the search algorithm.

In the following we discuss more sophisticated approaches which try to predict the final performance based on architecture and data information as well as the premature learning curve.
All these methods use a model to predict the final performance of a training job.
This model provides the probability that a premature learning curve will exceed a particular accuracy.
This is used to determine the likelihood that the current training job will provide a network with higher accuracy than the current best one.
If this likelihood does not exceed a predetermined threshold, training is terminated.

\paragraph{\cite{Domhan2015_Speeding}}
Typically learning curves are increasing and saturating functions.
\cite{Domhan2015_Speeding} build on this observation and select a set of eleven different increasing and saturating function families and combine them by a weighted sum to probabilistically model a learning curve.
However, since it does not learn across learning curves, it requires a relatively long premature learning curve in order to yield accurate predictions.

\paragraph{\cite{Klein2017_Learning}}
\cite{Klein2017_Learning} extend the work by \cite{Domhan2015_Speeding} to learn the learning curve prediction model across different learning curves.
For this purpose, the function families are combined with a Bayesian neural network which infers the parameters of the functions and their weights conditioned on the premature learning curve.
Parameters of the Bayesian neural network are learned on a set of mature learning curves.

\paragraph{\cite{Chandrashekaran2017_Speeding}}
The two previously discussed works make strong assumptions regarding the learning curves by using a set of function families which might not apply generally.
Moreover, noisy optimization or learning rate schedules may violate the assumption made about the learning curve behavior.
For instance, a step-wise decay of the learning rate violates the assumption that the learning curve is a quickly saturating function and a cyclic learning rate schedule violates the assumption that the learning curve is an increasing function.
\cite{Chandrashekaran2017_Speeding} overcome this problem by using mature learning curves observed during training rather than a predefined set of functions.
An affine transformation is used to reduce the squared error between all mature learning curves and a premature learning curve.
The prediction model is formed as an average of the best fits of mature learning curves.

\paragraph{\cite{Baker2018_Accelerating}}
All previously discussed learning curve predictors are based only on the premature learning curve as well as machine learning models specifically designed for learning curve prediction. 
\cite{Baker2018_Accelerating} approach this problem differently.
First, they enrich the representation and include additional features like the first and second derivative of the premature learning curve, properties of the architecture (number of parameters and layers) and chosen hyperparameters (e.g. initial learning rate and weight decay).
Second, they study a set of traditional regression models to solve the prediction task.
Empirically, they find that the $\nu$-Support Vector Regression~\citep{Scholkopf2000_New} performs best among other alternatives such as random forest or Bayesian linear regression.
Furthermore, they report that the premature performance can be a strong competitor method which in most cases outperforms the work by \cite{Domhan2015_Speeding} and \cite{Klein2017_Learning}.
While \cite{Domhan2015_Speeding} did not compare against the premature performance,  \cite{Klein2017_Learning} only report results with respect to the mean squared error.
However, the squared error is no relevant measure for this task and it is by no means surprising that the premature performance has a high squared error.
Instead, the ranking of learning curves is more important and thus a measure like $R^2$ (as used by \cite{Baker2018_Accelerating}) is more appropriate.
Finally, \cite{Baker2018_Accelerating} are also the first and only to use a more sophisticated learning curve prediction method in combination with a neural architecture search method.
They build upon their previous work~\citep{Baker2017_Designing}, report a speed up by a factor of 3.8, and find equally good neural architectures.

\section{Transfer Learning}\label{sec:transfer-learning}

Transfer learning refers to the idea of reusing knowledge gained when solving one task in order to solve a different but related task more efficiently.

In the context of neural architechtures search, many simple techniques have proven to be effective.
Some works~\citep{Baker2017_Designing} demonstrate that an architecture discovered for one dataset can be easily applied (without introducing significant changes) to a dataset with the same image dimensions.
\cite{Zoph2018_Learning} demonstrate easy transferability of architectures discovered on datasets with few data points and a small resolution (CIFAR-10) to datasets with significantly more data and a higher resolution (ImageNet).
Furthermore, architectures which perform well across a diverse set of tasks provide a competitive solution~\citep{Wistuba2019_Inductive}.
Other similar paradigms include searching on a subset of the data~\citep{Sabharwal2016_Selecting} and searching on a down-scaled version of the data~\citep{Hinz2018_Speeding}.
It is also a common practice to search for networks with few cell repetitions and for cells that use only few filters.
Then, the final model is obtained by increasing the number of filters and number of cell repetitions significantly to build larger models with better performance.

Transferring knowledge across tasks provides another alternative avenue that can be leveraged to improve the performance of neural architecture search methods.
Most search algorithms train multiple sample architectures during the search process.
In the same spirit as the use of surrogate models, training of these sampled architectures can be made more efficient with the use of additional information gained from training architectures on other datasets of tasks.
In order to learn dataset-specific properties, features which describe properties of the dataset, also called meta-features~\citep{Reif2014_Automatic}, are considered.
We discuss some of these approches in this section.

\paragraph{\cite{Wong2018_Transfer}}
\cite{Wong2018_Transfer} use transfer learing to automate the fine-tuning of architectures across tasks.
They extend the controller of \cite{Zoph2017_Neural} to include task-specific embeddings as input to the controller to account for the different tasks in the policy.
The controller generates a sequence of discrete actions that specify design choices for fine-tuning a network architecture.
This includes choosing the embeddings, the structure of the classifier and various hyperparameters.
The parameters of the controller are updated with REINFORCE.
In their work they explore transfer learning for both image classification and natural language processing tasks.

\paragraph{\cite{Istrate2019_TAPAS}}
The approach by \cite{Istrate2019_TAPAS} is based on a surrogate model which predicts the performance of neural architectures for a new dataset based on performance observations of neural architectures on other datasets.
They use as \emph{landmarker}~\citep{Pfahringer2000_Meta}, the accuracy of one computationally inexpensive a CNN model on new dataset, which is their only meta-feature.
This meta-feature is used to determine the most similar datasets.
It is also used together with the architecture encoding as input for the surrogate model.
The observations on datasets with similar validation accuracy are used to train the surrogate model.
An evolutionary algorithm with a setup similar to \cite{Real2017_Large} is used to obtain the the architecture that maximizes the surrogate model's prediction.
This architecture is the outcome of the search process and is trained from scratch.

\paragraph{\cite{Kokiopoulou2019_Fast}} 
\cite{Kokiopoulou2019_Fast} propose to automate fine-tuning for transferring models across different text classification tasks by using a surrogate model.
This model is trained to predict the validation accuracy for a given set of meta-features and architecture design choice.
Contrary to other works, they propose to learn the meta-features from the raw data rather than relying on manually engineered meta-features or embeddings.
The surrogate model consists of two components.
While the first component learns to compute predictive meta-features given a small amount of data, the second component predicts the validation accuracy for an architecture based on the architecture description and the meta-features.
These components are trained across multiple datasets and then used for a new dataset.
Instead of just predicting the accuracy of several architectures, the authors propose to search it by means of gradient ascent.
For this purpose they use a continuous parametrization for the neural architectures.

\paragraph{\cite{Xue_2019_Transferable}}
\cite{Xue_2019_Transferable} assume a lifelong neural architecture search setting with an incoming stream of datasets.
For each new dataset, a set of landmarkers is trained to establish a dataset representation.
Based on this representation, it is decided whether the dataset resembles a previously handled one, or whether it is a fundamentally different one.
For the former case, they avoid a search and reuse the solution for the previously handled dataset.
Otherwise, a new search with any arbitrary optimizer is initiated for the dataset.

\section{Constraints and Multiple Objectives}\label{sec:mo}

While it is important to find networks that yield high accuracy, sometimes for a practitioner it is also imperative to consider other objectives, such as the number of model parameters, the number of floating point operations, and device-specific statistics like the latency of the model.
There exist two different approaches which account for these additional objectives.
In the first approach, the conditions are added as constraints to the optimization problem so as to enforce requirements like fewer parameters or faster inference time.
The exact form of the constraints and the trade-off between different constraints can be adjusted as per practical requirements.
In the second approach, the problem is tackled as a multi-objective function optimization problem which yields a set of proposals as a solution \citep{Deb2014_Multi}.
This separates the decision making into two steps.
In the first step, a set of candidates is obtained without considering any trade-offs between the different objectives and then the decision for a solution is made in the second step.
This separation allows for reusing the set of candidate solution for different specifications.

\subsection{Constrained Optimization}
For some tasks modeled with deep learning, specific constraints $g_i$, such as, thresholds on inference time or memory requirement are explicitly stated.
For such cases the single-objective optimization problem defined in Equation~\eqref{eq:opt-problem-definition} then turns into a constrained optimization problem as defined by
\begin{subequations}
    \begin{alignat}{2}
        &\max_{\vec{\arch}\in\searchspace} &\qquad& \f\left(\vec{\arch}\right)\\
        &\text{subject to} & & g_i\left(\vec{\arch}\right)\leq c_i\ \forall i\in I\,.
    \end{alignat}
\end{subequations}
The optimization methods discussed in Section \ref{sec:opt} cannot be directly adopted for this scenario as they were designed to solve an unconstrained optimization problem.
However, the classical technique of penalty methods \citep{Fiacco1990_Nonlinear} can be used to form the unconstrained optimization given by
\begin{equation}
  \max_{\vec{\arch}\in\searchspace}\f\left(\vec{\arch}\right)\cdot\prod_{i\in I}\penaltyfct\left(g_i\left(\vec{\arch}\right),c_i\right)\,.
  \label{eq:mo-penalty-method}
\end{equation}
The function $\penaltyfct$ is a penalty function, that punishes any violations of  constraints.
Once a penalty function has been selected, most optimization method discussed in Section \ref{sec:opt} can be directly applied to solve the problem.
In the following, we discuss the various penalty functions considered in the literature along with the specific optimizer used by different approaches.

\cite{Tan2018_MnasNet} use
\begin{equation}
    \penaltyfct\left(g_i\left(\vec{\arch}\right),c_i\right)=\left[\frac{g_i\left(\vec{\arch}\right)}{c_i}\right]^{w_i}\,,
\end{equation}
as a penalty function, where $\vec{w}$ is treated as a hyperparameter and is set to suit the desired trade-off.
They use the reinforcement learning optimizer proposed by \cite{Zoph2017_Neural} in the factorized hierarchical search space as discussed in Section \ref{sub:space-cell}.
\cite{Zhou2018_Resource} propose to use
\begin{equation}
    \penaltyfct\left(g_i\left(\vec{\arch}\right),c_i\right)=\phi^{\max\left\{0, \left(g_i\left(\vec{\arch}\right)-c_i\right)/c_i\right\}}\,,
\end{equation}
where $\phi$ is a penalization constant in the range of 0 to 1 and use a similar approach to \cite{Cai2018_Efficient} to find the final architecture.
Specifically, a policy is learned that decides whether to add, remove or keep a layer as well as whether to alter its number of filters.
\cite{Hsu2018_MONAS} use a harder penalty function which returns 0 when the constraint is violated and 1 otherwise.
They use a reinforcement learning optimizer similar to the one proposed by \cite{Zoph2017_Neural} which predicts the hyperparameters of different layers.
Thus, the reward received by the controller is the accuracy for cases which do not violate any constraint and is 0 otherwise.
The authors evaluate their optimizer to select the layer-wise hyperparameters (number of filters, kernel size) of an AlexNet~\citep{Krizhevsky2012_Imagenet} and the cell-wise hyperparameters (stage, growth rate) of a CondenseNet~\citep{Huang2018_CondenseNet}.
We refer to Section~\ref{sub:opt-rl} for more details on the optimization methods.

\subsection{Multi-Objective Optimization}

Another approach to handle multiple fronts lies in the formalism of multi-objective optimization problem~\citep{Hwang2012_Multiple} defined as
\begin{equation}
  \max_{\vec{\arch}\in\searchspace} \f_1\left(\vec{\arch}\right),\f_2\left(\vec{\arch}\right),\ldots,\f_n\left(\vec{\arch}\right)\,.
\end{equation}
However, often there is no single optimal solution that simultaneously minimizes all these functions.
As some of the objectives can be conflicting, an optimal solution with respect to one objective may not be optimal with respect to another.
Therefore, in such scenarios, the task boils down to finding a set of solutions which are \textit{Pareto optimal}.
A solution is Pareto optimal if none of the objectives can be improved without worsening at least one other objective.
This means that for a Pareto optimal solution $\vec{\arch}$, no other solution $\vec{\arch}'\in\searchspace$ exists such that $\f_i\left(\vec{\arch}\right)\geq\f_i\left(\vec{\arch}'\right)$ with $\f_j\left(\vec{\arch}\right)>\f_j\left(\vec{\arch}'\right)$ for some $j$.
Otherwise, we say $\vec{\arch}'$ dominates $\vec{\arch}$, $\vec{\arch}'\prec\vec{\arch}$.
Finally, it is up to the user to select a model from the    set of Pareto optimal solutions, the Pareto front, maybe depending on different deployment scenarios.

\paragraph{Decomposition Methods}
One way to solve this problem is the decomposition approach.
A parameterized aggregation function $h$ is used to transform the multi-objective optimization problem into a single-objective optimization problem
\begin{equation}
  \max_{\vec{\arch}\in\searchspace} h\left(\left(\f_1\left(\vec{\arch}\right),\f_2\left(\vec{\arch}\right),\ldots,\f_n\left(\vec{\arch}\right)\right),\vec{w}\right)\,.
\end{equation}
Examples for $h$ are the weighted sum, weighted exponential sum, weighted min-max or weighted product \citep{Pineda2014_A}.
However, solving this problem for a fixed setting of the weights $\vec{w}$ will in most cases not find all Pareto optimal solutions.
Therefore, the problem is solved for different $\vec{w}$.
In neural architecture search this requires multiple optimization runs for several different weight vectors which is prohibitive.
Therefore, a common approach is to select an aggregation function and fix the weight vector according to domain knowledge and the desired trade-off between objectives.
Consequently, the multi-objective becomes a single-objective optimization problem which can be solved by any method discussed in Section \ref{sec:opt}.
\cite{Hsu2018_MONAS} propose to use the weighted sum as the aggregation function and use a reinforcement learning approach \citep{Zoph2017_Neural} to solve this problem.
Unfortunately, not every objective function is differentiable which is a requirement for some of the optimization methods discussed in Section~\ref{sub:opt-one-shot}.
\cite{Cai2019_Proxyless} demonstrate one way to overcome this obstacle.
They propose to replace non-differentiable objective functions with differentiable surrogate models.
Specifically, they use a surrogate model to predict the latency of an operation.

\paragraph{NSGA-II}
An alternative approach is to estimate architectures which are not dominated by the current solution set and evaluate their performance.
NSGA-II~\citep{Deb2002_NSGA} is an elitist evolutionary algorithm that is utilized predominantly.
The initial population is selected at random.
Candidate solutions are sorted into different lists, called fronts.
All the non-dominated solutions belong to the first front.
The solutions  in the $i^{th}$ front are only dominated by all the solutions in the $1,\ldots,i-1$ fronts.
Solutions within a front are sorted according to their crowding distance which is a measure for the density of solutions within this solution's region.
It is computed by the sum of all the neighborhood distances across all the objectives and ensures that the algorithm explores  diverse solutions in the search space.
This yields a ranking of all solutions where the solution in front 1 with lowest density is the best one.
Now, the top $k$ solutions are selected as parents, mutated, recombined and evaluated.

\cite{Kim2017_NEMO} and \cite{Lu2018_NSGA_NET} both employ NSGA-II in order to tackle the problem.
The early work of \cite{Kim2017_NEMO} encodes the genotype by means of layer type and number of outputs leading to only chain-structured networks.
In their experiments \cite{Lu2018_NSGA_NET} report results on both NASNet search space and the genotype representation by \cite{Xie2017_Genetic} (see Section \ref{sec:space} for details) and therefore are not limited to chain-structured networks.
They use the vanilla NSGA-II for modeling the evolutionary process.
However the sampling step in their approach comprises of two phases - exploration and exploitation.
For the exploration phase, the offspring is generated by applying mutations and cross-over.
In the exploitation phase new samples are obtained from a Bayesian network that models the distribution of previously trained architectures.

\paragraph{\cite{Elsken2019_Efficient}}
\cite{Elsken2019_Efficient} propose another evolutionary algorithm which shares many common aspects with their earlier work \citep{Elsken2018_Simple} (see Section~\ref{sub:opt-ea}).
No cross-over operations are used and mutations are function-preserving.
They extend the pool of mutations and include new ones that shrink an architecture by means of layer removal or filter reduction.
Similar to NSGA-II, their algorithm tries to achieve diverse solutions.
They take advantage of the fact that the objective functions can be divided into  \emph{cheap-to-evaluate objective functions} (e.g. number of parameters, inference time etc.) and \emph{expensive-to-evaluate objective functions} (e.g. validation accuracy).
A kernel density estimator is used to model the density of candidates with respect to the cheap-to-evaluate objectives in order to select candidates from low-density regions.
Finally, the expensive-to-evaluate objectives for the candidates are evaluated and the Pareto front is updated.

\paragraph{\cite{Smithson2016_Neural}}
\cite{Smithson2016_Neural} propose a surrogate model-based approach to search for a chain-structured architecture.
The search algorithm starts with a random parameter setting.
New candidates are sampled from a Gaussian distribution around previously evaluated architectures.
A surrogate model (artificial neural network) is used to predict the values for the expensive-to-evaluate objective functions, cheap-to-evaluate objectives are evaluated exactly.
Based on both the predictions and evaluations, it is determined whether this solution is dominated by an existing solution which will in turn determine the probability the candidate is accepted.
Every accepted candidate will be evaluated and thereafter the surrogate model is updated.

\paragraph{\cite{Dong2018_DPP}}
\cite{Dong2018_DPP} is another surrogate model-based approach which extends the work by \cite{Liu2018_Progressive} (see Section \ref{sub:opt-smbo}) to solve multi-objective optimization problems.
The only difference of the optimization method is how models are selected.
Instead of only considering model accuracy, they also consider all other objectives.
Feasible candidates are those which do not violate any constraints and are not dominated (according to the prediction of the surrogate model) by any known solution.

\subsection{Model Compression}

Motivated by the efficiency of network deployment on mobile devices, model compression has surfaced as a relevant goal for architecture design.
Neural architecture search methods in this context seek to find policies to prune a model so as to satisfy budget constraints in terms of number of parameters, without sacrificing a great deal in accuracy.
To this end, we discuss three important works that propose methods to automate model compression.

\paragraph{\cite{He2018_AMC}}
In their work, \cite{He2018_AMC} use a reinforcement controller to prune a trained network in a layer-wise fashion.
The controller takes as input a representation for each layer and is trained to output the desired sparsity ratio of this layer after compression.
They use deep deterministic policy gradient~\citep{Lillicrap2016_Continous} to update the parameters of the controller.
During the exploration phase, they evaluate the reward accuracy of the pruned model without fine-tuning and argue that it serves as a good proxy for the fine-tuned accuracy.

\paragraph{\cite{Ashok2018_N2N}}
\cite{Ashok2018_N2N} use reinforcement learning to learn two policies which sequentially remove and shrink layers for any given initial architecture with the aim of compressing it.
The architecture altered by the policies is trained from scratch with knowledge distillation~\citep{Hinton2015_Distilling}, where the initial architecture serves as the teacher.
REINFORCE is used to train the reinforcement learning controllers using following reward function with resembles the penalty function introduced in Equation~\eqref{eq:mo-penalty-method},
\begin{equation}
    \frac{\f\left(\vec{\arch}\right)}{\f\left(\vec{\arch}_{\text{init}}\right)} g\left(\vec{\arch}\right)\left(2-g\left(\vec{\arch}\right)\right)\,,
    \label{eq:mo-ashok-objective}
\end{equation}
where $g\left(\vec{\arch}\right)$ is the compression ratio, i.e. the number of parameters of $\vec{\arch}$ divided by number of parameters of the initial architecture $\vec{\arch}_{\text{init}}$.

\paragraph{\cite{Cao2019_Learnable}}
\cite{Cao2019_Learnable} maximize Equation~\eqref{eq:mo-ashok-objective} by using Bayesian optimization (Section~\ref{sub:opt-smbo}) with a Gaussian process built using a deep kernel~\citep{Wilson2016_Deep}.
The acquisition function is maximized by random exploration of the search space.
Candidates are derived from the initial architecture by randomly removing and shrinking layers as well as adding skip connections.
The authors highlight that the consideration of adding skip connection is crucial for the discovery of compressed architectures which are comparable to the initial architecture with respect to classification accuracy.

\section{Other Related Methods}\label{sec:other}

Methodologies originally developed for architecture search have been extended to automate other aspects of deep learning.
\begin{figure}[t]
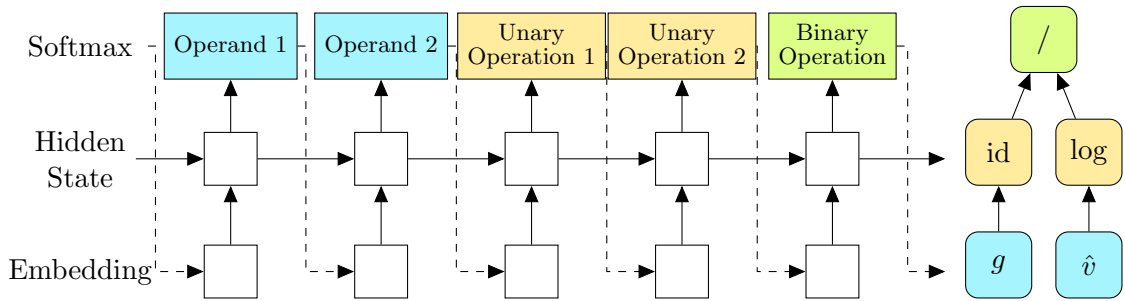

    \tikzsetnextfilename{other-controller}%
    \begin{tikzpicture}
\input{constants.tikz}
\tikzstyle{boxstyle}=[rectangle,draw=black,minimum size=20,align=center, font=\fontsize{9}{0}\selectfont]
\tikzstyle{opstyle}=[boxstyle,fill=colorOp, minimum height=25]
\tikzstyle{unaryopstyle}=[opstyle,fill=colorSegment]
\tikzstyle{binaryopstyle}=[opstyle,fill=colorCell]

\edef\prevhiddenlabel{labelinput}
\edef\xdim{2.0}

\node[rectangle,align=center](\prevhiddenlabel) at (0,0) {Hidden\\State};
\node[rectangle,align=center](labelemb) at (0,-1.5) {Embedding};
\node[rectangle,align=center](labelsm) at (0,1.5) {Softmax};

\node[opstyle](sm1) at (\xdim,1.5) {Operand 1};
\node[opstyle](sm2) at (2*\xdim,1.5) {Operand 2};
\node[unaryopstyle](sm3) at (3*\xdim,1.5) {Unary\\Operation 1};
\node[unaryopstyle](sm4) at (4*\xdim,1.5) {Unary\\Operation 2};
\node[binaryopstyle](sm5) at (5*\xdim,1.5) {Binary\\Operation};

\foreach \i in {1,...,5}
{
    \xdef\hiddenlabel{hidden\i}
    
    \node[boxstyle](\hiddenlabel) at (\xdim*\i,0) {};
    \draw[->]   (\prevhiddenlabel) -- (\hiddenlabel);
    \xdef\prevhiddenlabel{\hiddenlabel}
    
    \xdef\embeddinglabel{emb\i}
    \node[boxstyle](\embeddinglabel) at (\xdim*\i,-1.5) {};
    \draw[->]   (\embeddinglabel) -- (\hiddenlabel);
    
    \draw[->]   (\hiddenlabel) -- (sm\i);
}
\foreach \i in {1,...,4}
{
    \pgfmathtruncatemacro\result{\i + 1}
    \draw[->,dashed]   (sm\i) to (\xdim*\i+\xdim/2,1.5) to (\xdim*\i+\xdim/2,-1.5) to (emb\result);
}

\draw[->,dashed]   (0.9,1.5) to (\xdim/2,1.5) to (\xdim/2,-1.5) to (emb1);
\draw[->,dashed]   (sm5) to (\xdim*5.5,1.5) to (\xdim*5.5,-1.5)  to (\xdim*5.75,-1.5);
\draw[->]   (hidden5) to (\xdim*5.75,0);
\end{tikzpicture}%

    \tikzsetnextfilename{other-rmsprop}%
    \begin{tikzpicture}
\input{constants.tikz}
\tikzstyle{boxstyle}=[rectangle,draw=black,minimum size=25,align=center,rounded corners=1ex]
\tikzstyle{opstyle}=[boxstyle,fill=colorOp]
\tikzstyle{unaryopstyle}=[opstyle,fill=colorSegment]
\tikzstyle{binaryopstyle}=[opstyle,fill=colorCell]

\node[opstyle](op1) at (0,0) {$g$};
\node[opstyle](op2) at (1.2,0) {$\hat{v}$};
\node[unaryopstyle](uop1) at (0,1.5) {id};
\node[unaryopstyle](uop2) at (1.2,1.5) {$\log$};
\node[binaryopstyle](bop) at (0.6,3) {$/$};

\draw[->]   (op1) -- (uop1);
\draw[->]   (op2) -- (uop2);
\draw[->]   (uop1) -- (bop);
\draw[->]   (uop2) -- (bop);
\end{tikzpicture}%

  \caption{Controller output for one group. Each group consists of two operands and the unary operations applied to it as well as the binary operation which combines the outcome of the unary operations.}
  \label{fig:other-controller}
\end{figure}
For instance, \cite{Bello2017_Neural} encode a parameter optimization method used in training of a deep learning model by means of a graph structure that defines the update rules akin to classical gradient-based optimization.
An update is defined by several blocks as shown in Figure~\ref{fig:other-controller}.
A block consists of two operands, one unary function for each of the operands and a binary function.
The unary functions are applied on the operands and the binary function is applied to combine the respective outcomes.
The output of this block becomes a possible operand itself and can be used by other blocks.
This structure is stacked up to a depth of three.
Possible operands include different orders of the gradient ($g,g^2,g^3$), its bias-corrected running estimates, differently scaled versions of the parameter, random noise, constants, and the sign of the gradient.
Additionally the classical updates of Adam and RMSProp optimizer are also encapsulated as operands.
Examples for unary functions are the identity function, clipping functions, the sign function, functions that modify the gradient to zero with a certain probability, as well as logarithmic and exponential function.
Binary functions include addition, subtraction, multiplication, and division.
A reinforcement controller, similar to the one proposed by \cite{Zoph2017_Neural} (see Section \ref{sub:opt-rl}), is applied to search for an optimal structure.
The main difference in the work of \cite{Bello2017_Neural} is that they use Trust Region Policy Optimization~\citep{Schulman2015_Trust} instead of REINFORCE for updating the controller parameters.
A deep-learning optimizer sampled from the controller is used to train a simple two-layer convolutional neural network for only five epochs and the validation accuracy is used as the reward.

\cite{Ramachandran2018_Searching} uses the aforementioned setup and search space structure to search for activation functions.
They learn the controller with Proximal Policy Optimization~\citep{Schulman2017_Proximal}.
As they seek to search for activation functions, there setup comprises of only one operand which is the input of the activation function and a different set of unary and binary operations.
Examples for unary operations are powers of different degree, various sigmoid functions and the sine function.
The set of possible binary functions on the input $(x_1,x_2)$ is extended with additional functions like $x_1\cdot\sigma\left(x_2\right)$, maximum and minimum, and a weighted average.
The reward is estimated by training a ResNet-20 with the sampled activation function on CIFAR-10 for 10,000 steps.
The authors report the discovery of the following activation function
\begin{equation}
    \operatorname{swish}\left(x\right) = x \cdot \sigma\left(x\right)\,,
\end{equation}
where $\sigma$ is the logistic function.
In their experiments, this particular activation function turned out to be more effective than other commonly used activation functions and has been used by many other authors since.
However, it is interesting to note that the discovered function closely resembles the previously proposed swish activation in an earlier work by the authors.

\cite{Cubuk2018_AutoAugment} define a similar controller and train it with Proximal Policy Optimization to optimize data augmentation policies for image classification.
The controller predicts five sub-policies, each consists of two augmentation operations and always uses horizontal flipping, random crops and a cutout operation.
Examples for augmentation operations are rotation, sheering, control the brightness, contrast and sharpness of the image.
For every image, a random sub-policy is selected and applied.
The validation accuracy is used as the reward signal.
Finally, the best sub-policies of five policies are concatenated to the final augmentation policy which consists of 25 sub-policies.

\begin{table}
  \centering
  \caption{Classification error of various architectures with different augmentation strategies.
  Whenever two numbers are reported, the left one are the results obtained by us then the right one the results reported by \cite{Cubuk2018_AutoAugment}.
  \cite{Cubuk2018_AutoAugment} do not report results for mixup.}
  \label{tab:other-autoaugment-results}
  \begin{tabular}{p{2.6cm}ccc>{\centering\arraybackslash}p{1.6cm}p{1.6cm}>{\centering\arraybackslash}p{1.6cm}}
    \hline\noalign{\smallskip}
    Model & Baseline & Cutout & Mixup & Cutout +Mixup & Auto-Augment & Auto-Augment +Mixup\\
    \noalign{\smallskip}
    \hline
    \noalign{\smallskip}
    Wide-ResNet\vfill (28-10) & 3.93/3.87 & 2.89/3.08 & 3.24 & 2.95 & 2.71/2.68 & 2.60 \\
    Shake-Shake\vfill (26 2x32d) & 3.73/3.55 & 3.03/3.02 & 3.13 & 3.10 & 2.59/2.47 & 2.85\\
    Shake-Shake\vfill (26 2x96d) & 3.15/2.86 & 2.50/2.56 & 2.37 & 2.15 & 2.16/1.99 & 1.84\\
    Shake-Shake\vfill (26 2x112d) & 3.13/2.82 & 2.55/2.57 & 2.37 & 2.12 & 2.09/1.89 & 1.83\\
    PyramidNet\vfill +ShakeDrop & 2.77/2.67 & 2.09/2.31 & 1.90 & 1.51 & 1.62/1.48 & 1.33\\
    \hline
  \end{tabular}
\end{table}
\paragraph{Experiment}
We conduct an experiment to better understand the impact of established augmentation schemes and contrast it with the found policies.
As part of this we reconducted the experiments by \cite{Cubuk2018_AutoAugment} using their publicly available code.
A Wide-ResNet~\citep{Zagoruyko2016_Wide}, three versions of Shake-Shake~\citep{Gastaldi2017_Shake} and a PyramidNet~\citep{Yamada2016_Deep} with ShakeDrop~\citep{Yamada2018_ShakeDrop} are trained with different augmentation strategies as detailed in the following.
\emph{Baseline} is horizontal flipping and random crops and \emph{cutout} additionally uses cutout~\citep{DeVries2018_Improved}.
\emph{AutoAugment} extends the cutout policy by augmentation sub-policies found during the search.
We extend this setup further by considering mixup~\citep{Zhang2018_mixup} as another augmentation technique in order to see whether the AutoAugment policy can improve over standard state-of-the-art augmentation techniques.
As we see in Table~\ref{tab:other-autoaugment-results}, the combination of cutout with mixup in many cases is better than using just one of them.
Since AutoAugment is cutout plus other augmentation techniques, perhaps it is not surprising to notice a significant lift over cutout.
We consider the combination of cutout and mixup as a useful baseline which it is not able to outperform.
However, adding mixup to AutoAugment does not hurt the performance in most cases as expected.
In cases where the combination of cutout and mixup performed better than AutoAugment, this addition provides a significant improvement resulting in the best scores.
We are able to achieve an error of 1.33\% on the CIFAR-10 object recognition task.
To the best of our knowledge, the only work claiming a lower error on this task is the work by \cite{Huang2018_GPipe} which claims 1\% error.
However, they use a pretrained model on ImageNet which has 557 million parameters, while the PyramidNet used in our experiments has ``only'' 26.2 million parameters and can be trained on a single GPU.
More importantly, we do not use any pretrained weights and train the model from scratch.
Finally, we select eight augmentation policies at random and compare them to the discovered AutoAugment policies.
The results on Wide-ResNet yield a mean error of 2.81 and ranges from 2.67 to 2.97.
Thus, the mean error across eight random policies is worse than AutoAugment, however, some of them are better.
We observe a similar behavior for Shake-Shake (26 2x32d) and  (26 2x96d) where we obtain a mean error of 2.70 and 2.24, respectively.
The errors range from 2.46 to 2.85 and 2.02 to 2.40.
Two of the eight policies sampled at random yield better results than AutoAugment for every architecture, yet again, shows the importance to consider random as a baseline.
The search space for augmentation policies seems to be defined in such a way that any augmentation policy will perform better than cutout.
Only for Wide-ResNet the worst random policy considered by us is marginally worse.

\section{Outlook and Future Applications of Neural Architecture Search Methods}
In the broader context of automated deep learning, neural architecture search only tackles one component of the pipeline.
Moreover, most of the current approaches in neural architecture search only focus on CNNs for solving the task of object recognition and RNN cells for language modeling.
However, other interesting areas have already surfaced some of which have been discussed in sections \ref{sec:transfer-learning} to \ref{sec:other}.
For example, many recent works address the idea of using knowledge across different datasets to speed up the optimization processes (Section~\ref{sec:transfer-learning}).
Similarly, the search for architectures under constraints, with multiple objectives, and the automation of model compression has garnered significant attention in the past year (Section~\ref{sec:mo}).
The alignment of architecture search algorithms with other graph structures such as parameter optimization algorithms and activation functions or the search for data augmentation are exciting developments (Section~\ref{sec:other}).
Recently there have also been promising developments along other dimensions that address more complex tasks with image data such as object detection and segmentation~\citep{Zoph2018_Learning,Tan2018_MnasNet,Liu2019_Auto,Weng2019_NASUnet}, as well as works that address safety-critical issues such as the discovery of architectures that are robust against adversarial attacks~\citep{Cubuk2018_Intriguing,Sinn2019_Evolutionary}.
Furthermore, the existing techniques are extended to apply the architecture search for other types of networks.
For instance, there are first attempts to automatically optimize the architecture of autoencoders~\citep{Suganuma2018_Exploiting}, transformers~\citep{So2019_The} and graph CNNs~\citep{Gao2019_GraphNAS}.
In general, these techniques are increasingly being used to automate all components of the data science workflow.
The big remaining challenge is the joint optimization of all configuration parameters of the deep learning workflow.
So far, the individual components have been treated independently with some components still relying on manual configuration.
However, for true automation, it is inevitable that all sensitive parameters must be learned or searched for.
This includes data augmentation, the various search space hyperparameters, the choice of optimizer, the actual architecture choice and possibly even the final model compression.
Moreover, further development of any of these search methods needs to incorporate sample efficiency in its design, an aspect that is currently not handled by neural architecture search.

Given the large amounts of resources dedicated to the research in neural architecture search, it is imperative to question its meaningfulness and usefulness to research.
Although, one might hope to get insights into the understanding of the space of deep learning models or perhaps even the interpretability of deep learning in general, so far there have not been any dedicated efforts channelled into this direction.
Another important question, which we believe is often overlooked, pertains to the credibility of neural architecture search as a practically useful field of research.
It is crucial to emphasize the role of search space designs in the performance of obtained architectures, some of which are heavily inspired by the design of existing architectures.
Many architectures originate from the NASNet search space, which were found by various search algorithms.
Therefore, a legitimate question is to ask whether the search algorithms have really discovered new or innovative architectures, or whether they owe their success to the design and properties of the NASNet search space.
\cite{Li2019_Random} as well as the inventors of the NASNet search space, partially answer this question by showing that even a random search yields good, if not better, results.
Furthermore, these architectures ultimately do not seem to be significantly better than those developed by humans.
The human innovations that contribute to the invention of new architectures are difficult to discover given the restrictions of search spaces.
This brings us to the question of what to expect from the future developments of architecture search algorithms.
If we really seek to find new architectures with innovative elements that advance research in general, then perhaps we have to let go of the restrictive search spaces.
While a general automation of deep learning might answer most of these concerns, perhaps for now we need to ask ourselves why an architecture discovered by a search algorithm is better than the choice of an existing architecture in combination with hyperparameter optimization.
In terms of usefulness, the answer can not be that each dataset needs its own architecture.
Many architecture search papers show that found architectures can be easily transferred to other datasets, even other tasks, which is a well known property of deep learning architectures.
So why then should a practitioner commit to the use of a search algorithm for finding a new architecture, in particular since it does not free her from the optimization of the search space hyperparameters?

The general automation of deep learning is still in its infancy, and many of the aforementioned concerns remain unanswered for the time being. 
However, this remains an exciting domain and the future works will certainly highlight its practical relevance and usefulness.

\vskip 0.2in
\bibliography{references}

\begin{thebibliography}{121}
\providecommand{\natexlab}[1]{#1}
\providecommand{\url}[1]{\texttt{#1}}
\expandafter\ifx\csname urlstyle\endcsname\relax
  \providecommand{\doi}[1]{doi: #1}\else
  \providecommand{\doi}{doi: \begingroup \urlstyle{rm}\Url}\fi

\bibitem[Ashok et~al.(2018)Ashok, Rhinehart, Beainy, and Kitani]{Ashok2018_N2N}
Anubhav Ashok, Nicholas Rhinehart, Fares Beainy, and Kris~M. Kitani.
\newblock {N2N} learning: Network to network compression via policy gradient
  reinforcement learning.
\newblock In \emph{6th International Conference on Learning Representations,
  {ICLR} 2018, Vancouver, BC, Canada, April 30 - May 3, 2018, Conference Track
  Proceedings}, 2018.
\newblock URL \url{https://openreview.net/forum?id=B1hcZZ-AW}.

\bibitem[Auer(2002)]{Auer2002_Using}
Peter Auer.
\newblock Using confidence bounds for exploitation-exploration trade-offs.
\newblock \emph{Journal of Machine Learning Research}, 3:\penalty0 397--422,
  2002.
\newblock URL \url{http://jmlr.org/papers/v3/auer02a.html}.

\bibitem[Baker et~al.(2017)Baker, Gupta, Naik, and Raskar]{Baker2017_Designing}
Bowen Baker, Otkrist Gupta, Nikhil Naik, and Ramesh Raskar.
\newblock Designing neural network architectures using reinforcement learning.
\newblock In \emph{5th International Conference on Learning Representations,
  {ICLR} 2017, Toulon, France, April 24-26, 2017, Conference Track
  Proceedings}, 2017.
\newblock URL \url{https://openreview.net/forum?id=S1c2cvqee}.

\bibitem[Baker et~al.(2018)Baker, Gupta, Raskar, and
  Naik]{Baker2018_Accelerating}
Bowen Baker, Otkrist Gupta, Ramesh Raskar, and Nikhil Naik.
\newblock Accelerating neural architecture search using performance prediction.
\newblock In \emph{6th International Conference on Learning Representations,
  {ICLR} 2018, Vancouver, BC, Canada, April 30 - May 3, 2018, Workshop Track
  Proceedings}, 2018.
\newblock URL \url{https://openreview.net/forum?id=HJqk3N1vG}.

\bibitem[Bauer(1974)]{Bauer1974_Computational}
Friedrich~L Bauer.
\newblock Computational graphs and rounding error.
\newblock \emph{SIAM Journal on Numerical Analysis}, 11\penalty0 (1):\penalty0
  87--96, 1974.

\bibitem[Bello et~al.(2017)Bello, Zoph, Vasudevan, and Le]{Bello2017_Neural}
Irwan Bello, Barret Zoph, Vijay Vasudevan, and Quoc~V. Le.
\newblock Neural optimizer search with reinforcement learning.
\newblock In \emph{Proceedings of the 34th International Conference on Machine
  Learning, {ICML} 2017, Sydney, NSW, Australia, 6-11 August 2017}, pages
  459--468, 2017.
\newblock URL \url{http://proceedings.mlr.press/v70/bello17a.html}.

\bibitem[Bender et~al.(2018)Bender, Kindermans, Zoph, Vasudevan, and
  Le]{Bender2018_Understanding}
Gabriel Bender, Pieter-Jan Kindermans, Barret Zoph, Vijay Vasudevan, and Quoc
  Le.
\newblock Understanding and simplifying one-shot architecture search.
\newblock In Jennifer Dy and Andreas Krause, editors, \emph{Proceedings of the
  35th International Conference on Machine Learning}, volume~80 of
  \emph{Proceedings of Machine Learning Research}, pages 550--559,
  Stockholmsmässan, Stockholm Sweden, 10--15 Jul 2018. PMLR.
\newblock URL \url{http://proceedings.mlr.press/v80/bender18a.html}.

\bibitem[Bergstra and Bengio(2012)]{Bergstra2012_Random}
James Bergstra and Yoshua Bengio.
\newblock Random search for hyper-parameter optimization.
\newblock \emph{Journal of Machine Learning Research}, 13:\penalty0 281--305,
  2012.
\newblock URL \url{http://dl.acm.org/citation.cfm?id=2188395}.

\bibitem[Beyer and Schwefel(2002)]{Beyer2002_Evolution}
Hans{-}Georg Beyer and Hans{-}Paul Schwefel.
\newblock Evolution strategies - {A} comprehensive introduction.
\newblock \emph{Natural Computing}, 1\penalty0 (1):\penalty0 3--52, 2002.
\newblock \doi{10.1023/A:1015059928466}.
\newblock URL \url{https://doi.org/10.1023/A:1015059928466}.

\bibitem[Brock et~al.(2018)Brock, Lim, Ritchie, and Weston]{Brock2018_SMASH}
Andrew Brock, Theodore Lim, James~M. Ritchie, and Nick Weston.
\newblock {SMASH:} one-shot model architecture search through hypernetworks.
\newblock In \emph{6th International Conference on Learning Representations,
  {ICLR} 2018, Vancouver, BC, Canada, April 30 - May 3, 2018, Conference Track
  Proceedings}, 2018.
\newblock URL \url{https://openreview.net/forum?id=rydeCEhs-}.

\bibitem[Cai et~al.(2018{\natexlab{a}})Cai, Chen, Zhang, Yu, and
  Wang]{Cai2018_Efficient}
Han Cai, Tianyao Chen, Weinan Zhang, Yong Yu, and Jun Wang.
\newblock Efficient architecture search by network transformation.
\newblock In \emph{Proceedings of the Thirty-Second {AAAI} Conference on
  Artificial Intelligence, (AAAI-18), the 30th innovative Applications of
  Artificial Intelligence (IAAI-18), and the 8th {AAAI} Symposium on
  Educational Advances in Artificial Intelligence (EAAI-18), New Orleans,
  Louisiana, USA, February 2-7, 2018}, pages 2787--2794, 2018{\natexlab{a}}.
\newblock URL
  \url{https://www.aaai.org/ocs/index.php/AAAI/AAAI18/paper/view/16755}.

\bibitem[Cai et~al.(2018{\natexlab{b}})Cai, Yang, Zhang, Han, and
  Yu]{Cai2018_Path}
Han Cai, Jiacheng Yang, Weinan Zhang, Song Han, and Yong Yu.
\newblock Path-level network transformation for efficient architecture search.
\newblock In \emph{Proceedings of the 35th International Conference on Machine
  Learning, {ICML} 2018, Stockholmsm{\"{a}}ssan, Stockholm, Sweden, July 10-15,
  2018}, pages 677--686, 2018{\natexlab{b}}.
\newblock URL \url{http://proceedings.mlr.press/v80/cai18a.html}.

\bibitem[Cai et~al.(2019)Cai, Zhu, and Han]{Cai2019_Proxyless}
Han Cai, Ligeng Zhu, and Song Han.
\newblock Proxyless{NAS}: Direct neural architecture search on target task and
  hardware.
\newblock In \emph{Proceedings of the International Conference on Learning
  Representations, {ICLR} 2019, New Orleans, Louisiana, USA}, 2019.
\newblock URL \url{https://openreview.net/forum?id=HylVB3AqYm}.

\bibitem[Cao et~al.(2019)Cao, Wang, and Kitani]{Cao2019_Learnable}
Shengcao Cao, Xiaofang Wang, and Kris~M. Kitani.
\newblock Learnable embedding space for efficient neural architecture
  compression.
\newblock In \emph{Proceedings of the International Conference on Learning
  Representations, {ICLR} 2019, New Orleans, Louisiana, USA}, 2019.
\newblock URL \url{https://openreview.net/forum?id=S1xLN3C9YX}.

\bibitem[Casale et~al.(2019)Casale, Gordon, and Fusi]{Casale2019_ProbNAS}
Francesco~Paolo Casale, Jonathan Gordon, and Nicolo Fusi.
\newblock Probabilistic neural architecture search.
\newblock \emph{CoRR}, abs/1902.05116, 2019.
\newblock URL \url{http://arxiv.org/abs/1902.05116}.

\bibitem[Chandrashekaran and Lane(2017)]{Chandrashekaran2017_Speeding}
Akshay Chandrashekaran and Ian~R. Lane.
\newblock Speeding up hyper-parameter optimization by extrapolation of learning
  curves using previous builds.
\newblock In \emph{Machine Learning and Knowledge Discovery in Databases -
  European Conference, {ECML} {PKDD} 2017, Skopje, Macedonia, September 18-22,
  2017, Proceedings, Part {I}}, pages 477--492, 2017.
\newblock \doi{10.1007/978-3-319-71249-9\_29}.
\newblock URL \url{https://doi.org/10.1007/978-3-319-71249-9\_29}.

\bibitem[Chen et~al.(2016)Chen, Goodfellow, and Shlens]{Chen2016_Net2Net}
Tianqi Chen, Ian~J. Goodfellow, and Jonathon Shlens.
\newblock Net2net: Accelerating learning via knowledge transfer.
\newblock In Yoshua Bengio and Yann LeCun, editors, \emph{4th International
  Conference on Learning Representations, {ICLR} 2016, San Juan, Puerto Rico,
  May 2-4, 2016, Conference Track Proceedings}, 2016.
\newblock URL \url{http://arxiv.org/abs/1511.05641}.

\bibitem[Chen et~al.(2017)Chen, Li, Xiao, Jin, Yan, and Feng]{Chen2017_Dual}
Yunpeng Chen, Jianan Li, Huaxin Xiao, Xiaojie Jin, Shuicheng Yan, and Jiashi
  Feng.
\newblock Dual path networks.
\newblock In \emph{Advances in Neural Information Processing Systems 30: Annual
  Conference on Neural Information Processing Systems 2017, 4-9 December 2017,
  Long Beach, CA, {USA}}, pages 4470--4478, 2017.
\newblock URL \url{http://papers.nips.cc/paper/7033-dual-path-networks}.

\bibitem[Courbariaux et~al.(2015)Courbariaux, Bengio, and
  David]{Courbariaux2015_BinaryConnect}
Matthieu Courbariaux, Yoshua Bengio, and Jean{-}Pierre David.
\newblock Binaryconnect: Training deep neural networks with binary weights
  during propagations.
\newblock In \emph{Advances in Neural Information Processing Systems 28: Annual
  Conference on Neural Information Processing Systems 2015, December 7-12,
  2015, Montreal, Quebec, Canada}, pages 3123--3131, 2015.
\newblock URL
  \url{http://papers.nips.cc/paper/5647-binaryconnect-training-deep-neural-networks-with-binary-weights-during-propagations}.

\bibitem[Cubuk et~al.(2018{\natexlab{a}})Cubuk, Zoph, Man{\'{e}}, Vasudevan,
  and Le]{Cubuk2018_AutoAugment}
Ekin~Dogus Cubuk, Barret Zoph, Dandelion Man{\'{e}}, Vijay Vasudevan, and
  Quoc~V. Le.
\newblock Autoaugment: Learning augmentation policies from data.
\newblock \emph{CoRR}, abs/1805.09501, 2018{\natexlab{a}}.
\newblock URL \url{http://arxiv.org/abs/1805.09501}.

\bibitem[Cubuk et~al.(2018{\natexlab{b}})Cubuk, Zoph, Schoenholz, and
  Le]{Cubuk2018_Intriguing}
Ekin~Dogus Cubuk, Barret Zoph, Samuel~S. Schoenholz, and Quoc~V. Le.
\newblock Intriguing properties of adversarial examples.
\newblock In \emph{6th International Conference on Learning Representations,
  {ICLR} 2018, Vancouver, BC, Canada, April 30 - May 3, 2018, Workshop Track
  Proceedings}, 2018{\natexlab{b}}.
\newblock URL \url{https://openreview.net/forum?id=Skz1zaRLz}.

\bibitem[Deb(2014)]{Deb2014_Multi}
Kalyanmoy Deb.
\newblock Multi-objective optimization.
\newblock In \emph{Search methodologies}, pages 403--449. Springer, 2014.

\bibitem[Deb et~al.(2002)Deb, Agrawal, Pratap, and Meyarivan]{Deb2002_NSGA}
Kalyanmoy Deb, Samir Agrawal, Amrit Pratap, and T.~Meyarivan.
\newblock A fast and elitist multiobjective genetic algorithm: {NSGA-II}.
\newblock \emph{{IEEE} Trans. Evolutionary Computation}, 6\penalty0
  (2):\penalty0 182--197, 2002.
\newblock \doi{10.1109/4235.996017}.
\newblock URL \url{https://doi.org/10.1109/4235.996017}.

\bibitem[DeVries and Taylor(2017)]{DeVries2018_Improved}
Terrance DeVries and Graham~W. Taylor.
\newblock Improved regularization of convolutional neural networks with cutout.
\newblock \emph{CoRR}, abs/1708.04552, 2017.
\newblock URL \url{http://arxiv.org/abs/1708.04552}.

\bibitem[Domhan et~al.(2015)Domhan, Springenberg, and
  Hutter]{Domhan2015_Speeding}
Tobias Domhan, Jost~Tobias Springenberg, and Frank Hutter.
\newblock Speeding up automatic hyperparameter optimization of deep neural
  networks by extrapolation of learning curves.
\newblock In \emph{Proceedings of the Twenty-Fourth International Joint
  Conference on Artificial Intelligence, {IJCAI} 2015, Buenos Aires, Argentina,
  July 25-31, 2015}, pages 3460--3468, 2015.
\newblock URL \url{http://ijcai.org/Abstract/15/487}.

\bibitem[Dong et~al.(2018)Dong, Cheng, Juan, Wei, and Sun]{Dong2018_DPP}
Jin{-}Dong Dong, An{-}Chieh Cheng, Da{-}Cheng Juan, Wei Wei, and Min Sun.
\newblock Dpp-net: Device-aware progressive search for pareto-optimal neural
  architectures.
\newblock In Vittorio Ferrari, Martial Hebert, Cristian Sminchisescu, and Yair
  Weiss, editors, \emph{Computer Vision - {ECCV} 2018 - 15th European
  Conference, Munich, Germany, September 8-14, 2018, Proceedings, Part {XI}},
  volume 11215 of \emph{Lecture Notes in Computer Science}, pages 540--555.
  Springer, 2018.
\newblock \doi{10.1007/978-3-030-01252-6\_32}.
\newblock URL \url{https://doi.org/10.1007/978-3-030-01252-6\_32}.

\bibitem[Elsken et~al.(2018)Elsken, Metzen, and Hutter]{Elsken2018_Simple}
Thomas Elsken, Jan~Hendrik Metzen, and Frank Hutter.
\newblock Simple and efficient architecture search for convolutional neural
  networks.
\newblock In \emph{6th International Conference on Learning Representations,
  {ICLR} 2018, Vancouver, BC, Canada, April 30 - May 3, 2018, Workshop Track
  Proceedings}, 2018.
\newblock URL \url{https://openreview.net/forum?id=H1hymrkDf}.

\bibitem[Elsken et~al.(2019)Elsken, Metzen, and Hutter]{Elsken2019_Efficient}
Thomas Elsken, Jan~Hendrik Metzen, and Frank Hutter.
\newblock Efficient multi-objective neural architecture search via lamarckian
  evolution.
\newblock In \emph{Proceedings of the International Conference on Learning
  Representations, {ICLR} 2019, New Orleans, Louisiana, USA}, 2019.
\newblock URL \url{https://openreview.net/forum?id=ByME42AqK7}.

\bibitem[Fiacco and McCormick(1990)]{Fiacco1990_Nonlinear}
Anthony~V Fiacco and Garth~P McCormick.
\newblock \emph{Nonlinear programming: sequential unconstrained minimization
  techniques}, volume~4.
\newblock Siam, 1990.

\bibitem[Floreano et~al.(2008)Floreano, D{\"{u}}rr, and
  Mattiussi]{Floreano2008_Neuroevolution}
Dario Floreano, Peter D{\"{u}}rr, and Claudio Mattiussi.
\newblock Neuroevolution: from architectures to learning.
\newblock \emph{Evolutionary Intelligence}, 1\penalty0 (1):\penalty0 47--62,
  2008.
\newblock URL \url{https://doi.org/10.1007/s12065-007-0002-4}.

\bibitem[Gao et~al.(2019)Gao, Yang, Zhang, Zhou, and Hu]{Gao2019_GraphNAS}
Yang Gao, Hong Yang, Peng Zhang, Chuan Zhou, and Yue Hu.
\newblock Graphnas: Graph neural architecture search with reinforcement
  learning.
\newblock \emph{CoRR}, abs/1904.09981, 2019.
\newblock URL \url{http://arxiv.org/abs/1904.09981}.

\bibitem[Gastaldi(2017)]{Gastaldi2017_Shake}
Xavier Gastaldi.
\newblock Shake-shake regularization.
\newblock \emph{CoRR}, abs/1705.07485, 2017.
\newblock URL \url{http://arxiv.org/abs/1705.07485}.

\bibitem[Goldberg and Deb(1990)]{Goldberg1990_Tournament_Selection}
David~E. Goldberg and Kalyanmoy Deb.
\newblock A comparative analysis of selection schemes used in genetic
  algorithms.
\newblock In Gregory J.~E. Rawlins, editor, \emph{Proceedings of the First
  Workshop on Foundations of Genetic Algorithms. Bloomington Campus, Indiana,
  USA, July 15-18 1990.}, pages 69--93. Morgan Kaufmann, 1990.

\bibitem[Goodfellow et~al.(2016)Goodfellow, Bengio, and
  Courville]{Goodfellow2016_Deep}
Ian Goodfellow, Yoshua Bengio, and Aaron Courville.
\newblock \emph{Deep Learning}.
\newblock MIT Press, 2016.
\newblock \url{http://www.deeplearningbook.org}.

\bibitem[Ha et~al.(2017)Ha, Dai, and Le]{Ha2017_HyperNetworks}
David Ha, Andrew~M. Dai, and Quoc~V. Le.
\newblock Hypernetworks.
\newblock In \emph{5th International Conference on Learning Representations,
  {ICLR} 2017, Toulon, France, April 24-26, 2017, Conference Track
  Proceedings}, 2017.
\newblock URL \url{https://openreview.net/forum?id=rkpACe1lx}.

\bibitem[He et~al.(2018)He, Lin, Liu, Wang, Li, and Han]{He2018_AMC}
Yihui He, Ji~Lin, Zhijian Liu, Hanrui Wang, Li{-}Jia Li, and Song Han.
\newblock {AMC:} automl for model compression and acceleration on mobile
  devices.
\newblock In \emph{Computer Vision - {ECCV} 2018 - 15th European Conference,
  Munich, Germany, September 8-14, 2018, Proceedings, Part {VII}}, pages
  815--832, 2018.
\newblock \doi{10.1007/978-3-030-01234-2\_48}.
\newblock URL \url{https://doi.org/10.1007/978-3-030-01234-2\_48}.

\bibitem[Hinton et~al.(2015)Hinton, Vinyals, and Dean]{Hinton2015_Distilling}
Geoffrey~E. Hinton, Oriol Vinyals, and Jeffrey Dean.
\newblock Distilling the knowledge in a neural network.
\newblock \emph{CoRR}, abs/1503.02531, 2015.
\newblock URL \url{http://arxiv.org/abs/1503.02531}.

\bibitem[Hinz et~al.(2018)Hinz, Navarro{-}Guerrero, Magg, and
  Wermter]{Hinz2018_Speeding}
Tobias Hinz, Nicol{\'{a}}s Navarro{-}Guerrero, Sven Magg, and Stefan Wermter.
\newblock Speeding up the hyperparameter optimization of deep convolutional
  neural networks.
\newblock \emph{International Journal of Computational Intelligence and
  Applications}, 17\penalty0 (2):\penalty0 1850008, 2018.
\newblock \doi{10.1142/S1469026818500086}.
\newblock URL \url{https://doi.org/10.1142/S1469026818500086}.

\bibitem[Hochreiter and Schmidhuber(1997)]{Hochreiter1997_LSTM}
Sepp Hochreiter and J{\"{u}}rgen Schmidhuber.
\newblock Long short-term memory.
\newblock \emph{Neural Computation}, 9\penalty0 (8):\penalty0 1735--1780, 1997.
\newblock \doi{10.1162/neco.1997.9.8.1735}.
\newblock URL \url{https://doi.org/10.1162/neco.1997.9.8.1735}.

\bibitem[Howard et~al.(2017)Howard, Zhu, Chen, Kalenichenko, Wang, Weyand,
  Andreetto, and Adam]{Howard2017_Mobilenets}
Andrew~G. Howard, Menglong Zhu, Bo~Chen, Dmitry Kalenichenko, Weijun Wang,
  Tobias Weyand, Marco Andreetto, and Hartwig Adam.
\newblock Mobilenets: Efficient convolutional neural networks for mobile vision
  applications.
\newblock \emph{CoRR}, abs/1704.04861, 2017.
\newblock URL \url{http://arxiv.org/abs/1704.04861}.

\bibitem[Hsu et~al.(2018)Hsu, Chang, Juan, Pan, Chen, Wei, and
  Chang]{Hsu2018_MONAS}
Chi{-}Hung Hsu, Shu{-}Huan Chang, Da{-}Cheng Juan, Jia{-}Yu Pan, Yu{-}Ting
  Chen, Wei Wei, and Shih{-}Chieh Chang.
\newblock {MONAS:} multi-objective neural architecture search using
  reinforcement learning.
\newblock \emph{CoRR}, abs/1806.10332, 2018.

\bibitem[Hu et~al.(2018)Hu, Langford, Caruana, Horvitz, and Dey]{Hu2018_Macro}
Hanzhang Hu, John Langford, Rich Caruana, Eric Horvitz, and Debadeepta Dey.
\newblock Macro neural architecture search revisited.
\newblock In \emph{Workshop on Meta-Learning at NeurIPS 2018, MetaLearn 2018,
  3-8 December 2018, Montr{\'{e}}al, Canada.}, 2018.
\newblock URL
  \url{http://metalearning.ml/2018/papers/metalearn2018_paper16.pdf}.

\bibitem[Huang et~al.(2017)Huang, Liu, van~der Maaten, and
  Weinberger]{Huang2017_Densely}
Gao Huang, Zhuang Liu, Laurens van~der Maaten, and Kilian~Q. Weinberger.
\newblock Densely connected convolutional networks.
\newblock In \emph{2017 {IEEE} Conference on Computer Vision and Pattern
  Recognition, {CVPR} 2017, Honolulu, HI, USA, July 21-26, 2017}, pages
  2261--2269. {IEEE} Computer Society, 2017.
\newblock \doi{10.1109/CVPR.2017.243}.
\newblock URL \url{https://doi.org/10.1109/CVPR.2017.243}.

\bibitem[Huang et~al.(2018{\natexlab{a}})Huang, Liu, van~der Maaten, and
  Weinberger]{Huang2018_CondenseNet}
Gao Huang, Shichen Liu, Laurens van~der Maaten, and Kilian~Q. Weinberger.
\newblock Condensenet: An efficient densenet using learned group convolutions.
\newblock In \emph{2018 {IEEE} Conference on Computer Vision and Pattern
  Recognition, {CVPR} 2018, Salt Lake City, UT, USA, June 18-22, 2018}, pages
  2752--2761. {IEEE} Computer Society, 2018{\natexlab{a}}.
\newblock \doi{10.1109/CVPR.2018.00291}.
\newblock URL
  \url{http://openaccess.thecvf.com/content\_cvpr\_2018/html/Huang\_CondenseNet\_An\_Efficient\_CVPR\_2018\_paper.html}.

\bibitem[Huang et~al.(2018{\natexlab{b}})Huang, Cheng, Chen, Lee, Ngiam, Le,
  and Chen]{Huang2018_GPipe}
Yanping Huang, Yonglong Cheng, Dehao Chen, HyoukJoong Lee, Jiquan Ngiam,
  Quoc~V. Le, and Zhifeng Chen.
\newblock Gpipe: Efficient training of giant neural networks using pipeline
  parallelism.
\newblock \emph{CoRR}, abs/1811.06965, 2018{\natexlab{b}}.
\newblock URL \url{http://arxiv.org/abs/1811.06965}.

\bibitem[Hwang and Masud(2012)]{Hwang2012_Multiple}
C-L Hwang and Abu Syed~Md Masud.
\newblock \emph{Multiple objective decision making—methods and applications:
  a state-of-the-art survey}, volume 164.
\newblock Springer Science \& Business Media, 2012.

\bibitem[Istrate et~al.(2019)Istrate, Scheidegger, Mariani, Nikolopoulos,
  Bekas, and Malossi]{Istrate2019_TAPAS}
Roxana Istrate, Florian Scheidegger, Giovanni Mariani, Dimitrios~S.
  Nikolopoulos, Costas Bekas, and A.~Cristiano~I. Malossi.
\newblock {TAPAS:} train-less accuracy predictor for architecture search.
\newblock In \emph{Proceedings of the Thirty-Third {AAAI} Conference on
  Artificial Intelligence, (AAAI-19), Honolulu, Hawaii, USA}, 2019.

\bibitem[Jones(2001)]{Jones2001_A}
Donald~R. Jones.
\newblock A taxonomy of global optimization methods based on response surfaces.
\newblock \emph{J. Global Optimization}, 21\penalty0 (4):\penalty0 345--383,
  2001.
\newblock \doi{10.1023/A:1012771025575}.
\newblock URL \url{https://doi.org/10.1023/A:1012771025575}.

\bibitem[Jong(2006)]{DeJong2006_Evolutionary}
Kenneth A.~De Jong.
\newblock \emph{Evolutionary computation - a unified approach}.
\newblock {MIT} Press, 2006.
\newblock ISBN 978-0-262-04194-2.

\bibitem[Kandasamy et~al.(2018)Kandasamy, Neiswanger, Schneider, P{\'{o}}czos,
  and Xing]{Kandasamy2018_Neural}
Kirthevasan Kandasamy, Willie Neiswanger, Jeff Schneider, Barnab{\'{a}}s
  P{\'{o}}czos, and Eric~P. Xing.
\newblock Neural architecture search with bayesian optimisation and optimal
  transport.
\newblock In \emph{Advances in Neural Information Processing Systems 31: Annual
  Conference on Neural Information Processing Systems 2018, NeurIPS 2018, 3-8
  December 2018, Montr{\'{e}}al, Canada.}, pages 2020--2029, 2018.
\newblock URL
  \url{http://papers.nips.cc/paper/7472-neural-architecture-search-with-bayesian-optimisation-and-optimal-transport}.

\bibitem[Kim et~al.(2017)Kim, Reddy, Yun, and Seo]{Kim2017_NEMO}
Ye-Hoon Kim, Bhargava Reddy, Sojung Yun, and Chanwon Seo.
\newblock Nemo : Neuro-evolution with multiobjective optimization of deep
  neural network for speed and accuracy.
\newblock In \emph{AutoML Workshop at ICML 2017}, 2017.

\bibitem[Klein et~al.(2017)Klein, Falkner, Springenberg, and
  Hutter]{Klein2017_Learning}
Aaron Klein, Stefan Falkner, Jost~Tobias Springenberg, and Frank Hutter.
\newblock Learning curve prediction with bayesian neural networks.
\newblock In \emph{5th International Conference on Learning Representations,
  {ICLR} 2017, Toulon, France, April 24-26, 2017, Conference Track
  Proceedings}, 2017.
\newblock URL \url{https://openreview.net/forum?id=S11KBYclx}.

\bibitem[Kocsis and Szepesv{\'{a}}ri(2006)]{Kocsis2006_Bandit}
Levente Kocsis and Csaba Szepesv{\'{a}}ri.
\newblock Bandit based monte-carlo planning.
\newblock In Johannes F{\"{u}}rnkranz, Tobias Scheffer, and Myra Spiliopoulou,
  editors, \emph{Machine Learning: {ECML} 2006, 17th European Conference on
  Machine Learning, Berlin, Germany, September 18-22, 2006, Proceedings},
  volume 4212 of \emph{Lecture Notes in Computer Science}, pages 282--293.
  Springer, 2006.
\newblock URL \url{https://doi.org/10.1007/11871842\_29}.

\bibitem[Kokiopoulou et~al.(2019)Kokiopoulou, Hauth, Sbaiz, Gesmundo, Bartok,
  and Berent]{Kokiopoulou2019_Fast}
Efi Kokiopoulou, Anja Hauth, Luciano Sbaiz, Andrea Gesmundo, Gabor Bartok, and
  Jesse Berent.
\newblock Fast task-aware architecture inference.
\newblock \emph{CoRR}, abs/1902.05781, 2019.
\newblock URL \url{http://arxiv.org/abs/1902.05781}.

\bibitem[Krizhevsky(2009)]{Krizhevsky_CIFAR10}
Alex Krizhevsky.
\newblock Learning multiple layers of features from tiny images.
\newblock Technical report, Canadian Institute for Advanced Research, 2009.

\bibitem[Krizhevsky et~al.(2012)Krizhevsky, Sutskever, and
  Hinton]{Krizhevsky2012_Imagenet}
Alex Krizhevsky, Ilya Sutskever, and Geoffrey~E. Hinton.
\newblock Imagenet classification with deep convolutional neural networks.
\newblock In Peter~L. Bartlett, Fernando C.~N. Pereira, Christopher J.~C.
  Burges, L{\'{e}}on Bottou, and Kilian~Q. Weinberger, editors, \emph{Advances
  in Neural Information Processing Systems 25: 26th Annual Conference on Neural
  Information Processing Systems 2012. Proceedings of a meeting held December
  3-6, 2012, Lake Tahoe, Nevada, United States.}, pages 1106--1114, 2012.
\newblock URL
  \url{http://papers.nips.cc/paper/4824-imagenet-classification-with-deep-convolutional-neural-networks}.

\bibitem[LeCun et~al.(1998)LeCun, Bottou, Bengio, and
  Haffner]{Lecun1998_Gradient}
Yann LeCun, L{\'e}on Bottou, Yoshua Bengio, and Patrick Haffner.
\newblock Gradient-based learning applied to document recognition.
\newblock \emph{Proceedings of the IEEE}, 86\penalty0 (11):\penalty0
  2278--2324, 1998.
\newblock ISSN 0018-9219.
\newblock \doi{10.1109/5.726791}.

\bibitem[Li and Talwalkar(2019)]{Li2019_Random}
Liam Li and Ameet Talwalkar.
\newblock Random search and reproducibility for neural architecture search.
\newblock \emph{CoRR}, abs/1902.07638, 2019.
\newblock URL \url{http://arxiv.org/abs/1902.07638}.

\bibitem[Li et~al.(2017)Li, Jamieson, DeSalvo, Rostamizadeh, and
  Talwalkar]{Li2017_Hyperband}
Lisha Li, Kevin~G. Jamieson, Giulia DeSalvo, Afshin Rostamizadeh, and Ameet
  Talwalkar.
\newblock Hyperband: {A} novel bandit-based approach to hyperparameter
  optimization.
\newblock \emph{Journal of Machine Learning Research}, 18:\penalty0
  185:1--185:52, 2017.
\newblock URL \url{http://jmlr.org/papers/v18/16-558.html}.

\bibitem[Lillicrap et~al.(2016)Lillicrap, Hunt, Pritzel, Heess, Erez, Tassa,
  Silver, and Wierstra]{Lillicrap2016_Continous}
Timothy~P. Lillicrap, Jonathan~J. Hunt, Alexander Pritzel, Nicolas Heess, Tom
  Erez, Yuval Tassa, David Silver, and Daan Wierstra.
\newblock Continuous control with deep reinforcement learning.
\newblock In Yoshua Bengio and Yann LeCun, editors, \emph{4th International
  Conference on Learning Representations, {ICLR} 2016, San Juan, Puerto Rico,
  May 2-4, 2016, Conference Track Proceedings}, 2016.
\newblock URL \url{http://arxiv.org/abs/1509.02971}.

\bibitem[Liu et~al.(2018{\natexlab{a}})Liu, Zoph, Neumann, Shlens, Hua, Li,
  Fei{-}Fei, Yuille, Huang, and Murphy]{Liu2018_Progressive}
Chenxi Liu, Barret Zoph, Maxim Neumann, Jonathon Shlens, Wei Hua, Li{-}Jia Li,
  Li~Fei{-}Fei, Alan~L. Yuille, Jonathan Huang, and Kevin Murphy.
\newblock Progressive neural architecture search.
\newblock In \emph{Computer Vision - {ECCV} 2018 - 15th European Conference,
  Munich, Germany, September 8-14, 2018, Proceedings, Part {I}}, pages 19--35,
  2018{\natexlab{a}}.
\newblock \doi{10.1007/978-3-030-01246-5\_2}.
\newblock URL \url{https://doi.org/10.1007/978-3-030-01246-5\_2}.

\bibitem[Liu et~al.(2019{\natexlab{a}})Liu, Chen, Schroff, Adam, Hua, Yuille,
  and Fei{-}Fei]{Liu2019_Auto}
Chenxi Liu, Liang{-}Chieh Chen, Florian Schroff, Hartwig Adam, Wei Hua, Alan~L.
  Yuille, and Li~Fei{-}Fei.
\newblock Auto-deeplab: Hierarchical neural architecture search for semantic
  image segmentation.
\newblock \emph{CoRR}, abs/1901.02985, 2019{\natexlab{a}}.
\newblock URL \url{http://arxiv.org/abs/1901.02985}.

\bibitem[Liu et~al.(2018{\natexlab{b}})Liu, Simonyan, Vinyals, Fernando, and
  Kavukcuoglu]{Liu2018_Hierarchical}
Hanxiao Liu, Karen Simonyan, Oriol Vinyals, Chrisantha Fernando, and Koray
  Kavukcuoglu.
\newblock Hierarchical representations for efficient architecture search.
\newblock In \emph{6th International Conference on Learning Representations,
  {ICLR} 2018, Vancouver, BC, Canada, April 30 - May 3, 2018, Conference Track
  Proceedings}, 2018{\natexlab{b}}.
\newblock URL \url{https://openreview.net/forum?id=BJQRKzbA-}.

\bibitem[Liu et~al.(2019{\natexlab{b}})Liu, Simonyan, and Yang]{Liu2018_DARTS}
Hanxiao Liu, Karen Simonyan, and Yiming Yang.
\newblock {DARTS:} differentiable architecture search.
\newblock In \emph{Proceedings of the International Conference on Learning
  Representations, {ICLR} 2019, New Orleans, Louisiana, USA},
  2019{\natexlab{b}}.
\newblock URL \url{https://openreview.net/forum?id=S1eYHoC5FX}.

\bibitem[Lu et~al.(2018)Lu, Whalen, Boddeti, Dhebar, Deb, Goodman, and
  Banzhaf]{Lu2018_NSGA_NET}
Zhichao Lu, Ian Whalen, Vishnu Boddeti, Yashesh~D. Dhebar, Kalyanmoy Deb,
  Erik~D. Goodman, and Wolfgang Banzhaf.
\newblock {NSGA-NET:} {A} multi-objective genetic algorithm for neural
  architecture search.
\newblock \emph{CoRR}, abs/1810.03522, 2018.
\newblock URL \url{http://arxiv.org/abs/1810.03522}.

\bibitem[Luo et~al.(2018)Luo, Tian, Qin, Chen, and Liu]{Luo2018_NAO}
Renqian Luo, Fei Tian, Tao Qin, Enhong Chen, and Tie{-}Yan Liu.
\newblock Neural architecture optimization.
\newblock In \emph{Advances in Neural Information Processing Systems 31: Annual
  Conference on Neural Information Processing Systems 2018, NeurIPS 2018, 3-8
  December 2018, Montr{\'{e}}al, Canada.}, pages 7827--7838, 2018.
\newblock URL
  \url{http://papers.nips.cc/paper/8007-neural-architecture-optimization}.

\bibitem[Maddison et~al.(2017)Maddison, Mnih, and Teh]{Maddison2017_The}
Chris~J. Maddison, Andriy Mnih, and Yee~Whye Teh.
\newblock The concrete distribution: {A} continuous relaxation of discrete
  random variables.
\newblock In \emph{5th International Conference on Learning Representations,
  {ICLR} 2017, Toulon, France, April 24-26, 2017, Conference Track
  Proceedings}, 2017.
\newblock URL \url{https://openreview.net/forum?id=S1jE5L5gl}.

\bibitem[Mockus et~al.(1978)Mockus, Tie\v{s}is, and \v{Z}ilinskas]{Mockus1978}
Jonas Mockus, Vytautas Tie\v{s}is, and Antanas \v{Z}ilinskas.
\newblock The application of bayesian methods for seeking the extremum.
\newblock \emph{Towards Global Optimization}, 2:\penalty0 117--129, 1978.

\bibitem[Negrinho and Gordon(2017)]{Negrinho2017_DeepArchitect}
Renato Negrinho and Geoffrey~J. Gordon.
\newblock Deeparchitect: Automatically designing and training deep
  architectures.
\newblock \emph{CoRR}, abs/1704.08792, 2017.
\newblock URL \url{http://arxiv.org/abs/1704.08792}.

\bibitem[Pfahringer et~al.(2000)Pfahringer, Bensusan, and
  Giraud{-}Carrier]{Pfahringer2000_Meta}
Bernhard Pfahringer, Hilan Bensusan, and Christophe~G. Giraud{-}Carrier.
\newblock Meta-learning by landmarking various learning algorithms.
\newblock In \emph{Proceedings of the Seventeenth International Conference on
  Machine Learning {(ICML} 2000), Stanford University, Stanford, CA, USA, June
  29 - July 2, 2000}, pages 743--750, 2000.

\bibitem[Pham et~al.(2018)Pham, Guan, Zoph, Le, and Dean]{Pham2018_ENAS}
Hieu Pham, Melody Guan, Barret Zoph, Quoc Le, and Jeff Dean.
\newblock Efficient neural architecture search via parameters sharing.
\newblock In Jennifer Dy and Andreas Krause, editors, \emph{Proceedings of the
  35th International Conference on Machine Learning}, volume~80 of
  \emph{Proceedings of Machine Learning Research}, pages 4095--4104,
  Stockholmsmässan, Stockholm Sweden, 10--15 Jul 2018. PMLR.
\newblock URL \url{http://proceedings.mlr.press/v80/pham18a.html}.

\bibitem[Ramachandran et~al.(2018)Ramachandran, Zoph, and
  Le]{Ramachandran2018_Searching}
Prajit Ramachandran, Barret Zoph, and Quoc~V. Le.
\newblock Searching for activation functions.
\newblock In \emph{6th International Conference on Learning Representations,
  {ICLR} 2018, Vancouver, BC, Canada, April 30 - May 3, 2018, Workshop Track
  Proceedings}, 2018.
\newblock URL \url{https://openreview.net/forum?id=Hkuq2EkPf}.

\bibitem[Rasmussen and Williams(2006)]{Rasmussen2006_Gaussian}
Carl~Edward Rasmussen and Christopher K.~I. Williams.
\newblock \emph{Gaussian processes for machine learning}.
\newblock Adaptive computation and machine learning. {MIT} Press, 2006.
\newblock ISBN 026218253X.
\newblock URL \url{http://www.worldcat.org/oclc/61285753}.

\bibitem[Real et~al.(2017)Real, Moore, Selle, Saxena, Suematsu, Tan, Le, and
  Kurakin]{Real2017_Large}
Esteban Real, Sherry Moore, Andrew Selle, Saurabh Saxena, Yutaka~Leon Suematsu,
  Jie Tan, Quoc~V. Le, and Alexey Kurakin.
\newblock Large-scale evolution of image classifiers.
\newblock In Doina Precup and Yee~Whye Teh, editors, \emph{Proceedings of the
  34th International Conference on Machine Learning}, volume~70 of
  \emph{Proceedings of Machine Learning Research}, pages 2902--2911,
  International Convention Centre, Sydney, Australia, 06--11 Aug 2017. PMLR.
\newblock URL \url{http://proceedings.mlr.press/v70/real17a.html}.

\bibitem[Real et~al.(2019)Real, Aggarwal, Huang, and Le]{Real2019_Aging}
Esteban Real, Alok Aggarwal, Yanping Huang, and Quoc~V. Le.
\newblock Aging evolution for image classifier architecture search.
\newblock In \emph{Proceedings of the Thirty-Third {AAAI} Conference on
  Artificial Intelligence, (AAAI-19), Honolulu, Hawaii, USA}, 2019.

\bibitem[Reif et~al.(2014)Reif, Shafait, Goldstein, Breuel, and
  Dengel]{Reif2014_Automatic}
Matthias Reif, Faisal Shafait, Markus Goldstein, Thomas~M. Breuel, and Andreas
  Dengel.
\newblock Automatic classifier selection for non-experts.
\newblock \emph{Pattern Anal. Appl.}, 17\penalty0 (1):\penalty0 83--96, 2014.
\newblock \doi{10.1007/s10044-012-0280-z}.
\newblock URL \url{https://doi.org/10.1007/s10044-012-0280-z}.

\bibitem[Rummery and Niranjan(1994)]{Rummery1994_online}
G.~A. Rummery and M.~Niranjan.
\newblock On-line q-learning using connectionist systems.
\newblock Technical report, 1994.

\bibitem[Russakovsky et~al.(2015)Russakovsky, Deng, Su, Krause, Satheesh, Ma,
  Huang, Karpathy, Khosla, Bernstein, Berg, and Li]{Russakovsky2015_ImageNet}
Olga Russakovsky, Jia Deng, Hao Su, Jonathan Krause, Sanjeev Satheesh, Sean Ma,
  Zhiheng Huang, Andrej Karpathy, Aditya Khosla, Michael~S. Bernstein,
  Alexander~C. Berg, and Fei{-}Fei Li.
\newblock Imagenet large scale visual recognition challenge.
\newblock \emph{International Journal of Computer Vision}, 115\penalty0
  (3):\penalty0 211--252, 2015.
\newblock \doi{10.1007/s11263-015-0816-y}.
\newblock URL \url{https://doi.org/10.1007/s11263-015-0816-y}.

\bibitem[Sabharwal et~al.(2016)Sabharwal, Samulowitz, and
  Tesauro]{Sabharwal2016_Selecting}
Ashish Sabharwal, Horst Samulowitz, and Gerald Tesauro.
\newblock Selecting near-optimal learners via incremental data allocation.
\newblock In \emph{Proceedings of the Thirtieth {AAAI} Conference on Artificial
  Intelligence, February 12-17, 2016, Phoenix, Arizona, {USA.}}, pages
  2007--2015, 2016.
\newblock URL
  \url{http://www.aaai.org/ocs/index.php/AAAI/AAAI16/paper/view/12524}.

\bibitem[{Santiago Pineda} et~al.(2014){Santiago Pineda}, {Fraire Huacuja},
  Dorronsoro, Pecero, Santill{\'{a}}n, Barbosa, and {Soto
  Monterrubio}]{Pineda2014_A}
Alejandro {Santiago Pineda}, Héctor~Joaquín {Fraire Huacuja}, Bernab{\'{e}}
  Dorronsoro, Johnatan~E. Pecero, Claudia~G{\'{o}}mez Santill{\'{a}}n, Juan
  Javier~Gonz{\'{a}}lez Barbosa, and Jos{'{e}}~Carlos {Soto Monterrubio}.
\newblock A survey of decomposition methods for multi-objective optimization.
\newblock In \emph{Recent Advances on Hybrid Approaches for Designing
  Intelligent Systems}, pages 453--465. Springer, 2014.
\newblock \doi{10.1007/978-3-319-05170-3\_31}.
\newblock URL \url{https://doi.org/10.1007/978-3-319-05170-3\_31}.

\bibitem[Saxena and Verbeek(2016)]{Saxena2016_Convolutional}
Shreyas Saxena and Jakob Verbeek.
\newblock Convolutional neural fabrics.
\newblock In \emph{Advances in Neural Information Processing Systems 29: Annual
  Conference on Neural Information Processing Systems 2016, December 5-10,
  2016, Barcelona, Spain}, pages 4053--4061, 2016.
\newblock URL
  \url{http://papers.nips.cc/paper/6304-convolutional-neural-fabrics}.

\bibitem[Scarselli et~al.(2009)Scarselli, Gori, Tsoi, Hagenbuchner, and
  Monfardini]{Scarselli2009_The}
Franco Scarselli, Marco Gori, Ah~Chung Tsoi, Markus Hagenbuchner, and Gabriele
  Monfardini.
\newblock The graph neural network model.
\newblock \emph{{IEEE} Trans. Neural Networks}, 20\penalty0 (1):\penalty0
  61--80, 2009.
\newblock \doi{10.1109/TNN.2008.2005605}.
\newblock URL \url{https://doi.org/10.1109/TNN.2008.2005605}.

\bibitem[Sch{\"{o}}lkopf et~al.(2000)Sch{\"{o}}lkopf, Smola, Williamson, and
  Bartlett]{Scholkopf2000_New}
Bernhard Sch{\"{o}}lkopf, Alexander~J. Smola, Robert~C. Williamson, and
  Peter~L. Bartlett.
\newblock New support vector algorithms.
\newblock \emph{Neural Computation}, 12\penalty0 (5):\penalty0 1207--1245,
  2000.
\newblock \doi{10.1162/089976600300015565}.
\newblock URL \url{https://doi.org/10.1162/089976600300015565}.

\bibitem[Schulman et~al.(2015)Schulman, Levine, Abbeel, Jordan, and
  Moritz]{Schulman2015_Trust}
John Schulman, Sergey Levine, Pieter Abbeel, Michael~I. Jordan, and Philipp
  Moritz.
\newblock Trust region policy optimization.
\newblock In \emph{Proceedings of the 32nd International Conference on Machine
  Learning, {ICML} 2015, Lille, France, 6-11 July 2015}, pages 1889--1897,
  2015.
\newblock URL \url{http://jmlr.org/proceedings/papers/v37/schulman15.html}.

\bibitem[Schulman et~al.(2017)Schulman, Wolski, Dhariwal, Radford, and
  Klimov]{Schulman2017_Proximal}
John Schulman, Filip Wolski, Prafulla Dhariwal, Alec Radford, and Oleg Klimov.
\newblock Proximal policy optimization algorithms.
\newblock \emph{CoRR}, abs/1707.06347, 2017.
\newblock URL \url{http://arxiv.org/abs/1707.06347}.

\bibitem[Schuster and Paliwal(1997)]{Schuster1997_Bidirectional}
Mike Schuster and Kuldip~K. Paliwal.
\newblock Bidirectional recurrent neural networks.
\newblock \emph{{IEEE} Trans. Signal Processing}, 45\penalty0 (11):\penalty0
  2673--2681, 1997.
\newblock \doi{10.1109/78.650093}.
\newblock URL \url{https://doi.org/10.1109/78.650093}.

\bibitem[Sciuto et~al.(2019)Sciuto, Yu, Jaggi, Musat, and
  Salzmann]{Sciuto2019_Evaluating}
Christian Sciuto, Kaicheng Yu, Martin Jaggi, Claudiu Musat, and Mathieu
  Salzmann.
\newblock Evaluating the search phase of neural architecture search.
\newblock \emph{CoRR}, abs/1902.08142, 2019.
\newblock URL \url{http://arxiv.org/abs/1902.08142}.

\bibitem[Sinn et~al.(2019)Sinn, Wistuba, Buesser, Nicolae, and
  Tran]{Sinn2019_Evolutionary}
Mathieu Sinn, Martin Wistuba, Beat Buesser, Maria-Irina Nicolae, and Minh Tran.
\newblock Evolutionary search for adversarially robust neural networks.
\newblock \emph{Safe Machine Learning Workshop at {ICLR} 2019, New Orleans,
  Louisiana, USA}, 2019.

\bibitem[Smithson et~al.(2016)Smithson, Yang, Gross, and
  Meyer]{Smithson2016_Neural}
Sean~C. Smithson, Guang Yang, Warren~J. Gross, and Brett~H. Meyer.
\newblock Neural networks designing neural networks: multi-objective
  hyper-parameter optimization.
\newblock In \emph{{ICCAD}}, page 104. {ACM}, 2016.

\bibitem[Snoek et~al.(2012)Snoek, Larochelle, and Adams]{Snoek2012_Practical}
Jasper Snoek, Hugo Larochelle, and Ryan~P. Adams.
\newblock Practical bayesian optimization of machine learning algorithms.
\newblock In \emph{Advances in Neural Information Processing Systems 25: 26th
  Annual Conference on Neural Information Processing Systems 2012. Proceedings
  of a meeting held December 3-6, 2012, Lake Tahoe, Nevada, United States.},
  pages 2960--2968, 2012.
\newblock URL
  \url{http://papers.nips.cc/paper/4522-practical-bayesian-optimization-of-machine-learning-algorithms}.

\bibitem[So et~al.(2019)So, Le, and Liang]{So2019_The}
David So, Quoc Le, and Chen Liang.
\newblock The evolved transformer.
\newblock In \emph{Proceedings of the 36th International Conference on Machine
  Learning, {ICML} 2019, 9-15 June 2019, Long Beach, California, {USA}}, pages
  5877--5886, 2019.
\newblock URL \url{http://proceedings.mlr.press/v97/so19a.html}.

\bibitem[Suganuma et~al.(2017)Suganuma, Shirakawa, and Nagao]{Suganuma2017_A}
Masanori Suganuma, Shinichi Shirakawa, and Tomoharu Nagao.
\newblock A genetic programming approach to designing convolutional neural
  network architectures.
\newblock In \emph{Proceedings of the Genetic and Evolutionary Computation
  Conference, {GECCO} 2017, Berlin, Germany, July 15-19, 2017}, pages 497--504,
  2017.
\newblock \doi{10.1145/3071178.3071229}.
\newblock URL \url{https://doi.org/10.1145/3071178.3071229}.

\bibitem[Suganuma et~al.(2018)Suganuma, Ozay, and
  Okatani]{Suganuma2018_Exploiting}
Masanori Suganuma, Mete Ozay, and Takayuki Okatani.
\newblock Exploiting the potential of standard convolutional autoencoders for
  image restoration by evolutionary search.
\newblock In \emph{Proceedings of the 35th International Conference on Machine
  Learning, {ICML} 2018, Stockholmsm{\"{a}}ssan, Stockholm, Sweden, July 10-15,
  2018}, pages 4778--4787, 2018.
\newblock URL \url{http://proceedings.mlr.press/v80/suganuma18a.html}.

\bibitem[Sutton and Barto(1998)]{Sutton1998_Reinforcement}
Richard~S. Sutton and Andrew~G. Barto.
\newblock \emph{Reinforcement learning - an introduction}.
\newblock Adaptive computation and machine learning. {MIT} Press, 1998.
\newblock ISBN 0262193981.
\newblock URL \url{http://www.worldcat.org/oclc/37293240}.

\bibitem[Tan et~al.(2018)Tan, Chen, Pang, Vasudevan, and Le]{Tan2018_MnasNet}
Mingxing Tan, Bo~Chen, Ruoming Pang, Vijay Vasudevan, and Quoc~V. Le.
\newblock Mnasnet: Platform-aware neural architecture search for mobile.
\newblock \emph{CoRR}, abs/1807.11626, 2018.
\newblock URL \url{http://arxiv.org/abs/1807.11626}.

\bibitem[Villani(2008)]{Villani2008_Optimal}
C\'{e}dric Villani.
\newblock \emph{Optimal Transport: Old and New}.
\newblock Grundlehren der mathematischen Wissenschaften. Springer Berlin
  Heidelberg, 2008.
\newblock ISBN 9783540710509.

\bibitem[Wang et~al.(2019)Wang, Zhao, Jinnai, Tian, and
  Fonseca]{Wang2019_AlphaX}
Linnan Wang, Yiyang Zhao, Yuu Jinnai, Yuandong Tian, and Rodrigo Fonseca.
\newblock Alphax: exploring neural architectures with deep neural networks and
  monte carlo tree search.
\newblock \emph{CoRR}, abs/1903.11059, 2019.
\newblock URL \url{http://arxiv.org/abs/1903.11059}.

\bibitem[Watkins(1989)]{Watkins1989_learning}
Christopher John Cornish~Hellaby Watkins.
\newblock \emph{Learning from Delayed Rewards}.
\newblock PhD thesis, King's College, Cambridge, UK, May 1989.
\newblock URL \url{http://www.cs.rhul.ac.uk/~chrisw/new_thesis.pdf}.

\bibitem[Weng et~al.(2019)Weng, Zhou, Li, and Qiu]{Weng2019_NASUnet}
Yu~Weng, Tianbao Zhou, Yujie Li, and Xiaoyu Qiu.
\newblock Nas-unet: Neural architecture search for medical image segmentation.
\newblock \emph{{IEEE} Access}, 7:\penalty0 44247--44257, 2019.
\newblock \doi{10.1109/ACCESS.2019.2908991}.
\newblock URL \url{https://doi.org/10.1109/ACCESS.2019.2908991}.

\bibitem[Williams(1992)]{Williams1992_Simple}
Ronald~J. Williams.
\newblock Simple statistical gradient-following algorithms for connectionist
  reinforcement learning.
\newblock \emph{Machine Learning}, 8:\penalty0 229--256, 1992.
\newblock \doi{10.1007/BF00992696}.
\newblock URL \url{https://doi.org/10.1007/BF00992696}.

\bibitem[Wilson et~al.(2016)Wilson, Hu, Salakhutdinov, and
  Xing]{Wilson2016_Deep}
Andrew~Gordon Wilson, Zhiting Hu, Ruslan Salakhutdinov, and Eric~P. Xing.
\newblock Deep kernel learning.
\newblock In \emph{Proceedings of the 19th International Conference on
  Artificial Intelligence and Statistics, {AISTATS} 2016, Cadiz, Spain, May
  9-11, 2016}, pages 370--378, 2016.
\newblock URL \url{http://proceedings.mlr.press/v51/wilson16.html}.

\bibitem[Wistuba(2018{\natexlab{a}})]{Wistuba2018_Deep}
Martin Wistuba.
\newblock Deep learning architecture search by neuro-cell-based evolution with
  function-preserving mutations.
\newblock In \emph{{ECML/PKDD} {(2)}}, volume 11052 of \emph{Lecture Notes in
  Computer Science}, pages 243--258. Springer, 2018{\natexlab{a}}.
\newblock URL
  \url{http://www.ecmlpkdd2018.org/wp-content/uploads/2018/09/108.pdf}.

\bibitem[Wistuba(2018{\natexlab{b}})]{Wistuba2018_Practical}
Martin Wistuba.
\newblock Practical deep learning architecture optimization.
\newblock In \emph{5th {IEEE} International Conference on Data Science and
  Advanced Analytics, {DSAA} 2018, Turin, Italy, October 1-3, 2018}, pages
  263--272, 2018{\natexlab{b}}.
\newblock \doi{10.1109/DSAA.2018.00037}.
\newblock URL \url{https://doi.org/10.1109/DSAA.2018.00037}.

\bibitem[Wistuba and Pedapati(2019)]{Wistuba2019_Inductive}
Martin Wistuba and Tejaswini Pedapati.
\newblock Inductive transfer for neural architecture optimization.
\newblock \emph{CoRR}, abs/1903.03536, 2019.
\newblock URL \url{http://arxiv.org/abs/1903.03536}.

\bibitem[Wong et~al.(2018)Wong, Houlsby, Lu, and Gesmundo]{Wong2018_Transfer}
Catherine Wong, Neil Houlsby, Yifeng Lu, and Andrea Gesmundo.
\newblock Transfer learning with neural automl.
\newblock In \emph{Advances in Neural Information Processing Systems 31: Annual
  Conference on Neural Information Processing Systems 2018, NeurIPS 2018, 3-8
  December 2018, Montr{\'{e}}al, Canada.}, pages 8366--8375, 2018.
\newblock URL
  \url{http://papers.nips.cc/paper/8056-transfer-learning-with-neural-automl}.

\bibitem[Xie and Yuille(2017)]{Xie2017_Genetic}
Lingxi Xie and Alan~L. Yuille.
\newblock Genetic {CNN}.
\newblock In \emph{{IEEE} International Conference on Computer Vision, {ICCV}
  2017, Venice, Italy, October 22-29, 2017}, pages 1388--1397. {IEEE} Computer
  Society, 2017.
\newblock \doi{10.1109/ICCV.2017.154}.
\newblock URL \url{https://doi.org/10.1109/ICCV.2017.154}.

\bibitem[Xie et~al.(2017)Xie, Girshick, Doll{\'{a}}r, Tu, and
  He]{Xie2017_Aggregated}
Saining Xie, Ross~B. Girshick, Piotr Doll{\'{a}}r, Zhuowen Tu, and Kaiming He.
\newblock Aggregated residual transformations for deep neural networks.
\newblock In \emph{2017 {IEEE} Conference on Computer Vision and Pattern
  Recognition, {CVPR} 2017, Honolulu, HI, USA, July 21-26, 2017}, pages
  5987--5995, 2017.
\newblock \doi{10.1109/CVPR.2017.634}.
\newblock URL \url{https://doi.org/10.1109/CVPR.2017.634}.

\bibitem[Xie et~al.(2019{\natexlab{a}})Xie, Kirillov, Girshick, and
  He]{Xie2019_Exploring}
Saining Xie, Alexander Kirillov, Ross~B. Girshick, and Kaiming He.
\newblock Exploring randomly wired neural networks for image recognition.
\newblock \emph{CoRR}, abs/1904.01569, 2019{\natexlab{a}}.
\newblock URL \url{http://arxiv.org/abs/1904.01569}.

\bibitem[Xie et~al.(2019{\natexlab{b}})Xie, Zheng, Liu, and Lin]{Xie2019_SNAS}
Sirui Xie, Hehui Zheng, Chunxiao Liu, and Liang Lin.
\newblock {SNAS}: stochastic neural architecture search.
\newblock In \emph{Proceedings of the International Conference on Learning
  Representations, {ICLR} 2019, New Orleans, Louisiana, USA},
  2019{\natexlab{b}}.
\newblock URL \url{https://openreview.net/forum?id=rylqooRqK7}.

\bibitem[Xue et~al.(2019)Xue, Yan, Yan, Chu, Hu, and
  Lin]{Xue_2019_Transferable}
Chao Xue, Junchi Yan, Rong Yan, Stephen~M. Chu, Yonggang Hu, and Yonghua Lin.
\newblock Transferable automl by model sharing over grouped datasets.
\newblock In \emph{The IEEE Conference on Computer Vision and Pattern
  Recognition (CVPR)}, June 2019.

\bibitem[Yamada et~al.(2016)Yamada, Iwamura, and Kise]{Yamada2016_Deep}
Yoshihiro Yamada, Masakazu Iwamura, and Koichi Kise.
\newblock Deep pyramidal residual networks with separated stochastic depth.
\newblock \emph{CoRR}, abs/1612.01230, 2016.
\newblock URL \url{http://arxiv.org/abs/1612.01230}.

\bibitem[Yamada et~al.(2018)Yamada, Iwamura, and Kise]{Yamada2018_ShakeDrop}
Yoshihiro Yamada, Masakazu Iwamura, and Koichi Kise.
\newblock Shakedrop regularization.
\newblock \emph{CoRR}, abs/1802.02375, 2018.
\newblock URL \url{http://arxiv.org/abs/1802.02375}.

\bibitem[Zagoruyko and Komodakis(2016)]{Zagoruyko2016_Wide}
Sergey Zagoruyko and Nikos Komodakis.
\newblock Wide residual networks.
\newblock In \emph{Proceedings of the British Machine Vision Conference 2016,
  {BMVC} 2016, York, UK, September 19-22, 2016}, 2016.
\newblock URL \url{http://www.bmva.org/bmvc/2016/papers/paper087/index.html}.

\bibitem[Zhang et~al.(2019)Zhang, Ren, and Urtasun]{Zhang2019_Graph}
Chris Zhang, Mengye Ren, and Raquel Urtasun.
\newblock Graph hypernetworks for neural architecture search.
\newblock In \emph{Proceedings of the International Conference on Learning
  Representations, {ICLR} 2019, New Orleans, Louisiana, USA}, 2019.
\newblock URL \url{https://openreview.net/forum?id=rkgW0oA9FX}.

\bibitem[Zhang et~al.(2018{\natexlab{a}})Zhang, Ciss{\'{e}}, Dauphin, and
  Lopez{-}Paz]{Zhang2018_mixup}
Hongyi Zhang, Moustapha Ciss{\'{e}}, Yann~N. Dauphin, and David Lopez{-}Paz.
\newblock mixup: Beyond empirical risk minimization.
\newblock In \emph{6th International Conference on Learning Representations,
  {ICLR} 2018, Vancouver, BC, Canada, April 30 - May 3, 2018, Conference Track
  Proceedings}, 2018{\natexlab{a}}.
\newblock URL \url{https://openreview.net/forum?id=r1Ddp1-Rb}.

\bibitem[Zhang et~al.(2018{\natexlab{b}})Zhang, Zhou, Lin, and
  Sun]{Zhang2018_ShuffleNet}
Xiangyu Zhang, Xinyu Zhou, Mengxiao Lin, and Jian Sun.
\newblock Shufflenet: An extremely efficient convolutional neural network for
  mobile devices.
\newblock In \emph{2018 {IEEE} Conference on Computer Vision and Pattern
  Recognition, {CVPR} 2018, Salt Lake City, UT, USA, June 18-22, 2018}, pages
  6848--6856, 2018{\natexlab{b}}.
\newblock \doi{10.1109/CVPR.2018.00716}.
\newblock URL
  \url{http://openaccess.thecvf.com/content\_cvpr\_2018/html/Zhang\_ShuffleNet\_An\_Extremely\_CVPR\_2018\_paper.html}.

\bibitem[Zhang et~al.(2017)Zhang, Li, Loy, and Lin]{Zhang2017_PolyNet}
Xingcheng Zhang, Zhizhong Li, Chen~Change Loy, and Dahua Lin.
\newblock Polynet: {A} pursuit of structural diversity in very deep networks.
\newblock In \emph{2017 {IEEE} Conference on Computer Vision and Pattern
  Recognition, {CVPR} 2017, Honolulu, HI, USA, July 21-26, 2017}, pages
  3900--3908, 2017.
\newblock \doi{10.1109/CVPR.2017.415}.
\newblock URL \url{https://doi.org/10.1109/CVPR.2017.415}.

\bibitem[Zhong et~al.(2018)Zhong, Yan, Wu, Shao, and Liu]{Zhong2018_Practical}
Zhao Zhong, Junjie Yan, Wei Wu, Jing Shao, and Cheng{-}Lin Liu.
\newblock Practical block-wise neural network architecture generation.
\newblock In \emph{2018 {IEEE} Conference on Computer Vision and Pattern
  Recognition, {CVPR} 2018, Salt Lake City, UT, USA, June 18-22, 2018}, pages
  2423--2432, 2018.
\newblock \doi{10.1109/CVPR.2018.00257}.
\newblock URL
  \url{http://openaccess.thecvf.com/content\_cvpr\_2018/html/Zhong\_Practical\_Block-Wise\_Neural\_CVPR\_2018\_paper.html}.

\bibitem[Zhou et~al.(2018)Zhou, Ebrahimi, Arik, Yu, Liu, and
  Diamos]{Zhou2018_Resource}
Yanqi Zhou, Siavash Ebrahimi, Sercan~{\"{O}}mer Arik, Haonan Yu, Hairong Liu,
  and Greg Diamos.
\newblock Resource-efficient neural architect.
\newblock \emph{CoRR}, abs/1806.07912, 2018.
\newblock URL \url{http://arxiv.org/abs/1806.07912}.

\bibitem[Zoph and Le(2017)]{Zoph2017_Neural}
Barret Zoph and Quoc~V. Le.
\newblock Neural architecture search with reinforcement learning.
\newblock In \emph{5th International Conference on Learning Representations,
  {ICLR} 2017, Toulon, France, April 24-26, 2017, Conference Track
  Proceedings}, 2017.
\newblock URL \url{https://openreview.net/forum?id=r1Ue8Hcxg}.

\bibitem[Zoph et~al.(2018)Zoph, Vasudevan, Shlens, and Le]{Zoph2018_Learning}
Barret Zoph, Vijay Vasudevan, Jonathon Shlens, and Quoc~V. Le.
\newblock Learning transferable architectures for scalable image recognition.
\newblock In \emph{2018 {IEEE} Conference on Computer Vision and Pattern
  Recognition, {CVPR} 2018, Salt Lake City, UT, USA, June 18-22, 2018}, pages
  8697--8710, 2018.
\newblock \doi{10.1109/CVPR.2018.00907}.
\newblock URL
  \url{http://openaccess.thecvf.com/content\_cvpr\_2018/html/Zoph\_Learning\_Transferable\_Architectures\_CVPR\_2018\_paper.html}.

\end{thebibliography}

\end{document}